\documentclass[a4paper,fleqn]{cas-dc}
\renewcommand{\textcolor}[2]{{\color{black}#2}}

\usepackage[numbers]{natbib}

\usepackage{lscape}
\usepackage{rotating}
\usepackage{longtable}
\usepackage{adjustbox}
\usepackage{microtype}

\usepackage{caption}
\usepackage{subcaption}

\usepackage{enumerate}
\usepackage[shortlabels]{enumitem}
\usepackage{multirow}

\usepackage{lipsum}

\begin{document}
\let\WriteBookmarks\relax
\def\floatpagepagefraction{1}
\def\textpagefraction{.001}
\shorttitle{Distributed Intelligence on the Edge-to-Cloud Continuum: A Systematic Literature Review}
\shortauthors{D Rosendo et~al.}

\title [mode = title]{Distributed Intelligence on the Edge-to-Cloud Continuum:\\ A Systematic Literature Review}



\author[1]{Daniel Rosendo}[
                        orcid=0000-0003-1175-8426
                        ]


\address[1]{University of Rennes, Inria, CNRS, IRISA, Rennes, France}

\author[1]{Alexandru Costan}[
                        orcid=0000-0003-3111-6308
                        ]

\author[2]{Patrick Valduriez}[%
    orcid=0000-0001-6506-7538
   ]


\address[2]{University of Montpellier, Inria, CNRS, LIRMM, Montpellier, France}

\author[1]{Gabriel Antoniu}[
                        orcid=0000-0001-6525-3736
                        ]






\begin{abstract}
The explosion of data volumes generated by an increasing number of applications is strongly impacting the evolution of distributed digital infrastructures for data analytics and machine learning (ML). While data analytics used to be mainly performed on cloud infrastructures, the rapid development of IoT infrastructures and the requirements for low-latency, secure processing has motivated the development of edge analytics. Today, to balance various trade-offs, ML-based analytics tends to increasingly leverage an interconnected ecosystem that allows complex applications to be executed on hybrid infrastructures where IoT Edge devices are interconnected to Cloud/HPC systems in what is called the \emph{Computing Continuum}, the \emph{Digital Continuum}, or the \emph{Transcontinuum}.

Enabling learning-based analytics on such complex infrastructures is challenging. The large scale and optimized deployment of learning-based workflows across the Edge-to-Cloud Continuum requires extensive and reproducible experimental analysis of the application execution on representative testbeds. This is necessary to help understand the performance trade-offs that result from combining a variety of learning paradigms and supportive frameworks. A thorough experimental analysis requires the assessment of the impact of multiple factors, such as: model accuracy, training time, network overhead, energy consumption, processing latency, among others.

This review aims at providing a comprehensive vision of the main state-of-the-art libraries and frameworks for machine learning and data analytics available today. It describes the main learning paradigms enabling learning-based analytics on the Edge-to-Cloud Continuum. The main simulation, emulation, deployment systems, and testbeds for experimental research on the Edge-to-Cloud Continuum available today are also surveyed. Furthermore, we analyze how the selected systems provide support for experiment reproducibility. We conclude our review with a detailed discussion of relevant open research challenges and of future directions in this domain such as: holistic understanding of performance; performance optimization of applications;efficient deployment of Artificial Intelligence (AI) workflows on highly heterogeneous infrastructures; and reproducible analysis of experiments on the Computing Continuum.

\end{abstract}



\begin{keywords}
Edge computing \sep Distributed Intelligence \sep Big Data Analytics \sep Computing Continuum \sep Reproducibility
\end{keywords}

\maketitle

\section{Introduction}
\label{sec:introduction}

The current digital revolution is impacting human beings in the way they live, work, learn, and communicate. This has resulted in impressive progress in many areas such as Cloud Computing, High-Performance Computing (HPC), Artificial Intelligence (AI), Big Data Analytics, and the Internet of Things. Furthermore, new challenging application scenarios are emerging from a variety of domains such as autonomous vehicles, real-time manufacturing, precision agriculture, smart cities, to cite just a few~\cite{wang2018adaptive, midya2018multi}.

The explosion of data generated by many applications in the aforementioned areas and the need for real-time analytics and fast decision making has resulted in a shift of the data processing paradigms, as well as of Machine Learning (ML) paradigms, from centralized approaches towards decentralized and multi-tier computing infrastructures and services~\cite{mahmood2018fog}.
Data processing and AI workflows can no longer rely on traditional approaches that send all data to centralized and distant Cloud datacenters for processing or AI model training and inference. Instead, they need to leverage myriads of resources close to the data generation sites (\emph{i.e.}, in the Edge or Fog) in order to promptly extract insights~\cite{asch2018big} and satisfy the ultra-low latency requirements of applications, while keeping reasonable resource usage and preserving privacy constraints. In practice, to balance contradictory requirements, in many situations it makes sense to weight the respective benefits of centralization and decentralization and make appropriate trade-offs to smartly use the advantages of each type of infrastructure.

This contributes to the emergence of what is called the \emph{Computing Continuum}~\cite{etp4-hpc-20} (or the \emph{Digital Continuum} or the \emph{Transcontinuum}). It seamlessly combines resources and services at the center of the network (\emph{e.g.}, in Cloud datacenters), at its Edge, and \emph{in-transit}, along the data path. Typically, data is first generated and preprocessed (\emph{e.g.}, filtering, basic inference) on Edge devices, while Fog nodes further process partially aggregated data. Then, if required, data is transferred to HPC-enabled Clouds for Big Data analytics, Artificial Intelligence model training, and global simulations.

Due to the complexity incurred by application deployments on such highly distributed and heterogeneous Edge-to-Cloud infrastructures, the Computing Continuum vision remains to be realized in practice. Deploying, analyzing, and optimizing large-scale, real-life applications on such infrastructures requires configuring a myriad of system-specific parameters (\emph{e.g.,} from AI and Big Data systems, applications, ingestion systems, among others) and reconciling many requirements or constraints in terms of interoperability, mobility, communication latency, network efficiency, data privacy, and hardware resource consumption (\emph{e.g.,} GPU memory, CPU power, storage size, and others)~\cite{xia2018combining}.


Furthermore, enabling intelligence on the Edge-to-Cloud Continuum to allow fast and accurate decision making requires the efficient deployment of complex AI workflows on massively distributed infrastructures composed by heterogeneous resources. Therefore, enabling intelligence on the Computing Continuum requires the reproducible and extensive evaluations of AI workflow deployments exploring the combination of a variety of ML paradigms and frameworks and analyzing their performance trade-offs and impact on performance metrics such as model accuracy, training time, network overhead, energy consumption and application processing latency.

This systematic literature review provides a comprehensive vision of the main state-of-the-art libraries and frameworks for ML and Data Analytics. It also describes the main learning paradigms for enabling intelligence on the Computing Continuum.
The main contributions of this paper are:

\begin{enumerate}
    \item A taxonomy of Data Analytics and AI libraries and frameworks, and ML paradigms that may compose Edge-to-Cloud workflows to enable intelligence on the Computing Continuum.
    \item A synthetic presentation of the main systems for simulation, emulation, and deployment, as well as the relevant large scale testbeds for experimental evaluation of complex Edge-to-Cloud workflows.
    \item An analysis of how the studies included in our systematic review provide support for experiment reproducibility, an important requirement of the research community that allows scientific claims to be verified by others. We evaluated each article in terms of: (a) access to artifacts; (b) definition of the experimental setup; and (c) access to results.
    \item A discussion of the relevant open research challenges and future directions to enable intelligence on the Edge-to-Cloud Continuum, such as: holistic understanding of performance of applications; performance optimization of Edge-to-Cloud workflows; efficient deployment of complex AI workflows on highly heterogeneous infrastructures; and support of the reproducible analysis of Edge-to-Cloud experiments.
\end{enumerate}

The remainder of this paper is organized as follows. First, we compare our work with the existing surveys/reviews and motivate the need for our review in Section~\ref{sec:related-work}. In Section~\ref{sec:review-methodology}, we describe the methodology exploited to guide our systematic review. Then, we provide answers to the research questions raised by our methodology in Sections~\ref{sec:ai-cc} to~\ref{sec:issues}. In Section~\ref{sec:ai-cc}, we present the main frameworks and libraries for ML in the Edge and in the Cloud. In Section~\ref{sec:bda-cc} we present the main frameworks and libraries for Data Analytics. Next, Section~\ref{sec:ai-bda} discusses recent efforts on combining ML and Data Analytics across the Edge-to-Cloud Continuum, as well as the main learning paradigms used. Section~\ref{sec:experimental_research} presents the main systems for simulation, emulation and deployment for experimental research and the relevant large-scale testbeds. Furthermore, it presents how the selected studies provide support for the experiment reproducibility. Finally, Section~\ref{sec:major-findings} highlights the major findings and Section~\ref{sec:issues} discusses the research challenges in this area. Section~\ref{sec:conclusions} concludes this review.

\section{\textcolor{blue}{Related Work and Motivation}}
\label{sec:related-work}

\textcolor{blue}{Previous surveys and systematic reviews in the context of the Computing Continuum focused on a variety of domains, such as: resource management~\cite{bendechache2020simulating, mijuskovic2021resource}, security and privacy~\cite{ometov2022survey, ansari2020security}, architectures~\cite{debauche2021cloud, chao2019ecosystem, alli2020fog, hamdan2020edge}, robotics~\cite{vermesan2020internet}, blockchain~\cite{spataru2021review, gill2019transformative}, just to cite a few. The scope of our work is larger: while focusing on distributed intelligence on the continuum, we review articles in the fields of Machine Learning (ML) and Data Analytics (DA) applied on Edge, Cloud, and Edge-to-Cloud environments. Below we discuss how our work compares to recent surveys in these fields.}

\textcolor{blue}{
\textbf{Machine and Deep Learning across the Edge-to-Cloud Continuum.}
A recent survey~\cite{angel2022recent} explores evolving computing paradigms such as Edge, Fog, and Cloud highlighting the latest innovations resulted from their fusion with ML. The authors discuss open research challenges such as: scalability, deployment, failure management, hardware heterogeneity, resource management, security, and interoperability. Furthermore, they present future prospects, including: Big Data Analytics for fast data-driven decision making;  Artificial Intelligence to enhance resource management, energy management, security, and reliability; Serverless Computing for leveraging the infrastructure scalability and decreasing the application response time, latency, and energy consumption.}

\textcolor{blue}{A review on ML for data processing and management tasks across the Edge-to-Cloud continuum is presented in~\cite{samie2019cloud}. The authors categorize the usage of ML according to the application domain, ML techniques, input data type, and where they belong in the continuum. Besides, they discuss the research trends toward efficient ML on the edge, in particular: the optimization of ML techniques to reduce their power consumption, memory requirement, and computation intensity; efficient hardware for embedded ML; offloading ML tasks among Edge-to-Cloud resources; and collaborative ML training.}

\textcolor{blue}{In~\cite{mwase2022communication}, the authors review communication-efficient distributed Machine Learning strategies for the Edge-to-Cloud continuum. They introduce the principles of distributed ML operations and approaches of implementing parallelism and distribution. Furthermore, authors discuss communication inefficiencies in distributed ML on the Edge and the existing communication-efficient processing techniques for training in resource-limited devices. Lastly, they present research directions where further advancements in communication-efficient distributed ML may be made.}

\textcolor{blue}{A review of deep learning applications such as computer vision, virtual and augmented reality, and natural language processing running on Edge devices is presented in~\cite{chen2019deep}. Authors discuss edge-only, hybrid edge-cloud and distributed computing approaches to accelerate deep learning training and inference. They summarize the selected articles in terms of architecture, DNN model, application, key metrics, and Edge hardware used. Lastly, they discuss the challenges in deploying deep learning on Edge-to-Cloud environments, such as: management and scheduling of Edge resources, energy consumption, application migration, benchmarks, and privacy. In this research direction, authors~\cite{vestias2020moving} describe the methods and architectures to execute deep learning inference and training at the edge. They also discuss open issues regarding the deployment of deep learning at the edge.}

\textcolor{blue}{An overview of existing Edge Computing systems is presented in~\cite{liu2019survey}. It discusses techniques to support Deep Learning models at the Edge, for example: (1) systems and toolkits: OpenEI, a framework for Edge Intelligence; AWS IoT Greengrass, for ML Inference; Azure IoT Edge; and Cloud IoT Edge; and (2) open source Deep Learning packages: TensorFlow, Caffe2, PyTorch, MXNet, and some distributed Deep Learning models over Cloud and Edge such as DDNN and Neurosurgeon.} 

\textcolor{blue}{The confluence of IoT and AI detailing their potential applications and open issues is discussed in~\cite{mohammadi2018deep}. It presents the recent approaches for deploying DL on resource constrained Edge devices, Fog and Cloud. They also discuss the two main categories of IoT data generation, such as IoT streaming data and IoT Big Data, as well as their requirements for analytics. Lastly, authors highlight the challenges for the successful merging of DL and IoT applications, such as the lack of real-world datasets for IoT applications; the preprocessing of raw data for DL model training; ensuring data security and privacy in IoT applications; and online resource provisioning for IoT analytics; just to cite a few.}


\textcolor{blue}{
\textbf{Data Analytics across the Edge-to-Cloud Continuum.}
In~\cite{badidi2020fog} the authors provide a review focusing on the efforts of using Big Data Analytics solutions in the Edge-to-Cloud Continuum. They present the relevant Data Analytics platforms (\emph{e.g.}, Hadoop, Flink, Spark, Storm, Nifi, and others) and Machine Learning libraries (\emph{e.g.}, Spark MLlib, Tensorflow, Keras, Scikit-learn, \emph{etc.}) to enable a real-time Big Data pipeline from the Edge to the Cloud. Lastly, authors discuss the following open challenges: interoperability, characterizing smart city applications, and privacy issues.}

\textcolor{blue}{A survey on IoT Big Data Analytics covering Big Data generation, acquisition, storage, learning, and analytics is presented in~\cite{sezer2017context}. It discusses parallel processing models and engines for the analysis of Big Data such as Spark, Flink, and Storm. Regarding IoT Big Data learning, they present Machine Learning frameworks working on Big Data and processing in parallel such as Spark MLlib, SAMOA, and FlinkML. Lastly, authors highlight open issues related to Machine Learning and Big Data Analytics in IoT.}

\textcolor{blue}{A review on Edge, Fog, and Cloud computing infrastructures used for IoT Big Data Analytics is presented in~\cite{atitallah2020leveraging}. Authors review the combination of DL and Big Data Analytics in the development of smart cities and provide a comparison of deep learning frameworks and libraries; models; and datasets used in smart city applications. Furthermore, they review articles exploiting IoT and DL to develop intelligent applications and services for smart cities and outline the challenges in developing such applications.}

\textcolor{blue}{In summary, all these related works \emph{focus on specific domains} such as: Machine Learning on the Edge-to-Cloud continuum~\cite{angel2022recent, samie2019cloud, mwase2022communication}; Deep Learning mainly focusing on the Edge, but also discussing hybrid Edge-Cloud deployments~\cite{chen2019deep, vestias2020moving, liu2019survey, mohammadi2018deep}; and Data Analytics on Edge-to-Cloud environments~\cite{badidi2020fog, sezer2017context, atitallah2020leveraging}.}

\textcolor{blue}{
\paragraph{Motivation: study the challenges of ML and DA convergence across the Continuum.}
As opposed to these previous studies, we are interested in the specific issues (and the frameworks that address them) arising at the frontier of DL and ML, as this combination is rapidly gaining traction as a standard for analytics on the continuum. To the best of our knowledge, our literature review is the first to systematically explore the recent efforts and to summarize the existing approaches on applying Machine Learning, Data Analytics, and their combination to enable distributed intelligence on the Edge, Cloud, and Edge-to-Cloud continuum. Furthermore, our review is unique especially from two main perspectives: (1) it discusses relevant open challenges and research opportunities identified after reviewing the articles; and (2) it provides an extensive analysis of the articles in terms of experimental evaluations and validation, allowing to identify: \emph{(i)} the relevant large-scale testbeds; \emph{(ii)} simulation, emulation, and deployment systems; \emph{(iii)} the common ML/DA frameworks and libraries; metrics; the common models/algorithms, datasets, and Edge hardware; \emph{(iv)} the scale of the testbeds used to validate the proposed solutions; and \emph{(v)} their support for reproducibility.}

\section{Review Methodology}
\label{sec:review-methodology}

\begin{figure}[t]
  \centering
  \includegraphics[width=\linewidth]{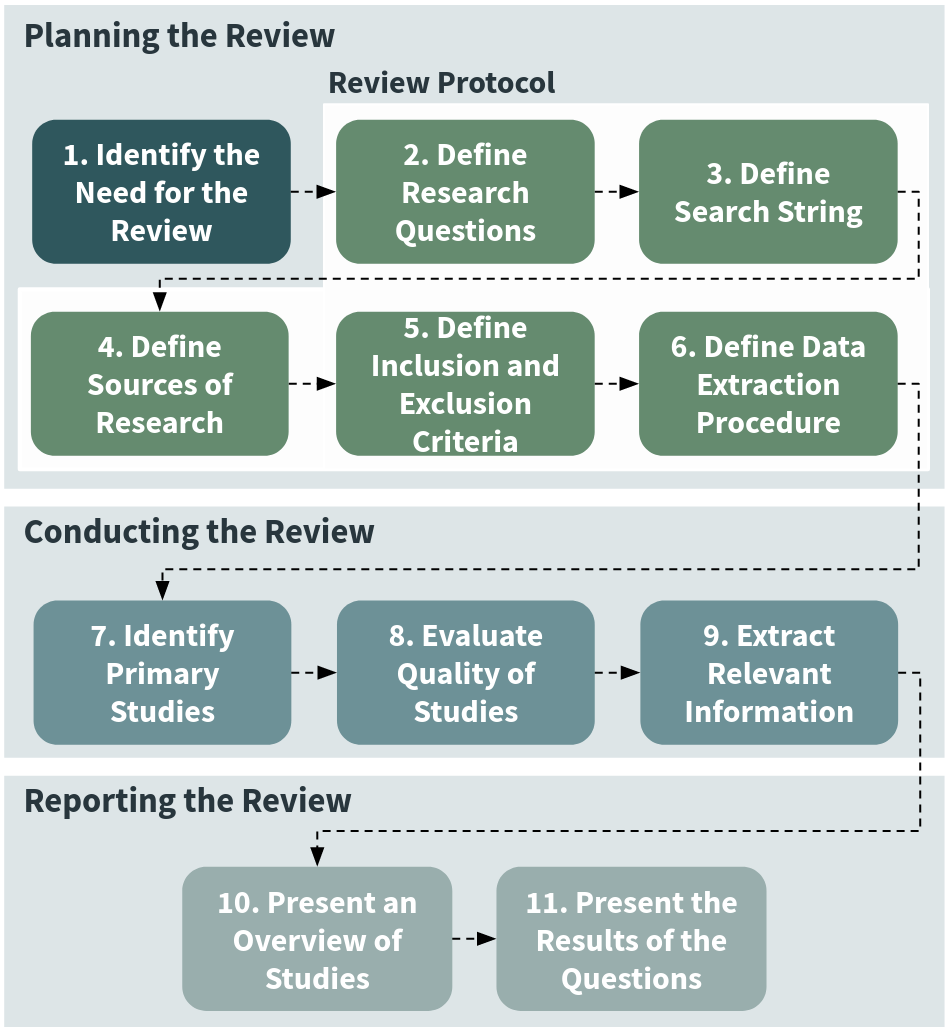}
  \caption{Systematic review methodology.}
  \label{fig:sr-meth}
\end{figure}

The systematic review methodology leveraged in this work is based on~\cite{keele2007guidelines, endo2016high}. The Figure~\ref{fig:sr-meth} illustrates the three main processes of the review, which are: 1) Planning the Review; 2) Conducting the Review; and 3) Reporting the Review. Next, we describe their corresponding activities in detail.

\subsection{Planning the Review}
In more and more application areas, we are witnessing the emergence of complex workflows that combine computing, analytics and learning. Such application workflows are evolving towards an interconnected ecosystem that often require a hybrid execution infrastructure from IoT devices to Cloud/HPC systems (aka \emph{Computing Continuum}). A holistic understanding of the complex continuum ecosystem is challenging.

\subsubsection{Identify the Need for the Review}
This systematic review aims to provide a taxonomy of libraries, frameworks, and learning paradigms that compose Edge-to-Cloud workflows to enable intelligent analytics. Furthermore, we highlight systems and testbeds that allow the analysis of such Edge-to-Cloud workflows, as well as the recent efforts to enable the computing continuum vision and the relevant research opportunities.

\subsubsection{Define the Research Questions}
\label{subsubsec:rq}
The objective of the systematic review is to answer the following research questions with a \textbf{focus on the Edge-to-Cloud Continuum}:


\begin{enumerate}[start=1,label={RQ\arabic*.}]
    \item What are the \textbf{main state-of-the-art methods} for \textbf{Machine Learning} and \textbf{Data Analytics}?
    \item How are the \textbf{existing Machine Learning} and \textbf{Data Analytics} approaches \textbf{combined} to enable \textbf{intelligence}?
    \item What are the existing solutions for \textbf{experimental research} and how do the selected studies support the \textbf{reproducibility} of the experiments?
    \item What are the \textbf{open challenges} and \textbf{research opportunities} in this area?
\end{enumerate}

\subsubsection{Define the Search String}
The keywords used in the search queries are: IoT, edge, fog, big data, stream processing, learning and intelligence. Therefore, the search string applied in the scientific databases is: "IoT" AND ("edge" OR "fog") AND "big data" AND "stream processing" AND ("learning" OR "intelligence").

\subsubsection{Define the Sources of Research}
The selected scientific databases are: ScienceDirect, ACM, IEEE Xplore, Springer Link, and Usenix (FAST, NSDI, ATC, HotEdge, and HotCloud).

\subsubsection{Define Inclusion and Exclusion Criteria}
The search scope of this study is limited to journal and conference articles, magazines and book chapters published between January 2016 and August 2021.

\textcolor{blue}{The main characteristics that the articles must present to be included in this systematic review are: (1) help to answer to at least one of the four research questions defined in Subsection~\ref{subsubsec:rq}; and (2) evaluate existing or propose novel systems, frameworks or architectures enabling intelligence on the Edge, Cloud, or Edge-to-Cloud environments. Articles not respecting these requirements are eliminated.}

\subsubsection{Define the Data Extraction Procedure}
\label{subsubsec:extraction-procedure}
The process of extracting information from the articles consists in filling a form that is designed to answer the research questions of the systematic review. Therefore, the form is structured as follows: title, publication year, scientific database, venue, resume of the contributions, framework/libraries for learning or analytics cited, experimental approach, and open challenges or future works.

\subsection{Conducting the Review}
The search string applied on the scientific databases returned a total of 1159 articles: 242 from ScienceDirect; 206 from ACM; 325 from IEEE Xplore; 290 from Springer Link; and 96 from Usenix. 

\subsubsection{Identify the Primary Studies}
\label{subsubsec:primary-studies}
As a refinement step, we started the screening process, which consists in reading the abstract and conclusions for each article. This refinement step eliminates out of scope articles. Finally, we selected a total of 69 papers for quality evaluation and extraction of relevant information.

\begin{figure*}[t]
\centering
\begin{subfigure}{.4\textwidth}
  \centering
  \includegraphics[width=\linewidth]{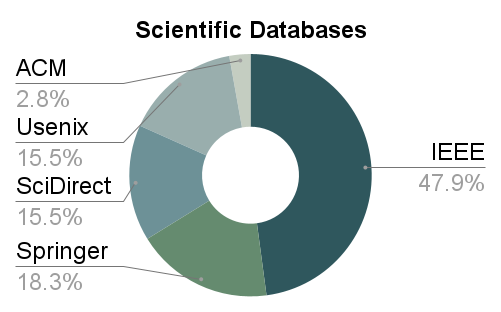}
  \caption{Selected articles by scientific database.}
  \label{fig:databases-percent}
\end{subfigure}%
\hspace{5mm} 
\begin{subfigure}{.4\textwidth}
  \centering
  \includegraphics[width=\linewidth]{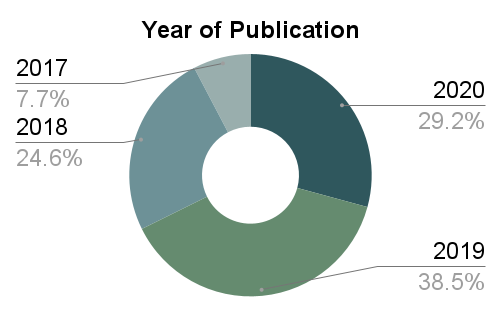}
  \caption{Selected articles by year of publication.}
  \label{fig:year-percent}
\end{subfigure}
\caption{Percentage of selected articles per year of publication and per scientific database.}
\label{fig:year-databases-percent}
\end{figure*}


\begin{table*}[t]
\small
\sffamily
\centering
\caption{Selected articles by area and computing paradigm.}
\label{tbl:articles-by-area}
\begin{tabular}{m{3.5cm}m{1.5cm}m{0.5cm}m{3.0cm}m{1.5cm}m{0.5cm}m{4.0cm}}
\hline
\rowcolor[HTML]{C4CDC1} 
\textbf{Area}                                                   & \textbf{Percentage}                              & \textbf{Qty.}                            & \textbf{Computing Paradigm} & \cellcolor[HTML]{C4CDC1}\textbf{Percentage} & \cellcolor[HTML]{C4CDC1}\textbf{Qty.} & \cellcolor[HTML]{C4CDC1}\textbf{Papers}                                                                                                                                                                                                                                                                                               \\ \hline
                                        &                          &                      & \cellcolor[HTML]{E0EBEA}Edge                        & \cellcolor[HTML]{E0EBEA}58\%                                      & \cellcolor[HTML]{E0EBEA}15                                        & \cellcolor[HTML]{E0EBEA}\cite{zhou2019distributing, kumar2019decaf, lu2019collaborative, zhang2018pcamp, nikouei2019toward, kukreja2019training, xu2019first, zhang2020deep, chen2019deep, guo2021mistify, deng2020edge, li2019edge, fafoutis2018extending, kumar2017resource, dey2019offloaded}                                     \\

\multirow{-2}{*}[10pt]{Machine Learning (ML)} & \multirow{-2}{*}[10pt]{35\%} & \multirow{-2}{*}[10pt]{24} & Edge-to-Cloud               & 42\%                                      & 9                                        & \cite{talagala2018eco, chen2019exploring, sarabia2019efficient, zhou2020saface, mrozek2020fall, loghin2020disruptions, ghosh2020edge, hong2019dlion, rocha2020distributed}                                                                                                                            \\ \hline
                                                                &                                                  &                                              & \cellcolor[HTML]{E0EBEA}Edge                        & \cellcolor[HTML]{E0EBEA}56\%                                      & \cellcolor[HTML]{E0EBEA}9                                         & \cellcolor[HTML]{E0EBEA}\cite{aral2020staleness, dautov2020stream, fu2019edgewise, alencar2020fot, hauswirth2020autonomous, verma2020smart, yang2017iot, huang2018building, dautov2017pushing}                                                                                                                                                               \\ 
\multirow{-2}{*}[3pt]{Data Analytics (DA)}                           & \multirow{-2}{*}[3pt]{23\%}                         & \multirow{-2}{*}[3pt]{16}                         & Edge-to-Cloud               & 44\%                                      & 7                                         & \cite{verma2017survey, dautov2019hierarchical, xu2021move, dautov2018data, sharma2017live, rosendo2020e2clab, kolodziej2019high}                                                                                                                                                                                                     \\ \hline
                                                                &                                                  &                                              & \cellcolor[HTML]{E0EBEA}Cloud                       & \cellcolor[HTML]{E0EBEA}36\%                                      & \cellcolor[HTML]{E0EBEA}10                                        & \cellcolor[HTML]{E0EBEA}\cite{ali2018recent, l2017machine, rao2019big, ulusar2020open, pathak2018construing, nguyen2019machine, lobo2020spiking, assefi2017big, nair2018applying, xian2020parallel}                                                                                                                                                          \\ 
\multirow{-2}{*}[5pt]{Combining ML and DA}                           & \multirow{-2}{*}[3pt]{42\%}                           & \multirow{-2}{*}[3pt]{29}                         & Edge-to-Cloud               & 64\%                                        & 19                                        & \cite{wang2018iot, sezer2017context, mohammadi2018deep, liu2019survey, zhang2019deep, prabhu2019fog, khayyam2020artificial, ali2020res, xu2018survey, yousefpour2019all, grzenda2020hybrid, diez2019data, perez2018resilient, fei2019cps, sankaranarayanan2020data, sarabia2020highly, paakkonen2020extending, kourtellis2021s2ce, rong2021edge} \\ \hline
\end{tabular}
\end{table*}



\subsubsection{Evaluate the Quality of the Studies}
The quality evaluation of the selected articles is based on checking if they are related to techniques or approaches that enable intelligent analytics on the Edge-to-Cloud continuum.

\subsubsection{Extract the Relevant Information}
For each one of the 69 articles selected in \ref{subsubsec:primary-studies} we read the whole paper to extract the relevant information and then fill the form defined in \ref{subsubsec:extraction-procedure}.

\subsection{Reporting the Review}
Lastly, two types of reports are issued: general statistics about the studies and answers to the questions raised by our methodology. 

\subsubsection{Present an Overview of the Studies}
Since the form is filled, we generate relevant statistics derived from a global analysis of the articles. Such statistics are aligned to the research questions defined in \ref{subsubsec:rq} and they are presented in the next sections.

\subsubsection{Present the Answers to the Research Questions}
Finally, according to the relevant information extracted from all articles, we structured the remaining sections of our review based on the research questions. From the information registered in the form, we grouped, defined taxonomies, and summarized all articles in order to answer the research questions.

\textcolor{blue}{Figure~\ref{fig:year-databases-percent} presents the percentage of selected papers per scientific database and per year of publication, respectively. We highlight that after the screening process, no article published in 2016 was selected. Table~\ref{tbl:articles-by-area} summarizes the selected articles by area and computing paradigm exploited.}

\section{
Machine Learning Methods on the Edge-to-Cloud Continuum}
\label{sec:ai-cc}

\begin{figure}[t]
  \centering
  \includegraphics[width=0.8\linewidth]{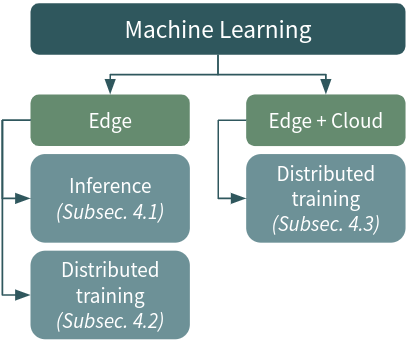}
  \caption{Taxonomy of learning methods on the Edge and Edge-to-Cloud Continuum.}  \label{fig:taxonomy-learning}
\end{figure}

Figure~\ref{fig:taxonomy-learning} presents the taxonomy of learning methods with a focus on the Edge and across the Edge-to-Cloud Continuum. The distributed training can be achieved across the Continuum or among Edge devices, while the inference is typically done at the Edge, for latency purposes. \textcolor{blue}{The Machine Learning frameworks/libraries identified in the articles are presented in Tables~\ref{tbl:ml-cloud} (designed for the Cloud) and~\ref{tbl:ml-edge} (designed for the Edge), respectively. Table~\ref{tbl:learning} characterizes the selected articles with respect to the resources exploited in the experimental evaluations, such as: frameworks and libraries; application/task; metrics; hardware; models; and datasets. Table~\ref{tbl:ai-quantitative} presents a quantitative analysis summarizing Table~\ref{tbl:learning}.} 

\subsection{Inference on the Edge}

\textcolor{blue}{Next, we present the recent efforts to enable inference on resource-limited Edge devices. The following papers focus on hardware and software aspects, as well as, frameworks and algorithms for the efficient inference on Edge devices.}

\textcolor{blue}{In~\cite{deng2020edge} the authors discuss the potential research directions to enable Edge Intelligence. First, they discuss how AI technologies may help to solve complex problems in Edge Computing, such as: service placement; resource provisioning; network planning; mobility management; among others. Furthermore, they explore issues on performing AI on resource-scarce Edge devices and the recent research efforts to solve problems in this direction, such as: frameworks for model training and inference; accelerating DNN computation on hardware; model compression; asynchronous model aggregation, among others.}

\textcolor{blue}{In~\cite{fafoutis2018extending} the authors investigate the benefits of using embedded Machine Learning in wearable sensors to increase battery lifetime. Their approach focus on optimizing data generation and transferring and uses Support Vector Machines for classification. Evaluations show that their approach significantly reduces the amount of data transferred and therefore extends the battery lifetime of resource-constrained sensors.}

\textcolor{blue}{A tree-based algorithm for efficient prediction on IoT devices is proposed in~\cite{kumar2017resource}. Named Bonsai, the algorithm was designed to be fast, accurate, compact and energy-efficient at prediction time. Evaluations show that Bonsai outperforms state-of-the-art algorithms such as kNN, SVM, and single hidden layer NN algorithms in terms of accuracy, model size, inference time, and energy consumption.}

\textcolor{blue}{A novel framework for collaborative DNN inference on Edge devices is proposed in~\cite{li2019edge}. Named Edgent, the framework exploits DNN computation partitioning and DNN right-sizing to enable low-latency inference. Authors evaluate Edgent under static and dynamic bandwidth environments and considered performance metrics for model inference such as: accuracy, latency requirements, and throughput. Through experiments on Raspberry Pi (acting as a mobile device), they demonstrate the effectiveness of Edgent towards low-latency Edge intelligence.}

\textcolor{blue}{In \cite{dey2019offloaded} the authors discuss strategies for accelerating DNN inference by partitioning the model between Edge devices. Next, they evaluate the implementation of an offloading system for Deep Learning inference in a Raspberry Pi 3 with the Intel Movidius hardware accelerator. Experimental results considering metrics such as processing latency, data transfer latency, and network bandwidth demonstrate that intelligent offloading may improve the performance when running in resource constrained Edge devices.}

In~\cite{zhou2019distributing} the authors propose three parallelism schemes (All-In-One, Pipeline, and Parallel) to deploy Deep Neural Networks (DNNs) on resource-constrained devices for inference. Each parallelism approach explores the finer granularity of containerizing a DNN model at the edge. Experimental evaluations show that parallelizing a VGG-16 model for inference starts to improve performance as network speed increases.

Since Edge devices are heterogeneous in terms of hardware characteristics and there is a variety of state-of-the-art Machine Learning packages that can be used for inference at the Edge, in~\cite{zhang2018pcamp} authors investigate how such learning packages perform on different Edge devices. They compare TensorFlow, Caffe2, MXNet, PyTorch, and TensorFlow-Lite running two trained CNN-based models (AlexNet as the large-scale model and SqueezeNet and MobileNet as the small-scale models) on Edge devices such as MacBook, FogNode, Jetson TX2, Raspberry Pi, and Nexus 6P. The performance comparison includes metrics such as latency, memory footprint, and energy.

In~\cite{guo2021mistify} the authors propose a framework to automatically port a Cloud-based model to a suite of models for Edge devices. Named Mistify, the framework decouples the model design (optimized for accuracy) and the deployment (optimized for resource efficiency) phases. Experimental results show that Mistify reduces the DNN porting time needed to cater to a wide spectrum of Edge deployment scenarios by more than 10 times.

\textcolor{blue}{In summary, the articles demonstrated that techniques to minimize the data transfer, reduce the model size, partitioning and offloading the model, and parallelizing the inference among Edge devices, are effective to improve the performance when running in resource constrained Edge devices.}

\subsection{Distributed Training on the Edge}

\textcolor{blue}{Next, we present the recent efforts to enable distributed Machine Learning and Deep Learning training on Edge devices. The following papers focus on lightweight Deep Learning models, learning paradigms, collaborative learning systems, and the frameworks and libraries for optimizing Deep Learning on mobile devices.}

In~\cite{kumar2019decaf} the authors propose a system for enabling iterative collaborative processing (ICP) in resource constrained Edge environments with a focus on Machine Learning applications (\emph{e.g.}, model training). The proposed system consists in a central controller that coordinates all the Edge devices (workers). The controller communicates the initial values of the model parameters to all the Edge devices and updates the model parameters at the end of every iteration using the individual model parameters from all the Edge devices. Lastly, it sends the updated parameters to the workers for next iteration. This process repeats until the model parameters have converged. 

CLONE~\cite{lu2019collaborative} is a collaborative learning setting on the Edge built on top of the Federated Learning algorithm and long short-term memory networks. In CLONE, the learning tasks are solved by a group of distributed Edge nodes. Each Edge node trains the neural network model locally based on its private data and uploads asynchronously the parameters to a \textit{Parameter EdgeServer}. The \textit{EdgeServer} aggregates those parameters and sends them back to Edge devices. Experimental results show that, compared to a stand-alone model training, CLONE reduces training time significantly without sacrificing prediction accuracy.

A Lightweight Convolutional Neural Network (L-CNN) is proposed in~\cite{nikouei2019toward} to enable real-time human identification on a network Edge using fewer resources and preserving the high accuracy of CNNs. In order to enhance performance on Edge, authors propose a hybrid lightweight tracking algorithm named Kerman (Kernelized Kalman filter). Kerman works along with L-CNN to further improve the speed and reliability of feature extraction for human abnormal behavior detection. Experimental results demonstrate that the proposed algorithms can track the humans as objects in real-time with decent accuracy at a resource consumption affordable by Edge devices.

Besides these novel approaches and systems to enable distributed training on the Edge, some recent efforts focus on understanding the performance of learning on the Edge. 


A survey on Deep Learning applied on mobile networking is presented in~\cite{zhang2019deep}. Mobile Data Analytics on Edge devices is achieved either through distributed Machine Learning systems such as MLbase, Gaia, TUX~\cite{xiao2017tux2}, and Adam; or, through Deep Learning libraries such as TensorFlow, Theano, PyTorch, and MXNET. Since mobile networks are ever changing, applications should learn and adapt fast to the domain changes. Therefore, authors discuss learning paradigms such as Online Learning, Lifelong Learning, and Transfer Learning. Lastly, a discussion on open source platforms (\emph{e.g.}, TensorFlow, Caffe, and NCNN) that seek to optimize Deep Learning on mobile devices is presented.

An empirical study of on-device Deep Learning for smartphones such as Android devices is presented in~\cite{xu2019first}. The study includes 21 frameworks based on their popularity (forks and stars on GitHub) in which authors investigate how those frameworks are used in DL applications. Another study~\cite{kukreja2019training} explores scenarios where it is advantageous to do training on the Edge. Experimental results show that peak memory footprint, which is crucial for training on Edge devices, can be reduced by checkpointing strategies such as full binomial checkpointing.



\textcolor{blue}{As a conclusion, the articles demonstrated that lightweight Deep Learning models may help to reduce the resource usage while preserving the model accuracy. Furthermore, collaborative model training strategies on resource-scarce Edge devices have been shown to be effective to reduce the training time without sacrificing accuracy.}


\subsection{Distributed training across the Edge-to-Cloud Continuum}

\textcolor{blue}{The articles presented in this section focus on deploying and distributing the processing of Machine Learning and Deep Learning workloads among Edge and Cloud environments. They propose novel systems and architectures and analyze the performance trade-offs of Cloud only \emph{vs.} Edge-to-Cloud collaborative training.}

\textcolor{blue}{An overview of challenges and of existing approaches to distributed Machine Learning for IoT applications in the Fog is presented in~\cite{rocha2020distributed}. The authors start by presenting the main challenges in processing IoT data, such as generation, transmission, and processing. Then, they highlight the challenges related to the execution of Machine Learning techniques in resource constrained Fog devices. Lastly, the authors present existing approaches to distribute intelligence on Fog devices with a focus on distributed processing and information sharing.}

\textcolor{blue}{\cite{loghin2020disruptions} provides a review of 5G on traditional and emerging technologies and share their ideas on future research challenges and opportunities. In particular, they exploit how 5G can help the development of Federated Learning. They present the domains impacted by 5G such as Edge computing, security and privacy, artificial intelligence, and database systems.}

\textcolor{blue}{A decentralized distributed Deep Learning system named DLion is proposed in~\cite{hong2019dlion}. DLion builds on top of TensorFlow and implements techniques such as compute capacity-aware batching, adaptive model parameter tuning, and network-aware data exchange features in order to reduce training time, improving model accuracy, and providing system scalability for Deep Learning in micro-clouds. Experiments compare DLion with existing distributed Deep Learning systems such as Gaia and Ako, showing that DLion reaches the target accuracy faster than them. Besides, the compute capacity-aware batching technique implemented in DLion helps to reduce the training time.}

\textcolor{blue}{The authors of~\cite{sarabia2019efficient} propose a container-based IoT gateway architecture for Ambient Assisted Living (AAL) scenarios to support the deployment of Deep Learning models. Such models are implemented and trained in the Cloud to detect the fall of people and deployed remotely on the Edge gateways to provide predictive analytics. Results show an improvement of the inference time compared to the Cloud-based approach.}

\textcolor{blue}{Still in the same direction of fall detection, authors of~\cite{mrozek2020fall} propose a system that detects falls on Edge devices (\emph{e.g.}, mobile phones) by using Boosted Decisions Trees. The proposed approach reduces the amount of data and network traffic sent to the Cloud and presents almost the same detection capabilities as the classification process performed in the Cloud.}

\textcolor{blue}{Authors of~\cite{zhou2020saface} propose SAFACE, a three-layer Edge computing system for face recognition. SAFACE employs Unsupervised Learning which can gradually fine-tune a portion of the face recognition model. The three-layer system consists of: Cloud, to train the CNN model; middle servers, for face recognition, fine-tune pre-trained CNN model, and context-aware scheduling; and Edge, for face detection. Experimental results demonstrate its advantages in improving recognition accuracy and reducing processing latency.}

\textcolor{blue}{In~\cite{ghosh2020edge}, a novel Edge-Cloud Machine Learning system is proposed. The system combines Edge and Cloud Computing for IoT Data Analytics by taking advantage of Edge nodes to reduce the network traffic and latency for Machine Learning tasks. The results show that using sliding window techniques, the network traffic can be reduced by up to 80\% without significant loss of accuracy.}

A novel architecture named Edge Cloud Orchestrator (ECO) is proposed in~\cite{talagala2018eco}. This architecture aims to orchestrate and manage Machine Learning deployments and execution across distributed layers in both Edge and Cloud. It supports deployment scenarios such as Federated Learning, Transfer Learning, and Staged Model Deployment. Furthermore, it supports Machine Learning engines and \textcolor{blue}{algorithms such as Spark MLlib (supports a variety of learning algorithms for classification, regression, clustering, among others), FlinkML (supports SVM, multiple linear regression, k-Nearest neighbors, among others), and TensorFlow (supports SVM, Gradient Boosting Machine, Random Forests, Naive Bayes, k-nearest neighbors, \emph{etc.}).}

In~\cite{chen2019exploring} the authors explore the use of Synthetic Gradients (SG) for model-parallel training of a Deep Neural Network (DNN) model. This approach distributes the training of the various layers in the Cloud and resource-limited Edge devices. They compare the feasibility of the SG approach with the conventional back propagation method and evaluate its accuracy and convergence speed considering a four-layered, an eight-layered, and a VGG16 model. Results show that the four-layered model presents comparable performance for SG and back propagation, but an accuracy degradation is observed for the VGG16 model using SG. Regarding the convergence speed, using SG, the model learns slower than the back propagation method even while increasing the number of layers in the model.

\textcolor{blue}{The articles previously presented demonstrate the benefits of collaborative Edge-to-Cloud training. The main performance improvements of such Edge-to-Cloud approaches refer to: reducing the training time without significant loss of accuracy; reducing the amount of data sent to the Cloud and thus the network traffic; and reducing the end-to-end processing latency of Machine Learning and Deep Learning applications.}

\begin{sidewaystable*}
\sffamily
\centering
\caption{\textcolor{blue}{Summary of artifacts, metrics, and hardware exploited in the Machine Learning experiments}}
\label{tbl:learning}
\begin{tabular}{m{1.0cm}m{3.0cm}m{3.0cm}m{4.5cm}m{4.5cm}m{3.0cm}m{2.0cm}}
\hline
\rowcolor[HTML]{C4CDC1} 
\textbf{Paper}               & \textbf{Framework/Library}                            & \textbf{Application/Task}                                   & \textbf{Metrics}                                                                                                                                       & \textbf{Hardware}                                                                                                                   & \textbf{Model}                                                                                          & \textbf{Dataset}                                                      \\ \hline
\rowcolor[HTML]{E0EBEA} 
\cite{zhang2018pcamp}           & TensorFlow, Caffe2, MXNet, PyTorch, and TensorFlow Lite & Inference                                            & inference time; memory footprint; and energy consumption                                                                                      & MacBook Pro, FogNode, Jetson TX2, Raspberry Pi 3 B+, Nexus 6P                                                                       & AlexNet and SqueezeNet                                                                                  & Not informed.                                                \\
\cite{zhou2019distributing}     & TensorFlow                                              & Inference                                            & inference time considering different computation and network conditions                                                                       & VMs with limited capabilities to emulate IoT devices (physical machine: 2x six-core Intel Xeon 2.40 GHz E5-2620 v3 CPUs, 64 GB RAM) & VGG-16                                                                                                  & Not informed.                                                \\
\rowcolor[HTML]{E0EBEA} 
\cite{kumar2019decaf}           & DeCaf, MOCHA                                            & Collaborative processing of Support Vector Machines  & model convergence speed; number of computations performed per iteration;                                                                      & Master: Intel(R) Xeon(R) CPU E5-2620 v3 2.40GHz; and Workers: MacBook Pro 2.9 GHz Intel Core i5                                     & Support Vector Machines                                                                                 & Not informed.                                                \\
\cite{chen2019exploring}        & TensorFlow                                              & Training with back propagation vs synthetic gradient & model accuracy and convergence speed                                                                                                          & 1x Nvidia K40 GPU and Xeon CPU and resource-limited containers to emulate Edge devices (one CPU and 1GB memory)                     & four-layered model; eight-layer model; and VGG16                                                        & MNIST                                                        \\
\rowcolor[HTML]{E0EBEA} 
\cite{lu2019collaborative}      & CLONE, Tensorflow, Keras, and Scikit-Learn              & Model training                                       & training time (from epoch); and evaluation scores including: precision, recall, accuracy, and F-measure                                       & 1x Intel FogNode (Parameter EdgeSever) and 2x Intel FodeNodes and 1x Jetson TX2 (Edge nodes: vehicles)                              & Random Forest (RF); Gradient Boosting Decision Tree (GBDT), and Long Short-Term Memory Networks (LSTMs) & Collected from a large EV company                            \\
\cite{hong2019dlion}            & DLion, TensorFlow                                       & Model training                                       & training time and model accuracy                                                                                                              & 4x machines: 2x 24 CPUs; and 2x 8 CPUs                                                                                              & 2Conv + 2FC                                                                                             & CIFAR10                                                      \\
\rowcolor[HTML]{E0EBEA} 
\cite{nikouei2019toward}        & MXNet                                                   & Human-object tracking (model training)               & performance of human-object detection (performance in FPS, CPU usage, memory usage, Average False Positive Rate, Average False Negative Rate) & Raspberry Pi 3 B; and Tinker Board                                                                                                  & Lightweight Convolutional Neural Network (L-CNN)                                                        & Pascal Visual Object Classes (VOC) including VOC07 and VOC12 \\
\cite{sarabia2019efficient}     & TensorFlow, Keras                                       & Fall detection (inference)                           & model performance: inference time, accuracy, and precision; and Edge gateway: CPU, memory, and power consumption                              & Raspberry Pi 2 B                                                                                                                    & Long Short-Term Memory Units (LSTM), Gated Recurrent Unit (GRU), Support vector machines (SVM), and k-nearest neighbors (kNN)                                      & SisFall                                                      \\

\cline{1-7}
\end{tabular}
\end{sidewaystable*}

\begin{sidewaystable*}
\sffamily
\centering
\begin{tabular}{m{1.0cm}m{3.0cm}m{3.0cm}m{4.5cm}m{4.5cm}m{3.0cm}m{1.5cm}}
\hline
\rowcolor[HTML]{C4CDC1} 
\textbf{Paper}               & \textbf{Framework/Library}                            & \textbf{Application/Task}                                   & \textbf{Metrics}                                                                                                                                       & \textbf{Hardware}                                                                                                                   & \textbf{Model}                                                                                          & \textbf{Dataset}                                                      \\ \hline

\cite{kukreja2019training}      & Tensorflow, Caffe, PyTorch                              & Model training                                       & memory footprint                                                                                                                              & ODROID XU4                                                                                                                          & ResNet                                                                                                  & Waggle-based data                                            \\

\rowcolor[HTML]{E0EBEA} 
\cite{zhou2020saface} & InsightFace, Pytorch, MXNet                                            & face recognition                                       & accuracy, speedup of partial fine-tuning, and throughput                                                                                                           & Hisilicon Hi3516CV500 IP Cameras; 1x Intel i7-6700k CPU and Nvidia GTX1080 GPU                  & MobileNet, Sphere20, ResNet50                                                 & private dataset                          \\

\cite{ghosh2020edge}        & Edge-Cloud ML system                     & human activity recognition           & model accuracy, network consumption                                                                                                    & simulated edge: VM 2GB of memory and single core CPU; cloud: 1x 6GB of memory and octa-core processor                                   & Feed-Forward NN (FFNN)                                                                                          & MHEALTH Mobile Health                                                      \\

\rowcolor[HTML]{E0EBEA} 
\cite{mrozek2020fall}           & CoreML                                                  & Fall detection (model training)                      & model accuracy                                                                                                                                & Apple devices                                                                                                                       & Boosted Decisions Trees                                                                                 & SisFall                                                      \\

\cite{guo2021mistify}           & Mistify, TensorFlow                                     & DNN model porting (Computer Vision and Natural Language Processing) & adaptation time; convergence speed; accuracy; and resource usage                                                                              & 1x Linux server (NVIDIA 2070 GPU); 1x server (NVIDIA P600 GPU); Google Edge TPU; and Samsung S9 smartphone                          & MobileNet, ResNet50, ResNeXt101, BiDAF, and BERT                                                        & ImageNet, Cifar100, and SQuADv1.1 \\

\rowcolor[HTML]{E0EBEA} 
\cite{li2019edge}  &	Edgent, BranchyNet, Chainer	 & inference  & accuracy, latency requirements, inference throughput, network bandwidth	 & Raspberry Pi and desktop PC (quad-core 3.40 GHz Intel processor; 8 GB RAM)  &	AlexNet, CIFAR-10  & Belgium 4G/LTE bandwidth logs; Oboe \\

\cite{fafoutis2018extending}  & N.A.  & classification  & accuracy and battery lifetime  & SPHERE  & SVM	 & generated with SPHERE wearable \\

\rowcolor[HTML]{E0EBEA} 
\cite{kumar2017resource}  & N.A.  & inference & inference time, energy consumption, accuracy, model size  & Arduino	 & Bonsai, kNN, SVM and single hidden layer neural network & Chars4K; CIFAR10, MNIST,  WARD, USPS, Eye, RTWhale, CUReT \\

\cite{dey2019offloaded}  & N.A.  & inference  & processing latency; data transfer latency; network bandwidth  & Raspberry Pi 3; Intel Movidius Neural Compute Stick; smartphones; laptops; Nvidia Jetson Tx2  & Squeezenet, AlexNet, Inception-v3  & Not informed. \\

\cline{1-7}

\end{tabular}
\end{sidewaystable*}

\begin{sidewaystable*}
\sffamily
\centering
\caption{\textcolor{blue}{Quantitative analysis of artifacts, metrics, and hardware exploited in the Machine Learning experiments. Percentages refer to the total number of papers in that domain.}}
\label{tbl:ai-quantitative}
\begin{tabular}{>{}l>{}r>{}l>{}l>{}l>{}r>{}l>{}l>{}l>{}r>{}l>{}r}
\hline

\rowcolor[HTML]{C4CDC1} 
\multicolumn{2}{c}{\cellcolor[HTML]{C4CDC1}\textbf{Framework/Library}} & \multicolumn{2}{c}{\cellcolor[HTML]{C4CDC1}\textbf{Metrics}}             & \multicolumn{2}{c}{\cellcolor[HTML]{C4CDC1}\textbf{Hardware (Edge)}}   & \multicolumn{2}{c}{\cellcolor[HTML]{C4CDC1}\textbf{Processors}} & \multicolumn{2}{c}{\cellcolor[HTML]{C4CDC1}\textbf{Model}}          & \multicolumn{2}{c}{\cellcolor[HTML]{C4CDC1}\textbf{Dataset}}       \\ \hline
\rowcolor[HTML]{E0EBEA} 
TensorFlow                           & 28\%                            & model accuracy     &  \cellcolor[HTML]{E0EBEA}31\% & Emulated                         & 26\%                        & CPU      &  \cellcolor[HTML]{E0EBEA}77\%     & SVM                                & 15\%                        & CIFAR                             & 22\%                        \\
MXNet                                 & 10\%                             & resource usage     &  16\%                         &  Raspberry Pi                                & 22\%                        & GPU      &  18\%                             & SqueezeNet                                  & 9\%                        & SisFall                           & 11\%                        \\
\rowcolor[HTML]{E0EBEA} 
PyTorch                          & 10\%                             & inference time     &  \cellcolor[HTML]{E0EBEA}16\% & NVIDIA Jetson TX2                      & 13\%                        & TPU      &  \cellcolor[HTML]{E0EBEA}5\%      & AlexNet                               & 9\%                         &   MNIST                         & 11\%                         \\
Keras                                & 7\%                             & convergence speed  &  13\%                         & Arduino                                & 4\%                        &          &                                                      & VGG                                & 9\%                         & Vehicle                              & 6\%                         \\
\rowcolor[HTML]{E0EBEA} 
Scikit-Learn                              & 7\%                             & energy consumption &  \cellcolor[HTML]{E0EBEA}13\% & Intel FogNode                          & 4\%                         & \cellcolor[HTML]{FFFFFF}          & \cellcolor[HTML]{FFFFFF}                                                     & ResNet                             & 9\%                         & VOC                               & 6\%                         \\
\cellcolor[HTML]{FFFFFF}Caffe2       & 6\%                             & network            &  13\%                         & Nexus 6P                               & 4\%                         &          &                                                      & MobileNet                          & 9\%                         & RTWhale                               & 6\%                         \\
\rowcolor[HTML]{E0EBEA} 
TensorFlow Lite                      & 3\%                             & \cellcolor[HTML]{FFFFFF}      &  \cellcolor[HTML]{FFFFFF}  & Tinker Board                           & 4\%                         &    \cellcolor[HTML]{FFFFFF}      &   \cellcolor[HTML]{FFFFFF}                                                   & K-NN                               & 6\%                         & CUReT                     & 6\%                         \\
DeCaf                                & 3\%                             &                    &                                                  & ODROID XU4                             & 4\%                         &          &                                                      & LSTM                                & 6\%                         & ImageNet                          & 6\%                         \\
\cellcolor[HTML]{E0EBEA}MOCHA        & \cellcolor[HTML]{E0EBEA}3\%     &                    &                                                  & \cellcolor[HTML]{E0EBEA}Apple devices  & \cellcolor[HTML]{E0EBEA}4\% &          &                                                      & \cellcolor[HTML]{E0EBEA}RF & \cellcolor[HTML]{E0EBEA}6\% & \cellcolor[HTML]{E0EBEA}SQuADv1.1 & \cellcolor[HTML]{E0EBEA}6\% \\
CLONE                                & 3\%                             &                    &                                                  & Edge TPU                               & 4\%                         &          &                                                      & GRU                            & 3\%                         & Chars4K                           & 6\%                         \\
\cellcolor[HTML]{E0EBEA}DLion        & \cellcolor[HTML]{E0EBEA}3\%     &                    &                                                  & \cellcolor[HTML]{E0EBEA}Samsung S9     & \cellcolor[HTML]{E0EBEA}4\% &          &                                                      & \cellcolor[HTML]{E0EBEA}GBDT       & \cellcolor[HTML]{E0EBEA}3\% & \cellcolor[HTML]{E0EBEA}WARD      & \cellcolor[HTML]{E0EBEA}6\% \\
Chainer                                    & 3\%                             &                    &                                                  & SPHERE                                 & 4\%                         &          &                                                      & L-CNN                              & 3\%                         & USPS                              & 6\%                         \\
\cellcolor[HTML]{E0EBEA}CoreML       & \cellcolor[HTML]{E0EBEA}3\%     &                    &                                                  &                                        & \multicolumn{1}{l}{}        &          &                                                      & \cellcolor[HTML]{E0EBEA}GoogleNet  & \cellcolor[HTML]{E0EBEA}3\% & \cellcolor[HTML]{E0EBEA}Eye       & \cellcolor[HTML]{E0EBEA}6\% \\
Mistify                              & 3\%                             &                    &                                                  &                                        & \multicolumn{1}{l}{}        &          &                                                      & DT                                 & 3\%                         &                            &                          \\
\cellcolor[HTML]{E0EBEA}Edgent       & \cellcolor[HTML]{E0EBEA}3\%     &                    &                                                  &                                        & \multicolumn{1}{l}{}        &          &                                                      & \cellcolor[HTML]{E0EBEA}FFNN        & \cellcolor[HTML]{E0EBEA}3\% & \cellcolor[HTML]{FFFFFF}     & \cellcolor[HTML]{FFFFFF} \\
BranchyNet                           & 3\%                             &                    &                                                  &                                        & \multicolumn{1}{l}{}        &          &                                                      & BDT                                & 3\%                         &                                   & \multicolumn{1}{l}{}        \\
      &       &                    &                                                  &                                        & \multicolumn{1}{l}{}        &          &                                                      & \cellcolor[HTML]{E0EBEA}ANN        & \cellcolor[HTML]{E0EBEA}3\% &                                   & \multicolumn{1}{l}{}        \\
 \hline
\end{tabular}
\end{sidewaystable*}

\subsection{Main takeaways}

This section aims to answer the following research question: \textit{What are the main state-of-the-art methods for Machine Learning on the Edge-to-Cloud Continuum?} We organize the existing approaches and the selected studies in two main categories, they are: inference and distributed training on the Edge; and distributed training combining Edge and Cloud. 



\textcolor{blue}{We highlight that there is not a single library or framework that fits all the needs given the heterogeneous and complex nature of the Edge-to-Cloud Computing Continuum. Therefore, the idea is to provide scientists and engineers a clear vision of the main solutions and existing artifacts and resources (\emph{e.g.}, frameworks/libraries; application/task; metrics; hardware; models; and datasets) so that they can easily identify which ones may be exploited to better attend to their project and research needs.}

\textcolor{blue}{Next, we summarize the \textbf{main limitations of performing Machine Learning over Edge devices}: \emph{(i)} \textbf{computing power:} Edge devices are typically limited in terms of accelerator memory (CPU, GPU, TPU) and storage, thus they can not handle large ML/DL models. This can be alleviated either by minimizing the model size (while maintaining accuracy) or by distributing the model across devices~\cite{dey2019offloaded, zhou2019distributing, guo2021mistify, kumar2019decaf, lu2019collaborative, nikouei2019toward, kukreja2019training, hong2019dlion, chen2019exploring}; \emph{(ii)} \textbf{network communication:} Edge devices are typically interconnected through wireless low-bandwidth and unreliable network links. They may become offline at any time for any reason or the network may be congested. This requires fault-tolerant and communication-efficient approaches for distributed model training~\cite{li2019edge, zhou2019distributing, ghosh2020edge, mwase2022communication}; and \emph{(iii)} \textbf{energy consumption:} Edge devices are typically battery-powered, thus they can not handle energy-intensive ML/DL tasks. This should be addressed by energy-efficient inference and training techniques to extend the battery lifetime~\cite{fafoutis2018extending, kumar2017resource}.}

The articles presented will serve as a basis to identify the recent efforts and also how such libraries and frameworks are being used for: collaborative learning on the Edge and Fog; deploying Neural Networks and performing distributed Machine Learning tasks on resource-constrained devices; how these libraries and frameworks perform on Edge devices; and performance trade-offs for training on the Edge vs. on the Cloud.

\section{
Data Analytics Methods on the Edge-to-Cloud Continuum}
\label{sec:bda-cc}

\begin{figure}[t]
  \centering
  \includegraphics[width=0.8\linewidth]{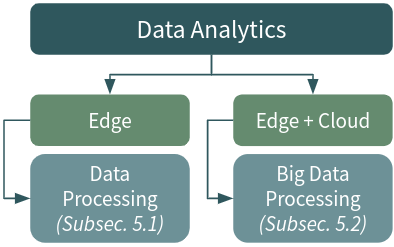}
  \caption{Taxonomy of analytics approaches on the Edge-to-Cloud Continuum.}
  \label{fig:taxonomy-analytics}
\end{figure}

Figure~\ref{fig:taxonomy-analytics} presents the taxonomy of analytics approaches with a focus on the Edge and Edge-to-Cloud Continuum. \textcolor{blue}{The Data Analytics frameworks identified in the articles are presented in Tables~\ref{tbl:da-cloud} (designed for the Cloud) and~\ref{tbl:da-edge} (designed for the Edge), respectively. Table~\ref{tbl:analytics} characterizes the selected articles with respect to resources exploited in the experimental evaluations, such as: frameworks; application/task; metrics; and hardware. Table~\ref{tbl:da-quantitative} presents quantitative analysis summarizing Table~\ref{tbl:analytics}.} In the next subsections, we present how these frameworks support analytics on the Edge-to-Cloud Continuum.

\subsection{Data Processing on the Edge}

\textcolor{blue}{Typically, IoT applications are latency-sensitive and they generate large amounts of data from sensors and Edge devices. Processing such data efficiently on the Edge of the network to obtain insights and react fast is critical. This section presents the recent efforts and novel systems proposed to distribute data processing among Edge devices to achieve high throughput and low latency.}

In~\cite{verma2020smart} the authors focus on using Edge computing for real-time analysis of healthcare systems. They discuss challenges of Edge computing such as: performance, deployment expertise in order to consider various parameters like infrastructure configurations, connectivity, and energy requirements; and data management.

A systematic study of data stream processing and analytics in the Fog considering four dimensions, such as system, data, human, and optimization is presented in~\cite{yang2017iot}. For each dimension, the authors present technical issues and new design challenges. For example, high throughput and low latency in stream processing systems can be achieved by optimizing their configurations such as: the number of bolts in the Storm DAG topology or the micro-batch size of Spark streaming, among others.

In the same direction of achieving high throughput and low latency, a novel stream processing engine focused on the Edge named EdgeWise is proposed in~\cite{fu2019edgewise}. The idea behind EdgeWise is the use of a congestion-aware scheduler and a fixed-size worker pool to improve throughput and latency. The authors compare EdgeWise with Storm deployed on a cluster of up to 8 Raspberry Pi nodes and they observe that EdgeWise reports up to 3 times improvement in throughput while keeping latency low.

\cite{dautov2017pushing}~proposes an approach to enable distributed data processing within a cluster of Edge devices. The proposed approach extends Apache NiFi core functionality to include three custom processors such as CaptureVideo, DetectFaces, and RecogniseFaces. Experiments show that the proposed approach has the potential to outperform the Cloud-enabled setup.

A novel distributed architecture that extends Apache NiFi to enable stream data processing at the Edge of the IoT network is presented in~\cite{dautov2020stream}. Edge Cluster Stream Processing (ECStream) allows time-constrained data-intensive applications to be entirely deployed and executed at the Edge and it is based on a task parallelism model where atomic tasks are offloaded to peer Edge devices, rather than the full workflow.

An Edge Intelligence framework for building service-oriented IoT is proposed in~\cite{huang2018building}. The framework allows developers to build stream processing capabilities on Edge server devices and use local streaming analytics to make IoT applications smart. Through annotation based programming primitives developers can design their local intelligent capabilities. The authors compare the latency of activity recognition engine implementations running on an Edge server and on the Cloud. Experiments show that the proposed framework can improve performance without degrading the recognition accuracy.

A new Fog platform for data stream analytics in IoT is proposed in~\cite{alencar2020fot}. It aims to exploit the computational capacity of Fog devices to process and analyze data without requiring a frequent use of Cloud resources. Experimental evaluations show that the proposed system can analyze data streams with low processing delay and low network utilization.

In~\cite{hauswirth2020autonomous} the authors propose Fed4Edge, a system that enables the coordination of resources available in Edge devices to process query pipelines in a collaborative way. Fed4Edge uses RDF Stream Processing (RSP) engines as autonomous processing agents. Large scale evaluations on a cluster of Raspberry Pi show that the scalability can be significantly improved by adding more Edge devices to a network of processing nodes.

A model synchronization mechanism for distributed and stateful data analytics named SCEDA is proposed in~\cite{aral2020staleness}. The authors use Reinforcement Learning to make dynamic scheduling decisions by learning individual network connectivity trends of Edge nodes as well as the significance of their updates. The proposed approach tackles the concept drift and connectivity issues in Edge data analytics to minimize its accuracy handicap without losing its timeliness benefits. Experimental results show that SCEDA can achieve a comparable level of accuracy as core data analytics.


\textcolor{blue}{In summary, the articles presented exploit the computational capacity of Edge devices to process and analyze data in a distributed way. The proposed approaches contribute to allow data-intensive and latency-sensitive applications to be entirely processed on Edge devices.}

\subsection{Big Data Processing across the Edge-to-Cloud}

\textcolor{blue}{The articles presented in this section focus on novel architectures and frameworks exploiting collaborative Edge-to-Cloud data processing for enabling real-time data analytics. The articles also aim to analyze the performance trade-offs of Cloud only \emph{vs.} Edge-to-Cloud collaborative data analytics.}

In~\cite{dautov2018data}, the authors propose a novel IoT distributed Stream Processing architecture that distributes the workload among a cluster of Edge devices. The proposed solution extends Apache NiFi with new services to discover and select devices able to perform offloaded tasks according to hardware and software requirements. The evaluation scenario consists of an intelligent surveillance system and the authors compare the performance of a cluster of Edge devices with a Cloud setup. Results show that an Edge cluster of 6 nodes performs up to 5-6 times faster than the Cloud deployments.

Later, the same authors proposed~\cite{dautov2019hierarchical} a distributed hierarchical data fusion architecture based on Complex Event Processing technology to handle streaming data. This approach produces timely and accurate results with minimum time delay, as soon as necessary information is generated and collected. The authors compare their solution (distributed hierarchical data fusion) with a traditional one (sending all low-level sensor readings to a central Cloud for analysis) and evaluations show that at lower levels (\emph{e.g.}, Edge and Fog) decisions can be taken 20x and 90x times faster.

In~\cite{sharma2017live}, the authors propose a novel framework of collaborative Edge-Cloud processing for enabling live data analytics in wireless IoT networks. They also present potential key enablers for the proposed framework and highlight some of the research directions for Big Data aware collaborative Edge-Cloud processing, such as adaptive learning/prediction algorithms. 

\textcolor{blue}{In~\cite{rosendo2020e2clab}, the authors propose a novel framework that allows the deployment and optimization~\cite{rosendo2021reproducible} of Big Data Analytics applications on the Edge-to-Cloud continuum. They illustrate and validate the framework with a smart surveillance application composed by data processing frameworks such as Edgent (on the Edge) and Apache Flink and Kafka (on the Cloud). Experimental evaluations exploit the performance trade-offs of Cloud-centric vs. hybrid Edge-Cloud processing approaches to understand how they impact metrics such as latency and throughput of the application.}

~\cite{xu2021move} proposes a novel framework that uses fine-grained stream processing to provide high resource utilization while meeting latency targets. Named Cameo, the framework dynamically calculates and propagates priorities of events based on user latency targets. Experiments show that Cameo reduces query latency in single and multi-tenant settings.

Several models, technologies and solutions for medical data processing and analysis are presented in~\cite{kolodziej2019high}. The authors illustrate examples of case studies and practical solutions composed of health sensor data processed with Kafka and Spark (an application predicting skin temperature based on heart rate and step count values) using medical datasets publicly available such as PhysioNet, UbiqLog and CrowdSignals.

In~\cite{verma2017survey}, the authors review the state of the art of the analytics network methodologies for real-time IoT analytics. They also present some real-time IoT analytics use cases and software platforms such as Flink, Spark, Storm, and Druid along with their network requirements. Lastly, they present research problems and future research directions focusing on the network methodologies for the real-time IoT analytics.


\textcolor{blue}{In summary, the articles demonstrate the benefits of collaborative Edge-to-Cloud data analytics. The main performance improvements highlighted by these studies regarding the comparison of Edge-to-Cloud \emph{vs.} Cloud only approaches refer to minimizing the processing latency of applications.}

\begin{sidewaystable*}
\centering
\sffamily
\caption{\textcolor{blue}{Summary of artifacts, metrics, and hardware exploited in the Data Analytics experiments.}}
\label{tbl:analytics}
\begin{tabular}{m{1.0cm}m{4.0cm}m{4.0cm}m{5.0cm}m{8.0cm}}
\hline
\rowcolor[HTML]{C4CDC1} 
\textbf{Paper}               & \textbf{Framework/Library} & \textbf{Application/Task}                                    & \textbf{Metrics}                                                                                                                                            & \textbf{Hardware}                                                                                                                                                    \\ \hline
\rowcolor[HTML]{E0EBEA} 
\cite{dautov2018data}           & NiFi                         & Surveillance System (collaborative data processing)   & processing time delay (network latency + data serialization + queueing)                                                                            & Edge setup: 8-Megapixel camera acted as the CCTV source; 3x Raspberry Pi 3; 3x Google Nexus 4. Cloud setup: Heroku (Intel Xeon CPU 2.5GHz, 512MB RAM) and Amazon EC2 (Intel Xeon CPU 2.4GHz, 4GB RAM)                         \\

\cite{dautov2019hierarchical}   & Flink                        & Smart Healthcare (data stream processing)             & time delay: difference between the moment when sensor data are first generated and the moment when valuable insights are drawn based on these data & Raspberry Pi 3; 1x machine 2.00 GHz Intel Core i7-4510U CPU, 16 GB RAM; and Google Compute Engine n1-standard-1;                                                                                  \\

\rowcolor[HTML]{E0EBEA} 
\cite{dautov2017pushing}        & Nifi                         & Face detection and recognition                        & processing time delay (image capture + face recognition + return results to the coordinator)                                                       & 5x VirtualBox instances of Raspberry Pi; Heroku; and Amazon Elastic Beanstalk;                                                                                                               \\


\cite{dautov2020stream}         & Nifi                         & Stream processing                                     & response time (time difference between an image is first sampled and the recognition task)                                                         & Raspberry Pi 3; Samsung Galaxy J5; Amazon EC2                                                                                                                                               \\
\rowcolor[HTML]{E0EBEA} 
\cite{fu2019edgewise}           & EdgeWise, Storm              & Stream processing                                     & throughput-latency performance                                                                                                                     & 8x Raspberry Pi 3 B; and 1x desktop machine                                                                                                                                 \\

\cite{alencar2020fot}           & Edgent, FoT-Stream, Kafka    & Stream processing                                     & processing delay and network utilization                                                                                                           & Sonoff SC; Raspberry Pi 3 B+; and Amazon EC2 servers                                                                                                       \\
\rowcolor[HTML]{E0EBEA} 
\cite{xu2021move}               & Cameo, Flink                 & Stream processing                                     & query latency                                                                                                                                      & DS12-v2 and DS11-v2 Azure virtual machines                                                                                                                                           \\

\cite{hauswirth2020autonomous}  & Fed4Edge                     & Stream processing                                     & query processing throughput                                                                                                                        & 85x Raspberry Pi B                                            \\
\rowcolor[HTML]{E0EBEA} 
\cite{huang2018building}        & Flink, Kafka                 & Activity recognition (random forest classifier model) & classification time; processing time; end to end latency; and activity recognition accuracy                                                        & Edge: Raspberry Pi 3; Cloud: Intel Xeon CPU E5-2640 v3 (2.6 GHz)  \\ 
\cite{rosendo2020e2clab}  & Flink, Kafka, Edgent                     & video stream processing                                     & end-to-end latency, processing throughput                                                                                                                        & 35x machines of the Grid5000 testbed                                            \\

\cline{1-5}

\end{tabular}
\end{sidewaystable*}

\begin{table*}[t]
\scriptsize
\sffamily
\centering
\caption{\textcolor{blue}{Quantitative analysis of artifacts, metrics, and hardware exploited in the Data Analytics experiments. Percentages refer to the total number of papers in that domain.}}
\label{tbl:da-quantitative}
\resizebox{\linewidth}{!}{
\begin{tabular}{>{}l>{}r>{}l>{}r>{}l>{}r>{}l>{}r}
\hline
\rowcolor[HTML]{C4CDC1} 
\multicolumn{2}{c}{\cellcolor[HTML]{C4CDC1}\textbf{Framework/Library}} & \multicolumn{2}{c}{\cellcolor[HTML]{C4CDC1}\textbf{Metrics}}            & \multicolumn{2}{c}{\cellcolor[HTML]{C4CDC1}\textbf{Hardware (Edge)}} & \multicolumn{2}{c}{\cellcolor[HTML]{C4CDC1}\textbf{Processors}}              \\ \hline
\rowcolor[HTML]{E0EBEA} 
Kafka                                & 19\%                            & end to end latency       & 47\%                                         & Raspberry Pi                           & 53\%                        & CPU                      & \cellcolor[HTML]{E0EBEA}100\% \\
Flink                                & 15\%                            & network                  & 16\%                                         & Emulated                               & 18\%                        & GPU                      & 0\%                           \\
\rowcolor[HTML]{E0EBEA} 
NiFi                                 & 11\%                            & model accuracy      & 16\%                                         & Arduino                                & 12\%                        & TPU                      & \cellcolor[HTML]{E0EBEA}0\%   \\
Storm                                & 11\%                            & throughput               & 16\%                                         & \cellcolor[HTML]{FFFFFF}Google Nexus 4 & \cellcolor[HTML]{FFFFFF}6\% & \cellcolor[HTML]{FFFFFF} & \cellcolor[HTML]{FFFFFF}                          \\
\rowcolor[HTML]{E0EBEA} 
Hadoop                               & 7\%                             & classification time      & 5\%                                          & Samsung Galaxy J5                      & 6\%                         & \cellcolor[HTML]{FFFFFF} & \cellcolor[HTML]{FFFFFF}                          \\
\cellcolor[HTML]{FFFFFF}Spark        & 7\%                             &            &                                           & \cellcolor[HTML]{FFFFFF}Sonoff SC      & \cellcolor[HTML]{FFFFFF}6\% & \cellcolor[HTML]{FFFFFF} & \cellcolor[HTML]{FFFFFF}                          \\
\rowcolor[HTML]{FFFFFF} 
\cellcolor[HTML]{E0EBEA}Edgent       & \cellcolor[HTML]{E0EBEA}4\%     &                          &                                              &                                        &                             &                          &                                                   \\
Flume                                & 4\%                             & \cellcolor[HTML]{FFFFFF} & \multicolumn{1}{l}{\cellcolor[HTML]{FFFFFF}} & \cellcolor[HTML]{FFFFFF}               & \cellcolor[HTML]{FFFFFF}    &                          &                                                   \\
\cellcolor[HTML]{E0EBEA}EdgeWise     & \cellcolor[HTML]{E0EBEA}4\%     & \cellcolor[HTML]{FFFFFF} & \multicolumn{1}{l}{\cellcolor[HTML]{FFFFFF}} & \cellcolor[HTML]{FFFFFF}               & \cellcolor[HTML]{FFFFFF}    &                          &                                                   \\
FoT-Stream                           & 4\%                             & \cellcolor[HTML]{FFFFFF} & \multicolumn{1}{l}{\cellcolor[HTML]{FFFFFF}} & \cellcolor[HTML]{FFFFFF}               & \cellcolor[HTML]{FFFFFF}    &                          &                                                   \\
\cellcolor[HTML]{E0EBEA}Cameo        & \cellcolor[HTML]{E0EBEA}4\%     & \cellcolor[HTML]{FFFFFF} & \multicolumn{1}{l}{\cellcolor[HTML]{FFFFFF}} & \cellcolor[HTML]{FFFFFF}               & \cellcolor[HTML]{FFFFFF}    &                          &                                                   \\
Fed4Edge                             & 4\%                             &                          & \multicolumn{1}{l}{}                         & \cellcolor[HTML]{FFFFFF}               & \cellcolor[HTML]{FFFFFF}    &                          &                                                   \\
\cellcolor[HTML]{E0EBEA}MOA          & \cellcolor[HTML]{E0EBEA}4\%     &                          & \multicolumn{1}{l}{}                         &                                        & \multicolumn{1}{l}{}        &                          &                                                   \\ \hline
\end{tabular}
}
\end{table*}

\subsection{Main takeaways}

This section aims to answer the following research question: \textit{What are the main state-of-the-art methods for data analytics on the Edge-to-Cloud Continuum?} We organize the existing frameworks and the selected studies in two main categories, they are: data processing on the Edge; and Edge-to-Cloud Big Data processing. 



From the analysis of the selected articles, we observe that, compared to the Cloud, a few stream processing frameworks tailored for the Edge exist, such as Apache NiFi, Edgent, and EdgeWise. We highlight that the recent works focus on proposing novel approaches for collaborative Edge-Cloud processing in order to enable live data analytics, instead of focusing on novel processing frameworks designed for running just on the Edge.

\section{
Approaches to Combine Machine Learning and Data Analytics 
}
\label{sec:ai-bda}

The convergence of Big Data and AI has become a research trend that grows over the years given the benefits of combining and applying both technologies in many areas, such as self-driving vehicles~\cite{grzywaczewski2017training}, precision agriculture~\cite{bhat2021big}, smart manufacturing~\cite{lee2020industrial}, among others. Combining Big Data and AI leverages advanced analytics capabilities and allows the efficient extraction of valuable insights from vast amounts of data~\cite{duan2019artificial}.

\begin{figure}[t]
  \centering
  \includegraphics[width=\linewidth]{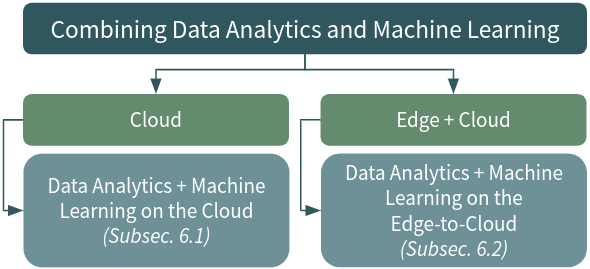}
  \caption{Taxonomy of approaches combining data analytics and learning on the Edge-to-Cloud Continuum.}
  \label{fig:taxonomy-learning-analytics}
\end{figure}

Next, we present the recent efforts on combining Big Data and AI to enable hybrid Cloud and Edge analytics. Figure~\ref{fig:taxonomy-learning-analytics} presents the taxonomy of approaches combining Machine Learning and Data Analytics with a focus on the Edge-to-Cloud Continuum.  \textcolor{blue}{The ML and Data Analytics frameworks/libraries identified in the articles are presented in Tables~\ref{tbl:ml-cloud} and~\ref{tbl:ml-edge} and Tables~\ref{tbl:da-cloud} and~\ref{tbl:da-edge}, respectively. Table~\ref{tbl:analytics-and-learning} characterizes the selected articles regarding resources exploited in the experimental evaluations, such as: frameworks/libraries; application/task; metrics; and hardware; models; and datasets. Table~\ref{tbl:learning-paradigms} presents the main state-of-the-art learning paradigms for collaborative learning. Table~\ref{tbl:ai-da-quantitative} presents a quantitative analysis summarizing Table~\ref{tbl:analytics-and-learning}.}

\subsection{Combining Data Analytics and Learning on the Cloud}

\textcolor{blue}{The articles presented in this section explore the joint usage of Data Analytics and Machine Learning frameworks and algorithms for analytics on Cloud resources. They also discuss the relevance of learning paradigms (\emph{e.g.}, Deep Learning, Online Learning, and Transfer Learning, among others) for Big Data Analytics.}

Distributed Cloud-based Machine Learning tools such as Mahout, Spark MLlib, and FlinkML are presented in~\cite{rao2019big}. Authors also present research directions and opportunities in the domain of developing parallel and distributed Machine Learning algorithms. For instance, they highlight that in streaming systems, there is a lack of online Machine Learning algorithms that are used to process real-time data to provide faster insights.

\textcolor{blue}{The survey~\cite{nguyen2019machine} presents ML and DL frameworks and libraries oriented towards fast processing and streaming of large-scale data, such as: TensorFlow, PyTorch, MXNet, Theano, FlinkML, and Spark MLlib. Authors highlight that there is no single tool suitable for every problem and often a combination of them is needed to succeed.}

\textcolor{blue}{Authors of~\cite{pathak2018construing} present key characteristics and challenges of handling Big Data. Regarding current trends in Big Data Analytics they refer to IoT and Edge analytics (to provide responses quickly as the events occur) and domain adaptation (where training data and test data are sampled from different distributions).}

In~\cite{ali2018recent} the authors survey existing solutions for Big Data stream processing in terms of learning type, supported languages, and supported Machine Learning tools. Authors discuss frameworks and platforms such as Apache Spark, MOA, Samza, Storm, and Kafka.

Another survey~\cite{ulusar2020open} presents existing open source tools for Big Data (\emph{e.g.}, Hadoop, Spark, Storm, and Flink) and Machine Learning (\emph{e.g.}, Mahout, Spark MLLib, and SAMOA). Besides, authors review Machine Learning algorithms in Big Data such as Supervised, Unsupervised, and Semi-supervised learning.

In~\cite{l2017machine} the authors present the challenges associated with Machine Learning in the context of Big Data and categorize them according to the Velocity, Volume, Variety, and Veracity dimensions of Big Data. They also present existing Machine Learning approaches and techniques for data manipulation (\emph{e.g.}, dimensionality reduction and data cleaning); processing manipulation (such as vertical/horizontal scaling and batch/stream oriented); and algorithm manipulation (\emph{e.g.}, algorithm modification with new paradigms). Lastly, they present learning paradigms relevant to Big Data such as Deep Learning, Lifelong Learning, Online Learning, and Transfer Learning.

\textcolor{blue}{A parallel Machine Learning algorithm for fault classification of mobile robotic roller bearings is proposed in~\cite{xian2020parallel}. The proposed algorithm combines Support Vector Machine with Spark to realize parallel operations. Experimental evaluations under different training set sizes demonstrate that the proposed algorithm (Spark SVM on Mesos) outperforms MapReduce SVM, Storm SVM, Radial Basis Function Neural Network (RBFNN), Deep Belief Network (DBN), presenting higher: classification accuracy, processing speed, and convergence rate.}

\textcolor{blue}{In~\cite{assefi2017big} the authors explore Spark MLlib with a variety of Big Data Machine Learning experiments on massive datasets to understand the qualitative and quantitative attributes of the platform. Experimental evaluations compare the performance of classification and clustering models (such as SVM, Decision Tree, Naive Bayes, Random Forest, and K-Means) on a variety of hardware and software configurations. Results show that with large datasets Spark MLlib outperforms Weka in terms of running time.}

\textcolor{blue}{In~\cite{nair2018applying} the authors propose a system for real-time health status prediction based on the Spark Big Data processing framework. The proposed system predicts user’s health status by applying a variety of decision tree models on streams of data. Experiments evaluate the generalization error of Decision Tree models based on maxDepth (tree depth) and maxBins (ordered splits) parameter values.}

\textcolor{blue}{In~\cite{ali2020res} the authors propose an Edge-Cloudlet-MultiResource three-tier architecture  to enable real-time processing of video streams. The proposed system performs Deep Learning inference on Cloudlets and distributes processing stages on the available resources using an algorithm to satisfy user Quality of Service requirements. Results show that for a 10K element data streams, with a frame rate of 15-100 per second, the job completion in the proposed system takes 49\% less time and saves 99\% bandwidth compared to a centralized Cloud-only based approach.}

An overview on Spiking Neural Networks (SNN) for Online Learning scenarios is presented in~\cite{lobo2020spiking}. According to the authors, the use of SNN in Online Learning allows fast real-time simulations of large networks and a low computational cost. SNN make possible the accumulation of knowledge as data become available without the requirement of storing and retraining the model with past samples. The authors also highlight research trends in the field of SNN and Online Learning such as Lifelong Machine Learning and Deep SNN Learning.




\begin{table*}[t]
\sffamily
\small
\centering
\caption{\textcolor{blue}{Quantitative analysis of Learning Paradigms for learning on the Edge-to-Cloud Continuum. Percentages refer to all papers reviewed which are discussing or exploiting in their experiments these learning paradigms.}}
\label{tbl:learning-paradigms}
\begin{tabular}{m{2.0cm}m{1.0cm}m{2.2cm}m{2.8cm}m{7.2cm}}
\hline
\rowcolor[HTML]{c4cdc1}
\textbf{Paper} & \textbf{Pct.} & \textbf{Learning Paradigm}         & \textbf{Main ideas}                                                                                     & \textbf{Characteristics}                                                                                                                                                                                                                                                                                                                                                                                                                                                                                                                                                                                                                   \\ \hline
\rowcolor[HTML]{E0EBEA} 
\cite{zhou2019distributing, kumar2019decaf, chen2019exploring, hong2019dlion, sankaranarayanan2020data, hong2019dlion, wang2018iot, l2017machine, sezer2017context, mohammadi2018deep, sharma2017live, liu2019survey, zhang2019deep, prabhu2019fog, khayyam2020artificial, xu2018survey, lobo2020spiking, yousefpour2019all, fei2019cps, aral2020staleness, paakkonen2020extending, rocha2020distributed} & 30\% & Distributed Machine/ Deep Learning~\cite{sergeev2018horovod}   & Exploit distributed resources of a cluster to speed up the convergence of model training.     & 

\begin{itemize}
    \item aims at parallelizing computing power.
    \item simultaneously places and processes data for model training and testing into a number of distributed nodes.
\end{itemize}
\\

\cite{grzenda2020hybrid, hong2019dlion, l2017machine, mohammadi2018deep, sharma2017live, zhang2019deep, rao2019big, ulusar2020open, nguyen2019machine, khayyam2020artificial, xu2018survey, lobo2020spiking, grzenda2020hybrid, diez2019data, fei2019cps, zhou2020saface} & 21\% & Online Learning or Sequential Learning~\cite{liang2006fast}    & Allow updating models upon arrival of new data without the need to retrain the complete model. & 

\begin{itemize}
    \item models learn one instance at a time: does not rebuild the model every time new data arrives, instead, it updates the existing knowledge based on the new incoming data.
    \item models are frequently updated: dynamic updates of the trained models are determining factors for a reliable and efficient analytic module.
    \item uses data streams for model training: can accommodate bigger datasets than batch learning and matches the need for stream analysis of IoT data. Besides, it can be efficiently deployed in an Edge device as they do not accumulate large amount of data.
\end{itemize}
 \\

\rowcolor[HTML]{E0EBEA} 
\cite{wang2018iot, sezer2017context, mohammadi2018deep, zhang2019deep, khayyam2020artificial, lobo2020spiking, fei2019cps, aral2020staleness, rocha2020distributed, chen2019deep, rong2021edge}  & 16\%  & Reinforcement Learning~\cite{sutton2018reinforcement}                                                               & Learning is based on feedback obtained from interactive actions in the environment.            & 

\begin{itemize}
    \item does not require a training data set: it interacts with the external environment to continuously adapt and learn on given points as a kind of feedback.
\end{itemize}
\\

\cite{kumar2019decaf, talagala2018eco, l2017machine, mohammadi2018deep, zhang2019deep, pathak2018construing, khayyam2020artificial, diez2019data, loghin2020disruptions, chen2019deep} & 14\% & Transfer Learning~\cite{pan2009survey} and Multi-task Learning~\cite{zhang2017survey} & Exploit the knowledge obtained from a task to improve generalization about another.           & 

\begin{itemize}
    \item useful when you have lack of training data sets.
    \item models may be reused as a starting point for predicting in another, but related, domain.
    \item compared to separated model training, it improves learning efficiency and prediction accuracy for the task-specific models.
\end{itemize}
\\

\rowcolor[HTML]{E0EBEA} 
\cite{lu2019collaborative, kumar2019decaf, talagala2018eco, mohammadi2018deep, zhang2019deep, diez2019data, aral2020staleness, loghin2020disruptions, chen2019deep, rong2021edge} & 14\%  & Federated Learning~\cite{bonawitz2019towards}                                                                   & Train a centralized model in a distributed way without the need to share private data.         & 

\begin{itemize}
    \item aims at training on heterogeneous datasets.
    \item data privacy: transmits and aggregates trained model parameters instead of training data.
    \item enables scalability: leverages the concurrent use of a high number of Edge devices to be independently trained and periodically synchronized through a central parameter server.
\end{itemize}
\\

\cite{l2017machine, zhang2019deep, lobo2020spiking, fei2019cps} & 5\%  & Lifelong Learning or Continual Learning~\cite{shin2017continual}   & Learning is continuous and knowledge is retained and used to solve different problems.         & 

\begin{itemize}
    \item does not rebuild the knowledge model every time a new piece of data arrives, but only updates the existing knowledge with the new incoming data.
    \item can accommodate bigger datasets than batch learning.
    \item may be a promising solution for lack of data, real-time processing, and concept drift.
\end{itemize}
                                                                                                                    \\ \hline

\end{tabular}
\end{table*}

\begin{sidewaystable*}
\sffamily
\centering
\caption{\textcolor{blue}{Summary of articles combining Data Analytics and Machine Learning in their experimental evaluations.}}
\label{tbl:analytics-and-learning}
\begin{tabular}{m{1.0cm}m{3.0cm}m{3.0cm}m{4.0cm}m{4.0cm}m{3.0cm}m{3.0cm}}
\hline
\rowcolor[HTML]{C4CDC1} 
\textbf{Paper}               & \textbf{Framework/Library} & \textbf{Application/Task}                                    & \textbf{Metrics}                                                                                                                                            & \textbf{Hardware} & \textbf{Model} & \textbf{Dataset}                                                                                                                                                    \\ \hline
\rowcolor[HTML]{E0EBEA} 
\cite{wang2018iot}           & 5G I-IoT framework                         & 5G channel utilization   & latency, spectrum efficiency, energy efficiency, data rates, reliability, QoS, and load                                                                            & simulation & NA  & NA                         \\


\cite{grzenda2020hybrid}        & Flink, Flume, Kafka, MOA          & Processing of vehicle location data                   & prediction accuracy                                                                                                                                & Not informed.  & Combined Stream and Neural Network (CSaNN), Combined Stream and Random Forest (CSaRF)  & Warsaw tram data (WAW)                                                                                                                                                               \\
\rowcolor[HTML]{E0EBEA} 
\cite{perez2018resilient}       & R                                                       & Traffic forecasting (model training)                 & amount of data collected by the Fog Nodes; impact of network between Fog Nodes and Cloud; model accuracy Edge vs Cloud;                       & Cluster of 8 servers: 2x Xeon E5-2630v4 (broadwell) and 128 GB of DDR4-2400 R ECCRAM. Edge devices: RaspberryPi 3 B                 & Conditional Restricted Boltzmann Machines (CRBM)                                                        & Floating Car Data (FCD)                                      \\

\cite{sankaranarayanan2020data} & Scikit-learn, Storm, Nifi, Hadoop, Kafka   & Distributed model training                            & network latency and service latency                                                                                                               & IoT device: 10x Arduino; Edge device: 4x Raspberry Pi 3; Edge gateway: 2x Intel E5645@2.4 GHz and memory of 24 GB   & Data Flow and Distributed Deep Neural Network (DF-DDNN) & IoT truck simulator by Horton works                                                                                 \\

\rowcolor[HTML]{E0EBEA} 
\cite{sarabia2020highly}        & Tensorflow, Keras, Scikit-Learn, Spark, Kafka                 & Fall detection system (model training)                & model accuracy, sensitivity, and precision                                                                                                         & Edge: Arduino Uno; Fog: Raspberry Pi 2 B; Cloud: 5x Amazon EC2 and 2x T2.micro; and 1x T2.medium   & RNN (LSTM/GRU) & SisFall                                                   \\

\cite{assefi2017big} & Spark, Weka   & inference                            & running time                                                                                                               & 2x virtual machines with 8 GB 4 vCPUs; and 16 GB 8 vCPUs   & SVM, Decision Tree, Naive Bayes, Random Forest, and K-Means & HEPMASS, SUSY, HIGGS, FLIGHT, HETROACT                                                                                 \\

\rowcolor[HTML]{E0EBEA} 
\cite{nair2018applying}        & Spark                  & health status prediction                & generalization error                                                                                                         & 1x Intel i5 and 8GB RAM; and 3x Amazon EC2   & Decision tree & heart disease                                                   \\

\cite{xian2020parallel}        & Spark, Storm, and Hadoop & fault classification                & classification accuracy, classification time, convergence rate                                                                                                         & 18 virtual machines on 6 physical machines   & SVM, Radial Basis Function Neural Network (RBFNN), Deep Belief Network (DBN) & fault pattern                                                   \\

\rowcolor[HTML]{E0EBEA} 
\cite{ali2020res}               & TensorFlow, Spark, Kafka, Storm, Hadoop                                              & Video stream analytics (deep learning inference)                    & job execution time; and bandwidth consumed                                                                                                    & Xeon E5 system with 8 cores and Nvidia GTX 970 GPU                                                                                  & MobileNet                  & Tiny ImageNet; and Stanford vehicle dataset and CASIA-Webface (face recognition) \\

\cline{1-7}

\end{tabular}
\end{sidewaystable*}

\begin{sidewaystable*}
\sffamily
\centering
\caption{\textcolor{blue}{Quantitative analysis of artifacts, metrics, and hardware exploited in the experiments combining Machine Learning and Data Analytics. Percentages refer to the total number of papers in that domain.}}
\label{tbl:ai-da-quantitative}
\begin{tabular}{>{}l>{}r>{}l>{}r>{}l>{}r>{}l>{}l>{}l>{}r>{}l>{}r}
\hline
\rowcolor[HTML]{C4CDC1} 
\multicolumn{2}{c}{\cellcolor[HTML]{C4CDC1}\textbf{Framework/Library}} & \multicolumn{2}{c}{\cellcolor[HTML]{C4CDC1}\textbf{Metrics}}                                    & \multicolumn{2}{c}{\cellcolor[HTML]{C4CDC1}\textbf{Hardware (Edge)}} & \multicolumn{2}{c}{\cellcolor[HTML]{C4CDC1}\textbf{Processors}}              & \multicolumn{2}{c}{\cellcolor[HTML]{C4CDC1}\textbf{Model}}          & \multicolumn{2}{c}{\cellcolor[HTML]{C4CDC1}\textbf{Dataset}}            \\ \hline
\rowcolor[HTML]{E0EBEA} 
Spark                                 & 19\%                           & network                                      &  \cellcolor[HTML]{E0EBEA}24\% & Emulated                          & 28\%                             & CPU                      &  \cellcolor[HTML]{E0EBEA}100\% & SVM                                   & 14\%                        & WAW                                      & 10\%                         \\
Kafka                                 & 15\%                           & model accuracy                               &  19\%                         & RaspberryPi                       & 21\%                             & GPU                      &  0\%                           & CSaNN                                 & 7\%                         & FCD                                      & 10\%                         \\
\rowcolor[HTML]{E0EBEA} 
Storm                                 & 11\%                           & processing time                              &  \cellcolor[HTML]{E0EBEA}14\% & Arduino                           & 10\%                             & TPU                      &  \cellcolor[HTML]{E0EBEA}0\%   & CSaRF                                 & 7\%                         & SisFall                                  & 10\%                         \\
Hadoop                                & 11\%                           & spectrum efficiency                          &  5\%                          & \cellcolor[HTML]{FFFFFF}          & \cellcolor[HTML]{FFFFFF}         & \cellcolor[HTML]{FFFFFF} & \cellcolor[HTML]{FFFFFF}                          & DF-DDNN                               & 7\%                         & HEPMASS                                  & 10\%                         \\
\rowcolor[HTML]{E0EBEA} 
Scikit-learn                          & 7\%                            & energy efficiency                            &  \cellcolor[HTML]{E0EBEA}5\%  & \cellcolor[HTML]{FFFFFF}          & \cellcolor[HTML]{FFFFFF}         & \cellcolor[HTML]{FFFFFF} & \cellcolor[HTML]{FFFFFF}                          & LSTM                                  & 7\%                         & SUSY                                     & 10\%                         \\
\cellcolor[HTML]{FFFFFF}Tensorflow    & 7\%                            & data rates                                   &  5\%                          & \cellcolor[HTML]{FFFFFF}          & \cellcolor[HTML]{FFFFFF}         & \cellcolor[HTML]{FFFFFF} & \cellcolor[HTML]{FFFFFF}                          & GRU                                   & 7\%                         & HIGGS                                    & 10\%                         \\
\rowcolor[HTML]{E0EBEA} 
5G I-IoT                              & 4\%                            & reliability                                  &  \cellcolor[HTML]{E0EBEA}5\%  & \cellcolor[HTML]{FFFFFF}          & \cellcolor[HTML]{FFFFFF}         & \cellcolor[HTML]{FFFFFF} & \cellcolor[HTML]{FFFFFF}                          & kNN                                   & 7\%                         & FLIGHT                                   & 10\%                         \\
Flink                                 & 4\%                            & QoS                                          & 5\%                                              & \cellcolor[HTML]{FFFFFF}          & \cellcolor[HTML]{FFFFFF}         &                          &                                                   & MobileNet                             & 7\%                         & HETROACT                                 & 10\%                         \\
\cellcolor[HTML]{E0EBEA}Flume         & \cellcolor[HTML]{E0EBEA}4\%    & \cellcolor[HTML]{E0EBEA}load                 & \cellcolor[HTML]{E0EBEA}5\%                      & \cellcolor[HTML]{FFFFFF}          & \cellcolor[HTML]{FFFFFF}         &                          &                                                   & \cellcolor[HTML]{E0EBEA}Decision Tree & \cellcolor[HTML]{E0EBEA}7\% & \cellcolor[HTML]{E0EBEA}Stanford vehicle & \cellcolor[HTML]{E0EBEA}10\% \\
MOA                                   & 4\%                            & amount of data collected                     & 5\%                                              & \cellcolor[HTML]{FFFFFF}          & \cellcolor[HTML]{FFFFFF}         &                          &                                                   & Naive Bayes                           & 7\%                         & CASIA-Webface                            & 10\%                         \\
\cellcolor[HTML]{E0EBEA}R             & \cellcolor[HTML]{E0EBEA}4\%    & \cellcolor[HTML]{E0EBEA}generalization error & \cellcolor[HTML]{E0EBEA}5\%                      & \cellcolor[HTML]{FFFFFF}          & \cellcolor[HTML]{FFFFFF}         &                          &                                                   & \cellcolor[HTML]{E0EBEA}Random Forest & \cellcolor[HTML]{E0EBEA}7\% & \cellcolor[HTML]{FFFFFF}                 & \cellcolor[HTML]{FFFFFF}     \\
Nifi                                  & 4\%                            & convergence rate                             & 5\%                                              & \cellcolor[HTML]{FFFFFF}          & \cellcolor[HTML]{FFFFFF}         &                          &                                                   & K-Means                               & 7\%                         & \cellcolor[HTML]{FFFFFF}                 & \cellcolor[HTML]{FFFFFF}     \\
\cellcolor[HTML]{E0EBEA}Keras         & \cellcolor[HTML]{E0EBEA}4\%    &                                              &                                                  &                                   & \multicolumn{1}{l}{}             &                          &                                                   & \cellcolor[HTML]{E0EBEA}Tiny ImageNet & \cellcolor[HTML]{E0EBEA}7\% & \cellcolor[HTML]{FFFFFF}                 & \cellcolor[HTML]{FFFFFF}     \\
Weka                                  & 4\%                            &                                              &                                                  &                                   & \multicolumn{1}{l}{}             &                          &                                                   & \cellcolor[HTML]{FFFFFF}              & \cellcolor[HTML]{FFFFFF}    & \cellcolor[HTML]{FFFFFF}                 &                              \\ \hline
\end{tabular}
\end{sidewaystable*}

\subsection{Combining Data Analytics and Learning on the Edge-to-Cloud Continuum}

\textcolor{blue}{Next, we present the recent efforts on combining state-of-the-art Big Data Analytics and Machine Learning approaches to enable intelligence on the Edge-to-Cloud Continuum. The following works focus on novel systems, frameworks and architectures.} 


A novel architectural design for enabling machine and deep learning over heterogeneous data streams on hybrid Cloud and Edge Computing infrastructures is proposed in~\cite{kourtellis2021s2ce}. Named Stream to Cloud and Edge (S2CE), the platform aims to enable mining of Big Data streams over Cloud and Edge. It provides functionalities of scalable processing, such as distributed processing, data fusion and preprocessing, and Cloud and Edge resource management.

In~\cite{wang2018iot} the authors propose 5G Intelligent Internet of Things (5G I-IoT). This approach is based on Big Data mining, Deep Learning, and Reinforcement Learning to process data intelligently and to optimize communication channels. The framework consists of three building blocks: (1) a processing center in the Cloud to handle real-time data for decision making using Deep Learning and Reinforcement Learning; (2) an object processor in the Fog to processes the raw data from sensing regions; and (3) the sensing regions in the Edge. Evaluations show that 5G I-IoT outperforms 4G-IoT and 5G-IoT in terms of effectiveness of channel utilization.

Authors of~\cite{sankaranarayanan2020data} propose a Data Flow and Distributed Deep Neural Network (DF-DDNN) that integrates data flow and distributed Deep Learning in the IoT-Edge environment to bring down the latency and increase accuracy. Experimental results show that the proposed solution enables a latency reduction of up to 33\% when compared to the existing traditional IoT-Cloud model. In~\cite{grzenda2020hybrid}, a hybrid technique combining batch learning, Online Learning, and stream mining to predict delays of public transport vehicles is proposed. The hybrid approach is validated by experiments with real public transport delay data streams.

Recent studies have also proposed novel architectures for collaborative Edge-Cloud learning and data analytics. A review of existing reference architecture designs of Big Data systems such as FAR-Edge~\cite{faredge} and Global Edge Computing Architecture~\cite{sitton2020geca} is presented in~\cite{paakkonen2020extending}. Authors propose a novel reference architecture design of a Big Data system with a focus on the utilization of ML in Edge Computing environments. In~\cite{khayyam2020artificial} the authors present an overview of AI approaches for Autonomous Vehicle (AV) and propose a concept architecture for integrating Artificial Intelligence with Edge Computing. They also discuss key issues and challenges on: data fusion, such as the reconstruction and understanding of the environment of AV; and Big Data Analytics for training systems and real-time decision-making of AV volumes of data.

A novel architecture that combines a data distribution layer connecting Fog nodes with a Cloud focusing on resilience, near real-time communication, and a traffic modeling approach is proposed in~\cite{perez2018resilient}. The modeling approach is an Online Machine Learning technique named Conditional Restricted Boltzmann Machines (CRBM) to learn and predict traffic telemetry. Experimental results show that the Cloud-based processing approach can produce severe impact in the accuracy of Cloud-learned models due to network connectivity outages between the Fog and the Cloud.

An Edge-Cloud collaborative computing platform for Artificial Intelligence of Things (AIoT) is proposed in~\cite{rong2021edge}. Named Sophon Edge, the platform helps to build and deploy AIoT applications efficiently. It addresses challenges related to building AIoT applications in practice, such as heterogeneity (\emph{e.g.,} communication protocols, data format, operating systems, among others) and accuracy of AI algorithms (\emph{e.g.}, model refinement and tuning).


Authors of~\cite{sarabia2020highly} propose a system to detect falls leveraging an Edge-Fog-Cloud architecture to deploy DL models into resource-constrained devices for DL inference. The architecture exploits Big Data Analytics resources for training DL models on the Cloud and performing inference on devices. They also present a practical and experimental deployment of DL models on Fog devices and the lightweight virtualization technologies, such as Docker containers, to optimize the resource usage. Their solution leverages the RNN (LSTM/GRU) algorithms since they are appropriate for sequential data such as IoT monitoring and they fulfill the resource constraint requirements and provide very high accuracy.

\textcolor{blue}{In summary, the systems, frameworks and architectures proposed by the articles aim to mainly: enable Cloud and Edge resource management; optimize network communication; provide efficient application deployments; and allow distributed data processing. Some approaches also exploit hybrid techniques combining batch learning, Online Learning, Reinforcement Learning, Deep Learning, among others to improve application performance (\emph{e.g.}, minimize latency, increase accuracy, \emph{etc.}).}

\subsection{Main takeaways}

This section aims to answer the following research question: \textit{How are the existing Machine Learning and Data Analytics approaches combined to enable intelligence on the Edge-to-Cloud Continuum?} We organize the selected studies in two main categories, they are: Data Analytics and Machine Learning on the Cloud; and Data Analytics combined with Machine Learning on the Edge-to-Cloud Continuum. 



We highlight that the recent efforts (\emph{e.g.} systems, frameworks and architectures) focus on applying, combining, and deploying Machine Learning paradigms such as Federated Learning, Transfer Learning, Multi-task Learning, Reinforcement Learning and Online Learning on distributed Edge devices for collaborative Edge-to-Cloud analytics. Such efforts focus mainly on addressing open challenges, such as: enabling fast and accurate predictive analytics; optimize communication channels and minimize connectivity issues in Edge data analytics; minimize the processing latency of applications to satisfy Quality of Service requirements; just to cite a few.

\section{
Experimental Research and Reproducibility
}
\label{sec:experimental_research}
In this Section we introduce the main state-of-the-art simulation, emulation, and deployment systems supporting experimental research on the Edge, Fog, Cloud, and Edge-to-Cloud Continuum~\cite{svorobej2019simulating, yousefpour2019all}. Besides, we discuss the relevant experimental testbeds enabling Edge-to-Cloud experiments. We then analyze the previously selected articles in terms of experimental evaluation aspects, such as the size of the experimental testbed and the support to the reproducibility of experiments.

\begin{table*}[t]
\small
\sffamily
\centering
\caption{Simulation, Emulation, and Deployment Systems for Experimental Research on the Edge-to-Cloud Continuum.}
\label{tbl:sed-exp}
\begin{tabular}{m{1.9cm}m{0.6cm}m{0.6cm}m{0.6cm}m{2.8cm}m{8.2cm}}
\hline
 
\rowcolor[HTML]{c4cdc1}\textbf{Simulation Systems} & \textbf{Edge} & \textbf{Fog} & \textbf{Cloud} & \textbf{Main Goal} & \textbf{Key Features} \\ \hline
 
\rowcolor[HTML]{E0EBEA} 
CloudSim~\cite{calheiros2011cloudsim}                                                    &      &     &  \checkmark   &  Modeling, simulation, and experimentation of Cloud infrastructures and application services.    &  

\begin{itemize}
    \item modeling and simulation of large scale Cloud computing data centers.
    \item modeling and simulation of virtualized server hosts and application containers.
    \item modeling and simulation of energy-aware computational resources.
    \item modeling and simulation of data center network topologies.
\end{itemize}

\\
SCORE~\cite{fernandez2018score}                                                       &      &     & \checkmark   &   Simulate energy- efficiency, security, and scheduling strategies in Cloud Computing environments.  &   

\begin{itemize}
    \item allows to prototype and compare different cluster scheduling strategies and policies.
    \item generates synthetic cluster workloads from empirical parameter distributions.
    \item allows the analysis of scheduling performance metrics.
\end{itemize}

\\
\rowcolor[HTML]{E0EBEA} 
ElasticSim~\cite{cai2017elasticsim}                                                 &      &     &  \checkmark  &   Simulate autoscaling algorithms.
  &   

\begin{itemize}
    \item supports resource runtime auto-scaling.
    \item supports stochastic task execution time modeling.
\end{itemize}

\\

iFogSim~\cite{gupta2017ifogsim}                                                     &      &  \checkmark   &    &   Modeling and simulation of resource management techniques in IoT, Edge and Fog Computing environments   &   

\begin{itemize}
    \item inherits a number of features from CloudSim.
    \item provides resource management techniques in IoT, Edge and Fog.
    \item allows the execution of multiple applications on the infrastructure at the same time.
    \item supports migration of application modules from one fog device to another.
\end{itemize}

\\
\rowcolor[HTML]{E0EBEA} 
FogNetSim~\cite{qayyum2018fognetsim++}                                                 &      &  \checkmark   &     &   Simulate distributed fog computing environments.   &   

\begin{itemize}
    \item covers the network aspects such as delay, packet error rate, transmission range, handover, scheduling, and heterogeneous mobile devices.
    \item allows to simulate a large fog network.
    \item allows to simulate heterogeneous devices with varying features.
    \item supports handover: allows static and dynamic nodes in the network.
\end{itemize}

\\

FogTorch~\cite{fogtorch}                                                  &      &  \checkmark   &  &  QoS-aware deployment of IoT applications through the Fog. &

\begin{itemize}
    \item allows the specification of a Fog infrastructure along processing (\emph{e.g.,} CPU cores, RAM memory, storage) and QoS (\emph{e.g.,} latency, bandwidth) capabilities.
    \item allows the specification of applications to be deployed along with needed IoT devices, processing and QoS requirements.
\end{itemize}

\\
\rowcolor[HTML]{E0EBEA} 
FogExplorer~\cite{hasenburg_supporting_2018}                                                 &      &  \checkmark   &    &    Simulate QoS and cost evaluation of fog-based IoT applications.  &    

\begin{itemize}
    \item simulates processing cost and processing time for individual application modules.
    \item simulates transmission cost and transmission time for individual data streams.
\end{itemize}

\\ \hline






\end{tabular}
\end{table*}

\begin{table*}[t]
\small
\sffamily
\centering
\begin{tabular}{m{1.9cm}m{0.6cm}m{0.6cm}m{0.6cm}m{2.8cm}m{8.2cm}}
\hline
\rowcolor[HTML]{c4cdc1}
\textbf{Simulation Systems} & \textbf{Edge} & \textbf{Fog} & \textbf{Cloud} & \textbf{Main Goal} & \textbf{Key Features} \\ \hline
\rowcolor[HTML]{E0EBEA} 
IoTSim-Edge~\cite{nandan2019iotsim}                                                 & \checkmark     &     &    &    Simulate the distribution and processing of streaming data generated by IoT devices in Edge computing environments.  &    

\begin{itemize}
    \item allows to define data analytic operations and their mapping to different parts of the infrastructure.
    \item supports modeling of heterogeneous IoT protocols along with their energy consumption profile.
    \item supports modeling of mobile devices and captures the effect of handoff caused by the movement of mobile devices.
\end{itemize}

\\

EdgeCloud-Sim~\cite{sonmez2018edgecloudsim}                                                &  \checkmark    &     &     &    Simulate environments specific to Edge Computing scenarios.  & 

\begin{itemize}
    \item considers computing and networking resources.
    \item supports network modeling specific to WLAN and WAN.
    \item supports device mobility model and provides realistic and tunable load generator.
\end{itemize}

\\ \hline

\rowcolor[HTML]{E0EBEA} 
\textcolor{blue}{YAFS}~\cite{lera2019yafs}                                                 & \checkmark     &     &    &    \textcolor{blue}{Analyze the design and deployment of applications through customized and dynamical strategies.}  &

\begin{itemize}
    \item \textcolor{blue}{allows dynamic scenarios: placement, path routing, orchestration, and workload movement.}
    \item \textcolor{blue}{supports placement allocation algorithms and orchestration algorithms.}
    \item \textcolor{blue}{provides functions to obtain metrics such as network utilization, network delay, response time, and waiting time.}
\end{itemize}

\\

\textcolor{blue}{XFogSim}~\cite{malik2020xfogsim}                                                &  \checkmark    &     &     &    \textcolor{blue}{Simulate federated fog computing environments.}  & 

\begin{itemize}
    \item \textcolor{blue}{provides resource allocation algorithms for resource sharing.}
    \item \textcolor{blue}{supports static and mobile nodes (handover mechanisms).}
    \item \textcolor{blue}{supports application evaluation in terms of: energy consumption, processing latency, scalability, and resource usage.}
\end{itemize}

\\ \hline

\rowcolor[HTML]{c4cdc1}
\textbf{Emulation Systems} & \textbf{Edge} & \textbf{Fog} & \textbf{Cloud} & \textbf{Main Goal} & \textbf{Key Features} \\ \hline
\rowcolor[HTML]{E0EBEA} 
EmuFog~\cite{mayer2017emufog}                                                      &      & \checkmark    &     &   Enable the design of Fog Computing infrastructures and the emulation of real applications and workloads.   &    

\begin{itemize}
    \item generates networks that can be emulated easily with MaxiNet~\cite{wette2014maxinet}.
    \item supports topologies from BRITE~\cite{medina2001brite} and Caida~\cite{caidanet}.
    \item places fog nodes based on user-defined constrains (\emph{e.g.,} network latency or resource constraints).
\end{itemize}

\\
Fogbed~\cite{coutinho2018fogbed}                                                        &      & \checkmark    &     &  Enable the rapid prototyping of Fog components in
virtualized environments.    &   

\begin{itemize}
    \item allows dynamic topology changes.
    \item provides traffic control links such as delay, rate, loss, and jitter.
    \item enables the deployment of Fog nodes as software containers under different network configurations.
\end{itemize}

\\

\rowcolor[HTML]{E0EBEA} 
\textcolor{blue}{RADICAL-DREAMER} \cite{radicaldreamer}                                                      & \checkmark      & \checkmark    & \checkmark   &   \textcolor{blue}{Emulate resource and task/workload definition in Edge-to-Cloud applications.}   &    

\begin{itemize}
    \item \textcolor{blue}{allows to evaluate workload and resource management aspects of applications.}
    \item \textcolor{blue}{supports modeling task placement in Edge-to-Cloud applications.}
    \item \textcolor{blue}{allows to evaluate deployment modalities and performance trade-offs.}
\end{itemize}

\\ \hline
\end{tabular}
\end{table*}

\begin{table*}[t]
\small
\sffamily
\centering
\begin{tabular}{m{1.9cm}m{0.6cm}m{0.6cm}m{0.6cm}m{2.8cm}m{8.2cm}}
\hline
\rowcolor[HTML]{c4cdc1}
\textbf{Deployment Systems}                                         & \textbf{Edge} & \textbf{Fog} & \textbf{Cloud} & \textbf{Main Goal} & \textbf{Key Features} \\ \hline
\rowcolor[HTML]{E0EBEA} 
\textbf{E2C}\textit{lab}~\cite{rosendo2020e2clab}                                                      & \checkmark     &  \checkmark   & \checkmark   &   Understand and optimize the performance of Edge-to-Cloud workflows through reproducible experiments on large-scale infrastructures.   &     

\begin{itemize}
    \item supports reproducible experiments.
    \item provides a \emph{Services} abstraction to support other applications.
    \item provides resource monitoring (\emph{e.g.,} CPU, GPU, memory, network) and network emulation to define Edge-to-Cloud constraints such as delay, loss and rate.
    \item maps application parts with the underlaying testbed.
    \item allows to optimize workflows through optimization libraries for hyperparameter search in Ray Tune~\cite{liaw2018tune}.
\end{itemize}

\\ 
KubeEdge~\cite{xiong2018extend}                                                 &  \checkmark    &     &   &    Deploy complex high level applications to the Edge.  &   

\begin{itemize}
    \item provides containerized application orchestration and device management to hosts at the Edge.
    \item provides core infrastructure support for networking, application deployment and metadata synchronization between Cloud and Edge.
    \item supports MQTT which enables Edge devices to access through Edge nodes.
\end{itemize}

\\

\rowcolor[HTML]{E0EBEA} Kubernetes~\cite{brewer2015kubernetes}    &      &     &  \checkmark   &    Manage and automate the deployment, scaling, and management of containerized applications across multiple hosts.  &       

\begin{itemize}
    \item provides mechanisms for deployment, maintenance, and scaling of applications.
    \item provides service discovery and load balancing.
    \item allows to automatically mount storage systems, such as local storage and public Cloud providers.
\end{itemize}

\\ \hline
\end{tabular}
\end{table*}

\subsection{Simulation, Emulation, and Deployment Systems for Experimental Research}
Table~\ref{tbl:sed-exp} summarizes the main open-source state-of-the-art simulation, emulation, and deployment systems for experimental research on the Edge-to-Cloud Continuum.

\subsubsection{Simulation Systems}
Building experimental testbed environments is expensive and brings challenges to conduct reproducible experiments. In this sense, simulation systems play an important role as they allow to analyze systems behavior at very large scale while easily tuning a myriad of configuration parameters. Next, we present simulation systems used in the modeling of Cloud, Fog, and Edge computing environments.

\paragraph{Cloud-based simulation systems.} 
CloudSim~\cite{calheiros2011cloudsim} framework allows modeling and simulation of Cloud computing infrastructures and services in a repeatable manner. CloudSim allows users to model the behavior data centers, Virtual Machines and resource provisioning policies. ElasticSim~\cite{cai2017elasticsim} is a workflow simulator that extends CloudSim. It focuses on supporting resource runtime auto-scaling and stochastic task execution time modeling. SCORE~\cite{fernandez2018score} allows the execution of heterogeneous workloads for simulating energy-efficient monolithic and parallel-scheduling models. 

\paragraph{Fog-based simulation systems.}
FogExplorer~\cite{hasenburg_supporting_2018} provides modeling and simulation to estimate QoS and cost evaluation of Fog-based IoT applications. FogExplorer allows users to choose good application designs during its design phase. FogTorch~\cite{fogtorch} aims to support the deployment of IoT applications in Fog infrastructures considering software, hardware and QoS requirements. FogNetSim++~\cite{qayyum2018fognetsim++} focuses on simulating large Fog networks and differs from others mainly by providing features that allow users to incorporate customized mobility models, scheduling algorithms, and manage handover mechanisms. \textcolor{blue}{XFogSim~\cite{malik2020xfogsim} extends FogNetSim++ to simulate federated fog computing environments. xFogSim is lightweight, configurable, scalable and introduces the concept of fog federation for resource sharing among fog locations. Furthermore, it allows users to evaluate applications in terms of energy consumption, processing latency, scalability, and resource usage. YAFS~\cite{lera2019yafs} aims to allow users to analyze application designs and incorporate strategies for placement, scheduling and routing. YAFS also supports dynamic allocation of new application modules, dynamic failures of network nodes, and user mobility. Furthermore, it facilitates the shareability of experiment results by generating logs of workload generation and computation, and link transmissions.} Lastly, iFogSim~\cite{gupta2017ifogsim} focuses on resource management techniques in IoT, Edge and Fog computing environments. iFogSim allows users to measure, in a repeatable manner, the impact of resource management techniques in terms of latency, network congestion, energy consumption, and cost.

\paragraph{Edge-based simulation systems.} 
EdgeCloudSim~\cite{sonmez2018edgecloudsim} focuses on Edge Computing scenarios and allows one to conduct experiments considering computational and networking resources. IoTSim-Edge~\cite{nandan2019iotsim} allows users to easily configure their Edge infrastructures and to capture the behavior of heterogeneous IoT and Edge devices in terms of sensing, processing, mobility, and data rate. Both Edge systems extend CloudSim.




\subsubsection{Emulation Systems}
Compared to simulation, the emulation approach provides more realistic results. While simulators mimic the behavior and configurations of a real device, emulation systems duplicate the hardware and software features of a real device~\cite{sreerangaraju}. Emulation systems are also a less expensive solution when compared to real deployments.

Fogbed~\cite{coutinho2018fogbed} allows resource provisioning emulation in Fog environments. It combines Containernet~\cite{peuster2018containernet} and Maxinet~\cite{wette2014maxinet} (both are extensions of the Mininet~\cite{kaur2014mininet} network emulator) to allow the use of virtual instances for resource provisioning emulation. 

EmuFog~\cite{mayer2017emufog} focuses on the design of Fog Computing infrastructures and the emulation of real applications and workloads. In EmuFog, users can: design the network topology; embed Fog nodes in the topology; and run Docker-based applications on those nodes connected by an emulated network.

\textcolor{blue}{RADICAL-DREAMER~\cite{radicaldreamer} provides the concepts of \textit{Task} and \textit{Workload} to model the characteristics of an application according to heterogeneous tasks. Besides, it provides the concept of \textit{Resource} to model distributed infrastructures. RADICAL-DREAMER allows users to evaluate deployment configurations, performance trade-offs, and workload placement strategies for Edge-to-Cloud applications~\cite{luckow2021exploring}.}

\subsubsection{Deployment Systems}
Deploying real-life applications on large-scale testbeds provides the most realistic results compared to simulation or emulation approaches. In this direction, a few systems have been proposed in the past few years.

E2Clab~\cite{rosendo2020e2clab} is a framework that implements a rigorous methodology for designing experiments with real-world workloads on the Edge-to-Cloud Continuum. E2Clab provides guidelines to move from real-world use cases to the design of relevant testbed setups for reproducible experiments enabling researchers to understand and optimize~\cite{rosendo2021reproducible} the performance of applications. The key features provided by E2Clab are~\cite{rosendo2021e2clab}: (1) reproducible experiments; (2) the mapping of applications parts executed across the computing continuum with the physical testbed; (3) the support for experiment variation and transparent scaling of the scenario; (4) network emulation to define Edge-to-Cloud communication constraints; (5) experiment deployment, monitoring and backup of results; and (6) the application optimization.

Kubernetes~\cite{brewer2015kubernetes} aims to simplify the deployment and management of services that compose an application by providing mechanisms for deployment, maintenance, and scaling. Using Kubernetes, users can manage containerized applications across multiple hosts. KubeEdge~\cite{xiong2018extend} builds on top of Kubernetes to extend Cloud capabilities to the Edge and allows containerized application orchestration and device management to hosts at the Edge. KubeEdge key features are: core infrastructure support for networking; application deployment; and metadata synchronization between Cloud and Edge.

\subsection{Large-Scale Experimental Testbeds for Edge-to-Cloud Experiments}

Several experimental testbeds allow researchers to evaluate their proposals in real-life settings by providing access to a large amount of resources
(grouped in homogeneous or heterogeneous clusters, upon convenience) and, more importantly, supported by some vibrant communities of users and solid technical teams. We cite here just a few. 

Grid5000~\cite{bolze:hal-00684943} is a large-scale French testbed for experimental research with a focus on parallel and distributed computing including Cloud, HPC, Big Data, and AI. Grid5000 is merging with FIT IoT-Lab to enable Edge-to-Cloud experiments. FIT IoT-Lab~\cite{adjih2015fit} is a large-scale multi-radio (\emph{e.g.}, IEEE 802.15.4, Bluetooth Low Energy, LoRa, etc.) and multi-platform (\emph{e.g.}, Arduino Zero, nRF52840-MDK, LoRa gateway, and many others) infrastructure for the Internet of Things. FIT IoT-Lab consists of more than 1.5K nodes and provides tools for monitoring energy consumption and network-related metrics, such as end-to-end delay, throughput and overhead. A recent effort on supporting experiments combining Grid5000 and FIT IoT-Lab testbeds is EnOSlib~\cite{cherrueau2021enoslib}. EnOSlib is a library which brings reusable building blocks for configuring the infrastructure, provisioning software on remote hosts as well as organizing the experimental workflow. 

Chameleon~\cite{keahey2020lessons} is a large-scale US experimental platform that aims to support Computer Science research in many areas, such as: systems, storage, networking, GPU, security, Artificial Intelligence, and High Performance Computing. CHI@Edge is an extension of Chameleon testbed that aims to support Edge Computing experiments. Combining Chameleon and CHI@Edge testbeds allows more realistic Edge-to-Cloud experiments since it provides access to real-life IoT/Edge devices such as Raspberry Pis, Jetson Nanos, among others. 

ORBIT~\cite{orbit2016open} (Open-Access Research Testbed for Next-Generation Wireless Networks) is based on a 20x20 two-dimensional grid of programmable radio nodes which can be interconnected into different topologies. ORBIT provides access to: radio resources, including WiFi 802.11a/b/g 802.11n 802.11ac, Bluetooth (BLE), ZigBee, and Software Defined Radio platforms; Software defined networking (SDN) resources; LTE and WiMAX base stations and clients; and Cloud resources such as nodes with Tesla-based GPUs.   

SmartSantander~\cite{sanchez2014smartsantander} is a large scale testbed composed of around 2000 IEEE 802.15.4 devices deployed in a 3-tiered architecture (IoT node, repeaters, and gateway node) deployment in the Spanish city of Santander. The testbed allows IoT native experimentation (\emph{e.g.} wireless sensor network experiments) and service provision experiments (\emph{e.g.} applications using real-time real-world sensor data).

Fed4FIRE+~\cite{demeester2016fed4fire} is a project offering the largest federation worldwide of Next Generation Internet (NGI) testbeds. Fed4FIRE aims to provide open, accessible and reliable experimental infrastructures supporting a wide variety of research, such as 5G, IoT, Cloud Computing, Wired and Wireless Computer Networking. The list of testbeds~\cite{nussbaum2019overview} federated with Fed4FIRE are: CityLab~\cite{struye2018citylab}, PlanetLab~\cite{fiuczynski2006planetlab}, ExoGENI~\cite{baldin2016exogeni}, Tengu~\cite{vanhove2015tengu}, NITOS~\cite{pechlivanidou2014nitos}, w-iLab~\cite{bouckaert2010w}, among others.


\begin{figure}[t]
  \centering
  \includegraphics[width=0.9\linewidth]{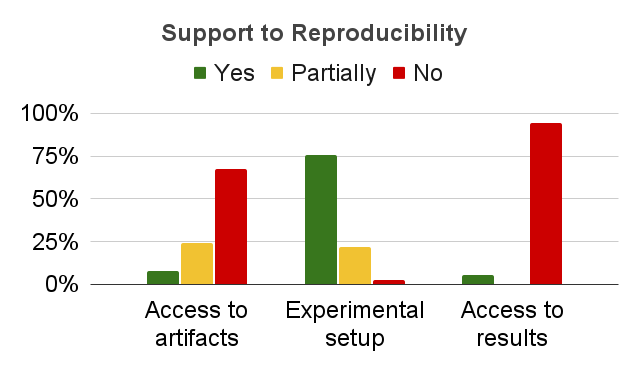}
  \caption{Support to the reproducibility of experiments provided by the selected studies.}
  \label{fig:exp-repro}
\end{figure}

\begin{figure}[t]
  \centering
  \includegraphics[width=0.92\linewidth]{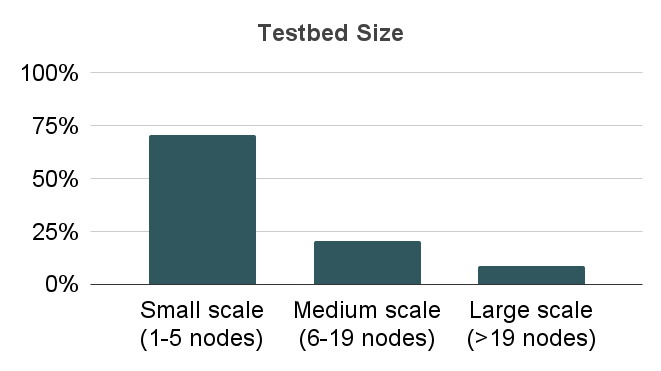}
  \caption{\textcolor{blue}{Testbed size used in the experimental evaluations: small scale~\cite{zhou2019distributing, kumar2019decaf, chen2019exploring, lu2019collaborative, zhang2018pcamp, hong2019dlion, wang2018iot, dautov2019hierarchical, sarabia2019efficient, dautov2017pushing, grzenda2020hybrid, zhou2020saface, mrozek2020fall, ali2020res, ghosh2020edge, alencar2020fot, xu2021move, guo2021mistify, li2019edge, fafoutis2018extending, kumar2017resource, dey2019offloaded, assefi2017big, nair2018applying}, medium scale~\cite{dautov2018data, perez2018resilient, sankaranarayanan2020data, sarabia2020highly, dautov2020stream, fu2019edgewise, xian2020parallel}, and large scale~\cite{hauswirth2020autonomous, rosendo2020e2clab, rosendo2021reproducible}}}
  \label{fig:exp-size}
\end{figure}

\begin{table}[t]
\centering
\caption{ACM Digital Library Terminology \cite{ferro2018sigir}}
\label{tbl:acm-terminology}
\resizebox{\linewidth}{!}{
\begin{tabular}{ll}
\hline
\rowcolor[HTML]{e2eaec}
\parbox[t]{2mm}{\rotatebox[origin=c]{90}{\textbf{\hspace{0.3cm}Repeatability\hspace{0.3cm}}}} & 
\textit{\begin{tabular}[c]{@{}l@{}} Same team, same experimental setup: the measurement can\\ be obtained with stated precision by the same team using\\ the same measurement procedure, the same measuring system,\\ under the same operating conditions, in the same location\\ on multiple trials. For computational experiments, this means\\ that a researcher can reliably repeat their own computation.\end{tabular}}
\\
\hline
\parbox[t]{2mm}{\rotatebox[origin=c]{90}{\textbf{\hspace{0.6cm}Replicability\hspace{0.6cm}}}} &
\textit{\begin{tabular}[c]{@{}l@{}} Different team, same experimental setup: the measurement can\\ be obtained with stated precision by a different team using\\ the same measurement procedure, the same measuring system, \\under the same operating conditions, in the same or a different\\ location on multiple trials. For computational experiments, this\\ means that an independent group can obtain the same result\\ using the author’s own artifacts.\end{tabular}}
\\
\hline
\rowcolor[HTML]{e2eaec}
\parbox[t]{2mm}{\rotatebox[origin=c]{90}{\textbf{\hspace{0.3cm}Reproducibility\hspace{0.3cm}}}} &
\textit{\begin{tabular}[c]{@{}l@{}} Different team, different experimental setup: the measurement\\ can be obtained with stated precision by a different team, a\\ different measuring system, in a different location on multiple\\ trials. For computational experiments, this means that an inde-\\pendent group can obtain the same result using artifacts which\\ they develop completely independently.\end{tabular}}\\ 
\hline
\end{tabular}}
\end{table}

\subsection{Support to Experimental Reproducibility}
\label{subsec:supp-rep}

A desired feature of any experimental research is that its scientific claims are verifiable by others in order to build upon them. This can be achieved through Repeatability, Replicability, and Reproducibility~\cite{ieee-computing, stodden2013best}. Find in Table~\ref{tbl:acm-terminology} the terminology proposed by the ACM Digital Library.

We evaluate the support to the reproducibility of experiments for each selected article. This evaluation is based on the following three main relevant aspects:

\textbf{Access to artifacts}: if authors provide access to a public repository with the artifacts used to run the experiments, such as: datasets, codes, applications, systems, configuration files, among others.

\textbf{Experimental setup}: if authors provide a description of the experimental setup, such as: hardware configuration of physical machines, software or systems used, network configurations, among others.

\textbf{Access to results}: if the computed experimental results are available in a public repository, such as: log files, files metric collected during runtime, monitoring data, code to plot charts, among others.


Figure~\ref{fig:exp-repro} summarizes the support to the reproducibility of experiments provided by the selected studies. Regarding the \textbf{access to artifacts}, 68\% of papers do not provide access to them, and just 24\% partially provide (a few artifacts, but not all). Analyzing the description of the \textbf{experimental setup}, 76\% of papers describe it in detail in a dedicated section of the paper, while 21\% only partially describe it and just 3\% do not provide enough information. Lastly, regarding the \textbf{access to results}, 95\% of the articles do not provide access and just 5\% provide a public repository with the results. In general, we notice a lack of support to the experimental reproducibility in the domain of Edge-to-Cloud experimental research.  

Lastly, Figure~\ref{fig:exp-size} presents the size of the testbeds used in the experimental evaluations. As one may note, 70\% of papers use small scale setups, composed by at most 5 machines or devices, while 20\% of them use testbed setups composed by 6 to 19 nodes, and just 10\% experiment in large scale setups with 20 nodes or more.

\subsection{Main takeaways}

This section aims to answer the following research question: \textit{What are the existing solutions for experimental research and how do the selected studies support the reproducibility of the experiments?} We identify and summarize the key characteristics of the main state-of-the-art simulation, emulation, and deployment systems. Furthermore, we discuss the recent efforts and initiatives merging large scale testbeds for enabling more realistic setups for Edge-to-Cloud experiments, such as: Grid'5000 and FIT IoT-Lab; Chameleon and CHI@Edge; and the Fed4FIRE project.

We also analyze the selected studies regarding their support to the reproducibility of experiments. \textcolor{blue}{As a conclusion, the results presented in Figure~\ref{fig:exp-repro} reinforce the need for rigorous experimental methodologies that provide guidelines to the reproducibility of experiments in the Edge-to-Cloud research domain. At the same time, Figure~\ref{fig:exp-size} highlights the need for methodologies and deployment systems guiding researchers to evaluate and validate their proposed approaches in large-scale environments.}

The development of novel systems, frameworks, or libraries abstracting the complexities of deploying Edge-to-Cloud workflows on large scale testbeds in addition with the management of the whole experimental cycle such as monitoring, gathering of results, and provenance of the experimental setup are extremely relevant. Recent advances in this direction exist, like the EnOSlib~\cite{cherrueau2021enoslib} library or the \textbf{E2C}\textit{lab}~\cite{rosendo2020e2clab} framework, but further advances are still needed.

\section{\textcolor{blue}{Major Findings}}
\label{sec:major-findings}

\begin{table}[t]
\small
\centering
\caption{Machine Learning frameworks/libraries designed for the Cloud. Marked with a $\star$ are the mostly exploited in the experiments.}
\label{tbl:ml-cloud}
\begin{tabular}{m{2.5cm}m{0.5cm}m{4.0cm}}
\hline
\rowcolor[HTML]{C4CDC1} 
\cellcolor[HTML]{C4CDC1}\textbf{Framework/Library} & \cellcolor[HTML]{C4CDC1}\textbf{Qty.} & \textbf{Paper}                                                                                                                                                                                                                                                                                                                                                                                                                                        \\ \hline
\cellcolor[HTML]{E0EBEA}$\star$ \textbf{Tensorflow}                                         & \cellcolor[HTML]{E0EBEA}24                                                        &\cellcolor[HTML]{E0EBEA} \cite{zhou2019distributing, chen2019exploring, lu2019collaborative, zhang2018pcamp, talagala2018eco, hong2019dlion, l2017machine, mohammadi2018deep, liu2019survey, zhang2019deep, prabhu2019fog, pathak2018construing, nguyen2019machine, sarabia2019efficient, kukreja2019training, xu2019first, yousefpour2019all, sarabia2020highly, paakkonen2020extending, rocha2020distributed, chen2019deep, guo2021mistify, rong2021edge, dey2019offloaded} \\ 
$\star$ \textbf{PyTorch}                                            & 10                                                        & \cite{zhang2018pcamp, mohammadi2018deep, liu2019survey, zhang2019deep, nguyen2019machine, kukreja2019training, xu2019first, yousefpour2019all, zhou2020saface, chen2019deep}                                                                                                                                                                                                                                                                         \\ 
\cellcolor[HTML]{E0EBEA}Caffe                                              & \cellcolor[HTML]{E0EBEA}7                                                         &\cellcolor[HTML]{E0EBEA} \cite{mohammadi2018deep, zhang2019deep, nguyen2019machine, kukreja2019training, xu2019first, chen2019deep, dey2019offloaded}                                                                                                                                                                                                                                                                                                                         \\ 
SAMOA                                              & 6                                                         & \cite{ali2018recent, l2017machine, rao2019big, ulusar2020open, lobo2020spiking, kourtellis2021s2ce}                                                                                                                                                                                                                                                                                                                                                  \\ \cellcolor[HTML]{E0EBEA}
Mahout                                             &\cellcolor[HTML]{E0EBEA} 6                                                         &\cellcolor[HTML]{E0EBEA} \cite{ali2018recent, l2017machine, yang2017iot, prabhu2019fog, rao2019big, ulusar2020open}                                                                                                                                                                                                                                                                                                                                                           \\ 
$\star$ \textbf{Scikit-Learn}                                       & 5                                                         & \cite{lu2019collaborative, nguyen2019machine, lobo2020spiking, sankaranarayanan2020data, sarabia2020highly}                                                                                                                                                                                                                                                                                                                                          \\ \cellcolor[HTML]{E0EBEA}
CNTK                                               & \cellcolor[HTML]{E0EBEA}5                                                         & \cellcolor[HTML]{E0EBEA}\cite{zhang2018pcamp, hong2019dlion, mohammadi2018deep, nguyen2019machine, xu2019first}                                                                                                                                                                                                                                                                                                                                                              \\ 
MOA                                                & 5                                                         & \cite{ali2018recent, l2017machine, lobo2020spiking, grzenda2020hybrid, kourtellis2021s2ce}                                                                                                                                                                                                                                                                                                                                                           \\ \cellcolor[HTML]{E0EBEA}
Spark MLlib                                        & \cellcolor[HTML]{E0EBEA}5                                                         &\cellcolor[HTML]{E0EBEA} \cite{ali2018recent, rao2019big, kolodziej2019high, ulusar2020open, nguyen2019machine}                                                                                                                                                                                                                                                                                                                                                               \\ 
$\star$ \textbf{Keras}                                              & 4                                                         & \cite{lu2019collaborative, nguyen2019machine, sarabia2019efficient, sarabia2020highly}                                                                                                                                                                                                                                                                                                                                                               \\ \cellcolor[HTML]{E0EBEA}

Chainer                                            & \cellcolor[HTML]{E0EBEA}4                                                         & \cellcolor[HTML]{E0EBEA}\cite{mohammadi2018deep, nguyen2019machine, xu2019first, li2019edge}                                                                                                                                                                                                                                                                                                                                                                                 \\

R                                                  & 3                                                         & \cite{ali2018recent, l2017machine, perez2018resilient}                                                                                                                                                                                                                                                                                                                                                                                               \\ \rowcolor[HTML]{E0EBEA}
Vowpal Wabbit                                      & 3                                                         & \cite{l2017machine, nguyen2019machine, kourtellis2021s2ce}                                                                                                                                                                                                                                                                                                                                                                                           \\ 
Theano                                             & 3                                                         & \cite{mohammadi2018deep, zhang2019deep, nguyen2019machine}                                                                                                                                                                                                                                                                                                                                                                                           \\ \rowcolor[HTML]{E0EBEA}

Gaia                                               & 2                                                         & \cite{hong2019dlion, zhang2019deep}                                                                                                                                                                                                                                                                                                                                                                                                                  \\ 
Flink ML                                           & 2                                                         & \cite{rao2019big, nguyen2019machine}                                                                                                                                                                                                                                                                                                                                                                                                                 \\ \rowcolor[HTML]{E0EBEA}
Mistify                                            & 1                                                         & \cite{guo2021mistify}                                                                                                                                                                                                                                                                                                                                                                                                                                \\ 
CLONE                                              & 1                                                         & \cite{lu2019collaborative}                                                                                                                                                                                                                                                                                                                                                                                                                           \\ \rowcolor[HTML]{E0EBEA}
DLion                                              & 1                                                         & \cite{hong2019dlion}                                                                                                                                                                                                                                                                                                                                                                                                                                 \\ 
Ako                                                & 1                                                         & \cite{hong2019dlion}                                                                                                                                                                                                                                                                                                                                                                                                                                 \\ \rowcolor[HTML]{E0EBEA}
TUX2                                               & 1                                                         & \cite{zhang2019deep}                                                                                                                                                                                                                                                                                                                                                                                                                                 \\ 
Weka                                               & 1                                                         & \cite{assefi2017big}                                                                                                                                                                                                                                                                                                                                                                                                                                 \\ \hline
\end{tabular}
\end{table}

\begin{table}[t]
\small
\centering
\caption{Machine Learning frameworks/libraries designed for the Edge. Marked with a $\star$ are the mostly exploited in the experiments.}
\label{tbl:ml-edge}
\begin{tabular}{m{2.5cm}m{0.5cm}m{4.0cm}}
\hline
\rowcolor[HTML]{C4CDC1} 
\cellcolor[HTML]{C4CDC1}\textbf{Framework/Library} & \cellcolor[HTML]{C4CDC1}\textbf{Qty.} & \textbf{Paper}                                                                                                                                                                             \\ \hline
\cellcolor[HTML]{E0EBEA}$\star$ \textbf{MXNet}                                              & \cellcolor[HTML]{E0EBEA} 11                                                        &\cellcolor[HTML]{E0EBEA}\cite{zhang2018pcamp, hong2019dlion, mohammadi2018deep, liu2019survey, zhang2019deep, nguyen2019machine, nikouei2019toward, xu2019first, yousefpour2019all, zhou2020saface, chen2019deep} \\  
$\star$ \textbf{Caffe2}                                             & 7                                                         & \cite{zhang2018pcamp, liu2019survey, zhang2019deep, nguyen2019machine, xu2019first, yousefpour2019all, chen2019deep}                                                                      \\ \cellcolor[HTML]{E0EBEA}
TF Lite                                            & \cellcolor[HTML]{E0EBEA}5                                                         & \cellcolor[HTML]{E0EBEA}\cite{zhang2018pcamp, liu2019survey, xu2019first, yousefpour2019all, loghin2020disruptions}                                                                               \\  
CoreML                                             & 4                                                         & \cite{liu2019survey, zhang2019deep, xu2019first, mrozek2020fall}                                                                                                                          \\ \cellcolor[HTML]{E0EBEA}
TF Federated                                       &\cellcolor[HTML]{E0EBEA} 2                                                         & \cellcolor[HTML]{E0EBEA}\cite{kumar2019decaf, loghin2020disruptions}                                                                                                                                              \\  
Neurosurgeon                                       & 2                                                         & \cite{kumar2019decaf, liu2019survey}                                                                                                                                                      \\ \cellcolor[HTML]{E0EBEA}
MOCHA                                              & \cellcolor[HTML]{E0EBEA}1                                                         & \cellcolor[HTML]{E0EBEA}\cite{kumar2019decaf}                                                                                                                                                                     \\  
FedProx                                            & 1                                                         & \cite{kumar2019decaf}                                                                                                                                                                     \\ \cellcolor[HTML]{E0EBEA}
TensorRT                                           & \cellcolor[HTML]{E0EBEA}1                                                         & \cellcolor[HTML]{E0EBEA}\cite{kumar2019decaf}                                                                                                                                                                     \\ \hline
\end{tabular}
\end{table}

\begin{table}[t]
\small
\centering
\caption{Data Analytics frameworks designed for the Cloud. Marked with a $\star$ are the mostly exploited in the experiments.}
\label{tbl:da-cloud}
\begin{tabular}{m{2.5cm}m{0.5cm}m{4.0cm}}
\hline
\rowcolor[HTML]{C4CDC1} 
\cellcolor[HTML]{C4CDC1}\textbf{Framework/Library} & \cellcolor[HTML]{C4CDC1}\textbf{Qty.} & \textbf{Paper}                                                                                                                                                                                                                                                                                                                                                                       \\ \hline
\cellcolor[HTML]{E0EBEA}$\star$ \textbf{Spark}                                              & \cellcolor[HTML]{E0EBEA}22                                                        & \cellcolor[HTML]{E0EBEA}\cite{talagala2018eco, ali2018recent, l2017machine, verma2017survey, sezer2017context, mohammadi2018deep, yang2017iot, liu2019survey, prabhu2019fog, rao2019big, kolodziej2019high, ulusar2020open, pathak2018construing, nguyen2019machine, lobo2020spiking, sarabia2020highly, kourtellis2021s2ce, rong2021edge, assefi2017big, nair2018applying, xian2020parallel} \\  
$\star$ \textbf{Flink}                                              & 15                                                        & \cite{talagala2018eco, verma2017survey, sezer2017context, mohammadi2018deep, liu2019survey, rao2019big, ulusar2020open, nguyen2019machine, dautov2019hierarchical, huang2018building, lobo2020spiking, grzenda2020hybrid, kourtellis2021s2ce, xu2021move, rosendo2020e2clab}                                                                                                        \\ \cellcolor[HTML]{E0EBEA}
$\star$ \textbf{Kafka}                                              &\cellcolor[HTML]{E0EBEA}15                                                        & \cellcolor[HTML]{E0EBEA}\cite{ali2018recent, yang2017iot, rao2019big, kolodziej2019high, ulusar2020open, pathak2018construing, nguyen2019machine, huang2018building, grzenda2020hybrid, sankaranarayanan2020data, sarabia2020highly, kourtellis2021s2ce, alencar2020fot, rosendo2020e2clab}                                                                                                     \\  
Storm                                              & 11                                                        & \cite{ali2018recent, l2017machine, verma2017survey, sezer2017context, mohammadi2018deep, ulusar2020open, lobo2020spiking, sankaranarayanan2020data, fu2019edgewise, kourtellis2021s2ce, xian2020parallel}                                                                                                                                                                           \\ \cellcolor[HTML]{E0EBEA}
Hadoop                                             & \cellcolor[HTML]{E0EBEA}8                                                         &\cellcolor[HTML]{E0EBEA} \cite{l2017machine, mohammadi2018deep, prabhu2019fog, rao2019big, ulusar2020open, pathak2018construing, nguyen2019machine, xian2020parallel}                                                                                                                                                                                                                                        \\  
Samza                                              & 4                                                         & \cite{ali2018recent, rao2019big, lobo2020spiking, kourtellis2021s2ce}                                                                                                                                                                                                                                                                                                               \\ \cellcolor[HTML]{E0EBEA}
Flume                                              & \cellcolor[HTML]{E0EBEA}3                                                         & \cellcolor[HTML]{E0EBEA}\cite{rao2019big, ulusar2020open, grzenda2020hybrid}                                                                                                                                                                                                                                                                                                                                \\  
Cameo                                              & 1                                                         & \cite{xu2021move}                                                                                                                                                                                                                                                                                                                                                                   \\ \cellcolor[HTML]{E0EBEA}
Druid                                              & \cellcolor[HTML]{E0EBEA}1                                                         & \cellcolor[HTML]{E0EBEA}\cite{verma2017survey}                                                                                                                                                                                                                                                                                                                                                              \\ \hline
\end{tabular}
\end{table}

\begin{table}[t]
\small
\centering
\caption{Data Analytics frameworks designed for the Edge. Marked with a $\star$ is the mostly exploited in the experiments.}
\label{tbl:da-edge}
\begin{tabular}{m{2.5cm}m{0.5cm}m{4.0cm}}
\hline
\rowcolor[HTML]{C4CDC1} 
\cellcolor[HTML]{C4CDC1}\textbf{Framework/Library} & \cellcolor[HTML]{C4CDC1}\textbf{Qty.} & \textbf{Paper}                                                                                            \\ \cellcolor[HTML]{E0EBEA}
$\star$ \textbf{Nifi}                                               & \cellcolor[HTML]{E0EBEA}5                                                         & \cellcolor[HTML]{E0EBEA}\cite{dautov2018data, dautov2017pushing, sankaranarayanan2020data, dautov2020stream, kourtellis2021s2ce} \\ 
Edgent                                             & 3                                                         & \cite{liu2019survey, alencar2020fot, rosendo2020e2clab}                                                  \\  \cellcolor[HTML]{E0EBEA}
EdgeWise                                           & \cellcolor[HTML]{E0EBEA}1                                                         & \cellcolor[HTML]{E0EBEA}\cite{fu2019edgewise}                                                                                    \\ \hline
\end{tabular}
\end{table}

\textcolor{blue}{We gained the following insights and learned some lessons from this systematic review:}

\begin{enumerate}
    \item \textcolor{blue}{The \textbf{most common AI frameworks and libraries} exploited in the articles are: Tensorflow and PyTorch for the Cloud; and MXNet and Caffe2 (now part of PyTorch) for Deep Learning on the Edge. In turn, the \textbf{most common Data Analytics frameworks} used in the articles are: Apache Spark, Apache Flink, and Apache Kafka for the Cloud; and Apache Nifi for the Edge. Very \textbf{few open source Data Analytics} frameworks designed \textbf{for the Edge} were identified.}
    \item \textcolor{blue}{The most cited AI learning paradigms are respectively: \textbf{Distributed ML/DL, Online Learning, Reinforcement Learning, Transfer/Multi-task Learning}, and \textbf{Federated Learning}. Although widely used, very few articles are exploring their \textbf{performance trade-offs at scale} and their potential jointly utilization across the Edge-to-Cloud Continuum.}
    \item \textcolor{blue}{The \textbf{hardware heterogeneity}, regarding Edge devices, is \textbf{not sufficiently analyzed} in the validation phase of the proposed systems, frameworks and architectures designed to enable intelligence on the Edge-to-Cloud Continuum. Evaluations mainly rely on  Raspberry Pi's or emulate resource-limited devices. Given the highly heterogeneity characteristic of the Edge-to-Cloud Continuum, it is strongly recommended that future works exploit \textbf{GPUs} and \textbf{TPUs} enabled devices, in addition to CPUs.}
    \item \textcolor{blue}{The majority of the articles proposing novel approaches or exploring existing solutions to enable distributed intelligence on the Edge-to-Cloud Continuum are performing evaluations on \textbf{small-scale testbeds} (\emph{e.g.}, with less than 6 machines or devices, in average). It is strongly recommended that the proposed distributed approaches for ML and DL training/inference or data stream processing be validated in larger-scale environments, to assess the issues of real-life Edge-to-Cloud applications.}
    \item \textcolor{blue}{We observed that the articles \textbf{do not follow systematic experimental methodologies}. Hence, most of them do not provide enough support to enable the \textbf{reproducibility} of the experiments by other researches. Despite most of the articles describing the experimental setup in a dedicated section of the paper, most of them do not provide access to public repositories sharing the artifacts used neither the results obtained. It is strongly recommended that future works consider adopting rigorous methodologies in their experimental evaluations. We highlight that relevant conferences and journals on Computer Science are adopting the \emph{Reproducibility Initiative}~\cite{parashar2022eic}, which consists in assigning reproducibility badges to articles submitting their artifacts for post-publication peer review.}
\end{enumerate}

\section{
Open Challenges and Research Opportunities
}
\label{sec:issues}
As presented in the previous sections, distributed digital infrastructures for Data Analytics and learning are now evolving towards an interconnected ecosystem allowing complex applications to be executed from IoT Edge devices to the HPC Cloud. Therefore, new challenging application scenarios are emerging from a variety of domains such as healthcare, asset monitoring in industry, precision agriculture and smart cities, where processing can no longer rely only on traditional approaches that send all data to centralized datacenters for Data Analytics and Machine Learning. Next, we present some of the relevant challenges and research opportunities to be addressed to enable the Computing Continuum vision.

\subsection{Understanding Performance of Application Workflows on the Edge-to-Cloud Continuum}

Understanding end-to-end performance on the complex Edge-to-Cloud heterogeneous ecosystem is challenging. Deploying large-scale real-life applications on such infrastructures requires configuring a myriad of system-specific parameters and reconciling many requirements or constraints in terms of hardware capacity, mobility, network efficiency, energy, and data privacy, with low-level infrastructure design choices. One important challenge is to accurately reproduce relevant behaviors of a given application workflow and representative settings of the physical infrastructure underlying this complex continuum. 

A first step towards reducing this complexity and enabling the Computing Continuum vision is to enable a \textbf{holistic understanding of performance} in such environments. That is, finding a rigorous approach to answering questions like: \emph{(1) How to identify infrastructure bottlenecks across the whole Edge-to-Cloud Continuum? (2) Which system parameters and network configurations impact on the application performance and how? (3) How Edge-to-Cloud hardware configurations impact on the energy consumption and on the processing latency of the application?}

Approaches based on workflow modeling \cite{sadiq2004data} and simulation or emulation, as presented in Table~\ref{tbl:sed-exp}, raise some important challenges in terms of specification, modeling, and validation in the context of the Computing Continuum~\cite{abreu2019comparative, svorobej2019simulating}. For example, it is increasingly difficult to model the heterogeneity and volatility of Edge devices or to assess the impact of the inherent complexity of hybrid Edge-Cloud deployments on performance. At this stage, \textbf{experimental evaluation} remains the main approach to gain accurate insights on performance metrics and to \textbf{build precise approximations} of the expected behavior of large-scale applications on the Computing Continuum, as a \textbf{first step prior to modeling}.

A key challenge in this context is to be able to \textbf{reproduce in a representative way the application behavior in a controlled environment}, for extensive experiments in a large-enough spectrum of potential configurations of the underlying hybrid Edge-Cloud infrastructure. However, this process is non-trivial due to the multiple combination possibilities of heterogeneous hardware and software resources, as well as, system components for Data Analytics and Machine Learning. Therefore, the Computing Continuum vision calls for novel approaches to \textbf{map the real-world application components and dependencies to infrastructure resources}.

Further research efforts shall necessarily focus on the design and implementation of novel methodologies and systems for large-scale experimental evaluation covering the characteristics of hybrid Edge-Cloud infrastructure deployments. Novel systems allowing \textbf{the combination of simulation and emulation} systems in addition to supporting \textbf{the deployment of state-of-the-art systems} for Data Analytics and Machine Learning \textbf{on real-world large-scale testbeds}, considering the same experimental evaluation package, would be relevant to accurately reproduce complex application behaviors.


\subsection{Optimizing the Performance of Edge-to-Cloud Application Workflows}

The optimization of application workflows on highly distributed and heterogeneous resources is challenging. Real-world applications deployed on hybrid Edge-to-Cloud infrastructures (\emph{e.g.,} smart factory~\cite{wang2018adaptive}, autonomous vehicles~\cite{midya2018multi}, among others) typically need to comply with many \textbf{conflicting constraints} related to hardware resource consumption (\emph{e.g.,} GPU memory, CPU power, main memory size, storage size and bandwidth), software components composing the application and requirements such as QoS, security, and privacy~\cite{xia2018combining}. 


Furthermore, Edge-to-Cloud deployment optimization problems aim at \textbf{optimizing metrics}~\cite{bellendorf2020classification, aslanpour2020performance} related to performance (\emph{e.g.}, execution time, latency, and throughput), resource usage, energy consumption, financial costs, and quality attributes (\emph{e.g.}, reliability, security, and privacy). Therefore, the parameter settings of the applications and the underlying infrastructure result in a complex multi-infrastructure configuration search space~\cite{ranjan2018next}.

Therefore, one important challenge is to accurately and efficiently answer questions like: \emph{(1) How to configure the hardware and system components to minimize processing latency and energy consumption?} \emph{(2) Where should the workflow components be executed across the Edge-to-Cloud Continuum to minimize communication costs and end-to-end latency?} \textcolor{blue}{\emph{(3) How to efficiently autoscale the application resources concerning workload fluctuations and infrastructure changes?}} 

Such optimization problems are of NP-hard complexity and multi-objective. Furthermore, the environment settings and configuration parameters are extremely vast and their combination of possibilities virtually unlimited~\cite{sharma2017live, xu2018survey}. Hence, the process of searching the ideal deployment and configuration of those real-life applications is challenging given the search space complexity: bad choices may result in increased financial expenses during deployment and production phases, decreased processing efficiency and poor user experience~\cite{verma2020smart}. 

Given these complexities, future research should focus on proposing \textbf{novel optimization methodologies} supporting the parallel deployment and evaluation of such complex application workflows on real-life large scale testbeds. The objective is two fold: speeding up the optimization computations, as well as obtaining more accurate results.  

Novel approaches should also rely on the development of \textbf{fully automated surrogate model building} to mimic and approximate the complex behavior of Edge-to-Cloud workflows and then perform \textbf{optimization and sensitivity analysis}. These new solutions may combine computationally tractable optimization techniques~\cite{pham2012intelligent} such as Bayesian Optimization~\cite{snoek2012practical} methods (\emph{e.g.,} Gaussian process (Kriging)~\cite{simpson2001kriging}, Decision Trees~\cite{wang2000optimization}, Random Forest~\cite{breiman2001random}, among others) to build surrogate models; and then combine with techniques such as evolutionary algorithms and swarm intelligence based algorithms (\emph{e.g.,} Genetic Algorithm~\cite{mirjalili2019genetic}, Differential Evolution~\cite{das2016recent}, Particle Swarm Optimization~\cite{du2016particle}, \emph{etc.}) to perform and speed up the optimization (\emph{e.g.,} to find the optimal deployment configuration using the built surrogate model).



\textcolor{blue}{Novel contributions are required for workload characterization and prediction, for autoscaling strategies to enable the efficient scaling of distributed application resources across the Edge-to-Cloud continuum, in response to workload fluctuations and infrastructure changes. Contributions in this context, should be aligned to the complex heterogeneous characteristics of the Computing Continuum paradigm, in terms of: computing resources; network constraints; and application requirements.}

\subsection{Enabling Intelligence on the Highly Heterogeneous Edge-to-Cloud Continuum}

The right selection of Machine Learning techniques for fast and accurate decision making on the highly heterogeneous (in terms of hardware and software) Edge-to-Cloud Continuum  requires extensive experiments and evaluations on real-life hybrid infrastructures combining HPC, Cloud, and Edge systems.

The goal is to understand how: (a) infrastructure design choices, (b) optimized learning algorithms with tunable parameters, and (c) the combination of learning paradigms impact on performance metrics such as memory usage, energy consumption, model accuracy, training time, network overhead, application processing latency, among others~\cite{rocha2020distributed}.

This comes down to answering questions like: \emph{(1) How to efficiently deploy complex AI workflows on heterogeneous and distributed infrastructures to reduce training time and improve model accuracy? (2) How to combine Machine Learning paradigms to leverage the massively distributed resources for training across the Edge-to-Cloud Continuum?}

A relevant challenge, worthy of further consideration, is \textbf{to understand the performance trade-offs at scale of combining a variety of learning paradigms} such as Reinforcement Learning~\cite{wiering2012reinforcement}, Deep Learning~\cite{goodfellow2016deep}, Online Learning~\cite{diez2019data}, Stream Learning~\cite{lobo2020spiking}, Lifelong Learning~\cite{fei2019cps}, Transfer Learning~\cite{diez2019data}, Federated Learning~\cite{aral2020staleness}, Distributed Learning~\cite{yousefpour2019all, zhang2020deep}, Multi-task Learning~\cite{zhang2017survey}, and others.

Approaches leveraging the \textbf{incremental evolution of models over time} (\emph{e.g.,} instead of reconstructing new models from scratch) should be considered for streaming data (\emph{e.g.,} instead of batch learning, where the whole training data set should be available for training). They are useful for applications that require high speed processing and analysis of data, and also to avoid the concept drift problem, where predictions become less accurate as time passes, or in cases where the accumulation of large volumes of data is impractical (\emph{e.g.,} due to memory, storage, and processing limitations of Edge devices)~\cite{xu2018survey}. 

Novel approaches should also leverage \textbf{the transfer of knowledge to/from different domains} (\emph{e.g.,} useful when data for training is scarce) and also take advantage of the parallelism and scalability provided by state-of-the-art distributed stream processing systems (\emph{e.g.,} Flink, Spark, \emph{etc.}) combined with Machine Learning paradigms~\cite{fei2019cps} in order to speed up the training and inference time.

Other open challenges~\cite{nikouei2019toward, loghin2020disruptions, prabhu2019fog} include: exploring the massively distributed Edge devices for AI training to achieve scalable and distributed deployment of models on Edge-to-Cloud infrastructures; applying Neural Architecture Search~\cite{elsken2019neural} and Hyperparameter Search~\cite{claesen2015hyperparameter} to obtain Deep Learning networks that require less resource without losing accuracy; and exploring Knowledge Distillation~\cite{chen2019deep} (\emph{i.e.,} transferring knowledge from a large model to a smaller model without loss of validity) to leverage model deployment on resource-limited devices.

Lastly, further research is needed on novel approaches proposing rigorous methodologies and systems for reproducible experimental evaluations to enable \textbf{the performance comparison of AI models and learning paradigms} deployed on large scale and heterogeneous Edge-to-Cloud infrastructures. Such approaches should publish the experimental artifacts on public repositories to allow their reproducibility~\cite{chen2019deep}.

These directions are still ongoing and active research areas in the Big Data and AI communities, and as presented in this systematic review, we have not seen reported studies exploring such challenges at large scale on hybrid Edge-to-Cloud infrastructures.

\subsection{Supporting Reproducible Analysis of Complex Edge-to-Cloud Workflows}


Given the relevance of experimental reproducibility in scientific research to allow the verification of the scientific claims and also to evolve the studies, in addition to the lack of support to the reproducibility of experiments identified in recent articles, as presented in Subsection~\ref{subsec:supp-rep}, future research efforts should focus on the design and implementation of rigorous methodologies for experimental reproducibility.

Supporting reproducibility of experiments carried out on large scale distributed and heterogeneous infrastructures is non-trivial. The experimental methodology, the artifacts used, and the data captured should provide additional context that more accurately explains the experiment execution and results.

One relevant challenge is to provide mechanisms to allow researchers to \textbf{repeat}, \textbf{replicate}, and \textbf{reproduce} the scientific claims and to help them answer questions like: \emph{(1) What machines/devices were used to execute the entire workflow? (2) What steps were invoked during the workflow execution? (3) Which infrastructure configurations and application parameters produced these results?}

Therefore, novel approaches should focus on enabling the repeatability, replicability and reproducibility of experiments. This requires the \textbf{definition of rigorous experimentation methodologies} (\emph{e.g.} well-defined description of: hardware and software resources required to run the experiments and their configurations, network setups, resource interconnections, and workflow execution logic); the \textbf{access to the experimental artifacts} (\emph{e.g.} datasets, scripts, libraries, applications, systems, configuration files, among others); and the development of \textbf{mechanisms to automatically manage the data derived from experiments}, including: the data provenance capture (\emph{e.g.} runtime configuration of physical machines, software and systems setups, network configurations, \emph{etc.}) and the access to results (\emph{e.g.} log files, metrics collected during execution, monitoring data, code to plot results, among others).

In particular, an important challenge is \textbf{the data provenance} capture on such highly heterogeneous and distributed infrastructures. It requires the design and development of novel provenance systems to efficiently capture data from heterogeneous hardware resources ranging from HPC/Cloud servers to resource constrained Edge devices (\emph{e.g.} requires smart data capture strategies to reduce capture overhead) interconnected by different network capabilities (\emph{e.g.} requires provenance data transmission balancing to mitigate the network overhead).


\section{Conclusions}
\label{sec:conclusions}
In this paper, we did a systematic review of the current state-of-the-art methods to enable intelligence on the Edge-to-Cloud Continuum.
First, we discussed the main libraries and frameworks for Machine Learning and Deep Learning inference, centralized training, and distributed training with a focus on the Edge and Cloud. We also presented the main methods for data processing on the Edge, as well as the methods for Big Data stream analytics across the Edge-to-Cloud Continuum.

We reviewed the recent systems, frameworks and architectures that combine Machine Learning and Data Analytics through the main state-of-the-art learning paradigms such as Online Learning, Transfer Learning, Federated Learning, among others, for collaborative Edge-to-Cloud training and decision making.
Finally, we discussed experimental research that covers the whole Edge-to-Cloud Continuum with a focus on simulation, emulation, and deployment systems, as well as large-scale experimental testbeds and how the studies included in our systematic review provide support for experiment reproducibility. 

There are several open challenges left to realize the Computing Continuum vision. We highlighted the complexity of a holistic understanding of performance of application workflows deployed on the Continuum, as well as the performance optimization of such applications on highly distributed and heterogeneous environments. Many advances are yet required to enable intelligence across the Edge-to-Cloud Continuum in an efficient way.
In particular, there is a need for approaches that allow the optimized deployment of complex AI workflows to reduce training time and improve model accuracy, and novel ideas that leverage the massively distributed Edge-to-Cloud resources for fast decision making. Given the lack of support for reproducibility, there is also a need for approaches that help scientists to repeat, replicate, and reproduce the analysis of complex Edge-to-Cloud workflows in a large scale. These challenges can be addressed through novel methodologies, algorithms, systems and frameworks.

\section*{Acknowledgments}
This work was funded by Inria through the HPC-BigData Inria Challenge (IPL) and by French ANR OverFlow project (ANR-15- CE25-0003).



\printcredits

\bibliographystyle{cas-model2-names}

\bibliography{cas-refs}

\begin{thebibliography}{170}
\expandafter\ifx\csname natexlab\endcsname\relax\def\natexlab#1{#1}\fi
\providecommand{\url}[1]{\texttt{#1}}
\providecommand{\href}[2]{#2}
\providecommand{\path}[1]{#1}
\providecommand{\DOIprefix}{doi:}
\providecommand{\ArXivprefix}{arXiv:}
\providecommand{\URLprefix}{URL: }
\providecommand{\Pubmedprefix}{pmid:}
\providecommand{\doi}[1]{\href{http://dx.doi.org/#1}{\path{#1}}}
\providecommand{\Pubmed}[1]{\href{pmid:#1}{\path{#1}}}
\providecommand{\bibinfo}[2]{#2}
\ifx\xfnm\relax \def\xfnm[#1]{\unskip,\space#1}\fi
\bibitem[{Abreu et~al.(2019)Abreu, Velasquez, Curado and
  Monteiro}]{abreu2019comparative}
\bibinfo{author}{Abreu, D.P.}, \bibinfo{author}{Velasquez, K.},
  \bibinfo{author}{Curado, M.}, \bibinfo{author}{Monteiro, E.},
  \bibinfo{year}{2019}.
\newblock \bibinfo{title}{{A Comparative Analysis of Simulators for the Cloud
  to Fog Continuum}}.
\newblock \bibinfo{journal}{Simulation Modelling Practice and Theory} ,
  \bibinfo{pages}{102029}.
\bibitem[{Adjih et~al.(2015)Adjih, Baccelli, Fleury, Harter, Mitton, Noel,
  Pissard-Gibollet, Saint-Marcel, Schreiner, Vandaele et~al.}]{adjih2015fit}
\bibinfo{author}{Adjih, C.}, \bibinfo{author}{Baccelli, E.},
  \bibinfo{author}{Fleury, E.}, \bibinfo{author}{Harter, G.},
  \bibinfo{author}{Mitton, N.}, \bibinfo{author}{Noel, T.},
  \bibinfo{author}{Pissard-Gibollet, R.}, \bibinfo{author}{Saint-Marcel, F.},
  \bibinfo{author}{Schreiner, G.}, \bibinfo{author}{Vandaele, J.}, et~al.,
  \bibinfo{year}{2015}.
\newblock \bibinfo{title}{Fit iot-lab: A large scale open experimental iot
  testbed}, in: \bibinfo{booktitle}{2015 IEEE 2nd World Forum on Internet of
  Things (WF-IoT)}, \bibinfo{organization}{IEEE}. pp.
  \bibinfo{pages}{459--464}.
\bibitem[{Alencar et~al.(2020)Alencar, Rios, Santana and
  Prazeres}]{alencar2020fot}
\bibinfo{author}{Alencar, B.M.}, \bibinfo{author}{Rios, R.A.},
  \bibinfo{author}{Santana, C.}, \bibinfo{author}{Prazeres, C.},
  \bibinfo{year}{2020}.
\newblock \bibinfo{title}{Fot-stream: A fog platform for data stream analytics
  in iot}.
\newblock \bibinfo{journal}{Computer Communications} .
\bibitem[{Ali and Abdullah(2018)}]{ali2018recent}
\bibinfo{author}{Ali, A.H.}, \bibinfo{author}{Abdullah, M.Z.},
  \bibinfo{year}{2018}.
\newblock \bibinfo{title}{Recent trends in distributed online stream processing
  platform for big data: Survey}, in: \bibinfo{booktitle}{2018 1st Annual
  International Conference on Information and Sciences (AiCIS)},
  \bibinfo{organization}{IEEE}. pp. \bibinfo{pages}{140--145}.
\bibitem[{Ali et~al.(2020)Ali, Anjum, Rana, Zamani, Balouek-Thomert and
  Parashar}]{ali2020res}
\bibinfo{author}{Ali, M.}, \bibinfo{author}{Anjum, A.}, \bibinfo{author}{Rana,
  O.}, \bibinfo{author}{Zamani, A.R.}, \bibinfo{author}{Balouek-Thomert, D.},
  \bibinfo{author}{Parashar, M.}, \bibinfo{year}{2020}.
\newblock \bibinfo{title}{Res: Real-time video stream analytics using edge
  enhanced clouds}.
\newblock \bibinfo{journal}{IEEE Transactions on Cloud Computing} .
\bibitem[{Alli and Alam(2020)}]{alli2020fog}
\bibinfo{author}{Alli, A.A.}, \bibinfo{author}{Alam, M.M.},
  \bibinfo{year}{2020}.
\newblock \bibinfo{title}{The fog cloud of things: A survey on concepts,
  architecture, standards, tools, and applications}.
\newblock \bibinfo{journal}{Internet of Things} \bibinfo{volume}{9},
  \bibinfo{pages}{100177}.
\bibitem[{Angel et~al.(2022)Angel, Ravindran, Vincent, Srinivasan and
  Hu}]{angel2022recent}
\bibinfo{author}{Angel, N.A.}, \bibinfo{author}{Ravindran, D.},
  \bibinfo{author}{Vincent, P.}, \bibinfo{author}{Srinivasan, K.},
  \bibinfo{author}{Hu, Y.C.}, \bibinfo{year}{2022}.
\newblock \bibinfo{title}{Recent advances in evolving computing paradigms:
  Cloud, edge, and fog technologies}.
\newblock \bibinfo{journal}{Sensors} \bibinfo{volume}{22},
  \bibinfo{pages}{196}.
\bibitem[{Ansari et~al.(2020)Ansari, Alsamhi, Qiao, Ye and
  Lee}]{ansari2020security}
\bibinfo{author}{Ansari, M.S.}, \bibinfo{author}{Alsamhi, S.H.},
  \bibinfo{author}{Qiao, Y.}, \bibinfo{author}{Ye, Y.}, \bibinfo{author}{Lee,
  B.}, \bibinfo{year}{2020}.
\newblock \bibinfo{title}{Security of distributed intelligence in edge
  computing: Threats and countermeasures}, in: \bibinfo{booktitle}{The
  cloud-to-thing continuum}. \bibinfo{publisher}{Palgrave Macmillan, Cham}, pp.
  \bibinfo{pages}{95--122}.
\bibitem[{Aral et~al.(2020)Aral, Erol-Kantarci and
  Brandi{\'c}}]{aral2020staleness}
\bibinfo{author}{Aral, A.}, \bibinfo{author}{Erol-Kantarci, M.},
  \bibinfo{author}{Brandi{\'c}, I.}, \bibinfo{year}{2020}.
\newblock \bibinfo{title}{Staleness control for edge data analytics}.
\newblock \bibinfo{journal}{Proceedings of the ACM on Measurement and Analysis
  of Computing Systems} \bibinfo{volume}{4}, \bibinfo{pages}{1--24}.
\bibitem[{Asch et~al.(2018)Asch, Moore, Badia, Beck, Beckman, Bidot, Bodin,
  Cappello, Choudhary, de~Supinski et~al.}]{asch2018big}
\bibinfo{author}{Asch, M.}, \bibinfo{author}{Moore, T.},
  \bibinfo{author}{Badia, R.}, \bibinfo{author}{Beck, M.},
  \bibinfo{author}{Beckman, P.}, \bibinfo{author}{Bidot, T.},
  \bibinfo{author}{Bodin, F.}, \bibinfo{author}{Cappello, F.},
  \bibinfo{author}{Choudhary, A.}, \bibinfo{author}{de~Supinski, B.}, et~al.,
  \bibinfo{year}{2018}.
\newblock \bibinfo{title}{Big data and extreme-scale computing: Pathways to
  convergence - toward a shaping strategy for a future software and data
  ecosystem for scientific inquiry}.
\newblock \bibinfo{journal}{The International Journal of High Performance
  Computing Applications} \bibinfo{volume}{32}, \bibinfo{pages}{435--479}.
\bibitem[{Aslanpour et~al.(2020)Aslanpour, Gill and
  Toosi}]{aslanpour2020performance}
\bibinfo{author}{Aslanpour, M.S.}, \bibinfo{author}{Gill, S.S.},
  \bibinfo{author}{Toosi, A.N.}, \bibinfo{year}{2020}.
\newblock \bibinfo{title}{Performance evaluation metrics for cloud, fog and
  edge computing: A review, taxonomy, benchmarks and standards for future
  research}.
\newblock \bibinfo{journal}{Internet of Things} , \bibinfo{pages}{100273}.
\bibitem[{Assefi et~al.(2017)Assefi, Behravesh, Liu and Tafti}]{assefi2017big}
\bibinfo{author}{Assefi, M.}, \bibinfo{author}{Behravesh, E.},
  \bibinfo{author}{Liu, G.}, \bibinfo{author}{Tafti, A.P.},
  \bibinfo{year}{2017}.
\newblock \bibinfo{title}{Big data machine learning using apache spark mllib},
  in: \bibinfo{booktitle}{2017 ieee international conference on big data (big
  data)}, \bibinfo{organization}{IEEE}. pp. \bibinfo{pages}{3492--3498}.
\bibitem[{Atitallah et~al.(2020)Atitallah, Driss, Boulila and
  Gh{\'e}zala}]{atitallah2020leveraging}
\bibinfo{author}{Atitallah, S.B.}, \bibinfo{author}{Driss, M.},
  \bibinfo{author}{Boulila, W.}, \bibinfo{author}{Gh{\'e}zala, H.B.},
  \bibinfo{year}{2020}.
\newblock \bibinfo{title}{Leveraging deep learning and iot big data analytics
  to support the smart cities development: Review and future directions}.
\newblock \bibinfo{journal}{Computer Science Review} \bibinfo{volume}{38},
  \bibinfo{pages}{100303}.
\bibitem[{Badidi et~al.(2020)Badidi, Mahrez and Sabir}]{badidi2020fog}
\bibinfo{author}{Badidi, E.}, \bibinfo{author}{Mahrez, Z.},
  \bibinfo{author}{Sabir, E.}, \bibinfo{year}{2020}.
\newblock \bibinfo{title}{Fog computing for smart cities’ big data management
  and analytics: A review}.
\newblock \bibinfo{journal}{Future Internet} \bibinfo{volume}{12},
  \bibinfo{pages}{190}.
\bibitem[{Baldin et~al.(2016)Baldin, Chase, Xin, Mandal, Ruth, Castillo,
  Orlikowski, Heermann and Mills}]{baldin2016exogeni}
\bibinfo{author}{Baldin, I.}, \bibinfo{author}{Chase, J.},
  \bibinfo{author}{Xin, Y.}, \bibinfo{author}{Mandal, A.},
  \bibinfo{author}{Ruth, P.}, \bibinfo{author}{Castillo, C.},
  \bibinfo{author}{Orlikowski, V.}, \bibinfo{author}{Heermann, C.},
  \bibinfo{author}{Mills, J.}, \bibinfo{year}{2016}.
\newblock \bibinfo{title}{Exogeni: A multi-domain infrastructure-as-a-service
  testbed}, in: \bibinfo{booktitle}{The GENI Book}.
  \bibinfo{publisher}{Springer}, pp. \bibinfo{pages}{279--315}.
\bibitem[{{Barba} and {Thiruvathukal}(2017)}]{ieee-computing}
\bibinfo{author}{{Barba}, L.A.}, \bibinfo{author}{{Thiruvathukal}, G.K.},
  \bibinfo{year}{2017}.
\newblock \bibinfo{title}{{Reproducible Research for Computing in Science
  Engineering}}.
\newblock \bibinfo{journal}{Computing in Science Engineering}
  \bibinfo{volume}{19}, \bibinfo{pages}{85--87}.
\bibitem[{Bellendorf and Mann(2020)}]{bellendorf2020classification}
\bibinfo{author}{Bellendorf, J.}, \bibinfo{author}{Mann, Z.{\'A}.},
  \bibinfo{year}{2020}.
\newblock \bibinfo{title}{Classification of optimization problems in fog
  computing}.
\newblock \bibinfo{journal}{Future Generation Computer Systems}
  \bibinfo{volume}{107}, \bibinfo{pages}{158--176}.
\bibitem[{Bendechache et~al.(2020)Bendechache, Svorobej, Takako~Endo and
  Lynn}]{bendechache2020simulating}
\bibinfo{author}{Bendechache, M.}, \bibinfo{author}{Svorobej, S.},
  \bibinfo{author}{Takako~Endo, P.}, \bibinfo{author}{Lynn, T.},
  \bibinfo{year}{2020}.
\newblock \bibinfo{title}{Simulating resource management across the
  cloud-to-thing continuum: A survey and future directions}.
\newblock \bibinfo{journal}{Future Internet} \bibinfo{volume}{12},
  \bibinfo{pages}{95}.
\bibitem[{Bhat and Huang(2021)}]{bhat2021big}
\bibinfo{author}{Bhat, S.A.}, \bibinfo{author}{Huang, N.F.},
  \bibinfo{year}{2021}.
\newblock \bibinfo{title}{Big data and ai revolution in precision agriculture:
  Survey and challenges}.
\newblock \bibinfo{journal}{IEEE Access} .
\bibitem[{Bolze et~al.(2006)Bolze, Cappello, Caron, Dayde, Desprez, Jeannot,
  J{\'e}gou, Lanteri, Leduc, Melab, Mornet, Namyst, Primet, Qu{\'e}tier,
  Richard, Talbi and Touche}]{bolze:hal-00684943}
\bibinfo{author}{Bolze, R.}, \bibinfo{author}{Cappello, F.},
  \bibinfo{author}{Caron, E.}, \bibinfo{author}{Dayde, M.},
  \bibinfo{author}{Desprez, F.}, \bibinfo{author}{Jeannot, E.},
  \bibinfo{author}{J{\'e}gou, Y.}, \bibinfo{author}{Lanteri, S.},
  \bibinfo{author}{Leduc, J.}, \bibinfo{author}{Melab, N.},
  \bibinfo{author}{Mornet, G.}, \bibinfo{author}{Namyst, R.},
  \bibinfo{author}{Primet, P.}, \bibinfo{author}{Qu{\'e}tier, B.},
  \bibinfo{author}{Richard, O.}, \bibinfo{author}{Talbi, E.G.},
  \bibinfo{author}{Touche, I.}, \bibinfo{year}{2006}.
\newblock \bibinfo{title}{{Grid'5000: A Large Scale And Highly Reconfigurable
  Experimental Grid Testbed}}.
\newblock \bibinfo{journal}{{International Journal of High Performance
  Computing Applications}} \bibinfo{volume}{20}, \bibinfo{pages}{481--494}.
\newblock \URLprefix \url{https://hal.inria.fr/hal-00684943},
  \DOIprefix\doi{10.1177/1094342006070078}.
\bibitem[{Bonawitz et~al.(2019)Bonawitz, Eichner, Grieskamp, Huba, Ingerman,
  Ivanov, Kiddon, Kone{\v{c}}n{\`y}, Mazzocchi, McMahan
  et~al.}]{bonawitz2019towards}
\bibinfo{author}{Bonawitz, K.}, \bibinfo{author}{Eichner, H.},
  \bibinfo{author}{Grieskamp, W.}, \bibinfo{author}{Huba, D.},
  \bibinfo{author}{Ingerman, A.}, \bibinfo{author}{Ivanov, V.},
  \bibinfo{author}{Kiddon, C.}, \bibinfo{author}{Kone{\v{c}}n{\`y}, J.},
  \bibinfo{author}{Mazzocchi, S.}, \bibinfo{author}{McMahan, H.B.}, et~al.,
  \bibinfo{year}{2019}.
\newblock \bibinfo{title}{Towards federated learning at scale: System design}.
\newblock \bibinfo{journal}{arXiv preprint arXiv:1902.01046} .
\bibitem[{Bouckaert et~al.(2010)Bouckaert, Vandenberghe, Jooris, Moerman and
  Demeester}]{bouckaert2010w}
\bibinfo{author}{Bouckaert, S.}, \bibinfo{author}{Vandenberghe, W.},
  \bibinfo{author}{Jooris, B.}, \bibinfo{author}{Moerman, I.},
  \bibinfo{author}{Demeester, P.}, \bibinfo{year}{2010}.
\newblock \bibinfo{title}{The w-ilab. t testbed}, in:
  \bibinfo{booktitle}{International Conference on Testbeds and Research
  Infrastructures}, \bibinfo{organization}{Springer}. pp.
  \bibinfo{pages}{145--154}.
\bibitem[{Breiman(2001)}]{breiman2001random}
\bibinfo{author}{Breiman, L.}, \bibinfo{year}{2001}.
\newblock \bibinfo{title}{Random forests}.
\newblock \bibinfo{journal}{Machine learning} \bibinfo{volume}{45},
  \bibinfo{pages}{5--32}.
\bibitem[{Brewer(2015)}]{brewer2015kubernetes}
\bibinfo{author}{Brewer, E.A.}, \bibinfo{year}{2015}.
\newblock \bibinfo{title}{Kubernetes and the path to cloud native}, in:
  \bibinfo{booktitle}{Proceedings of the sixth ACM symposium on cloud
  computing}, pp. \bibinfo{pages}{167--167}.
\bibitem[{Brogi and Forti(2017)}]{fogtorch}
\bibinfo{author}{Brogi, A.}, \bibinfo{author}{Forti, S.}, \bibinfo{year}{2017}.
\newblock \bibinfo{title}{Qos-aware deployment of iot applications through the
  fog}.
\newblock \bibinfo{journal}{IEEE Internet of Things Journal}
  \bibinfo{volume}{4}, \bibinfo{pages}{1185--1192}.
\newblock \DOIprefix\doi{10.1109/JIOT.2017.2701408}.
\bibitem[{Cai et~al.(2017)Cai, Li and Li}]{cai2017elasticsim}
\bibinfo{author}{Cai, Z.}, \bibinfo{author}{Li, Q.}, \bibinfo{author}{Li, X.},
  \bibinfo{year}{2017}.
\newblock \bibinfo{title}{Elasticsim: A toolkit for simulating workflows with
  cloud resource runtime auto-scaling and stochastic task execution times}.
\newblock \bibinfo{journal}{Journal of Grid Computing} \bibinfo{volume}{15},
  \bibinfo{pages}{257--272}.
\bibitem[{Caida(July 28, 2021)}]{caidanet}
\bibinfo{author}{Caida}, \bibinfo{year}{July 28, 2021}.
\newblock \bibinfo{title}{About caida}.
\newblock \bibinfo{howpublished}{\url{https://www.caida.org/about/}}.
\bibitem[{Calheiros et~al.(2011)Calheiros, Ranjan, Beloglazov, De~Rose and
  Buyya}]{calheiros2011cloudsim}
\bibinfo{author}{Calheiros, R.N.}, \bibinfo{author}{Ranjan, R.},
  \bibinfo{author}{Beloglazov, A.}, \bibinfo{author}{De~Rose, C.A.},
  \bibinfo{author}{Buyya, R.}, \bibinfo{year}{2011}.
\newblock \bibinfo{title}{Cloudsim: a toolkit for modeling and simulation of
  cloud computing environments and evaluation of resource provisioning
  algorithms}.
\newblock \bibinfo{journal}{Software: Practice and experience}
  \bibinfo{volume}{41}, \bibinfo{pages}{23--50}.
\bibitem[{Chao et~al.(2019)Chao, Peng, Xu and Zhang}]{chao2019ecosystem}
\bibinfo{author}{Chao, L.}, \bibinfo{author}{Peng, X.}, \bibinfo{author}{Xu,
  Z.}, \bibinfo{author}{Zhang, L.}, \bibinfo{year}{2019}.
\newblock \bibinfo{title}{Ecosystem of things: Hardware, software, and
  architecture}.
\newblock \bibinfo{journal}{Proceedings of the IEEE} \bibinfo{volume}{107},
  \bibinfo{pages}{1563--1583}.
\bibitem[{Chen and Ran(2019)}]{chen2019deep}
\bibinfo{author}{Chen, J.}, \bibinfo{author}{Ran, X.}, \bibinfo{year}{2019}.
\newblock \bibinfo{title}{Deep learning with edge computing: A review.}
\newblock \bibinfo{journal}{Proceedings of the IEEE} \bibinfo{volume}{107},
  \bibinfo{pages}{1655--1674}.
\bibitem[{Chen et~al.(2019)Chen, Zhao, Li and Zhao}]{chen2019exploring}
\bibinfo{author}{Chen, Y.}, \bibinfo{author}{Zhao, K.}, \bibinfo{author}{Li,
  B.}, \bibinfo{author}{Zhao, M.}, \bibinfo{year}{2019}.
\newblock \bibinfo{title}{Exploring the use of synthetic gradients for
  distributed deep learning across cloud and edge resources}, in:
  \bibinfo{booktitle}{2nd $\{$USENIX$\}$ Workshop on Hot Topics in Edge
  Computing (HotEdge 19)}.
\bibitem[{Cherrueau et~al.(2021)Cherrueau, Delavergne, van Kempen, Lebre,
  Pertin, Balderrama, Simonet and Simonin}]{cherrueau2021enoslib}
\bibinfo{author}{Cherrueau, R.A.}, \bibinfo{author}{Delavergne, M.},
  \bibinfo{author}{van Kempen, A.}, \bibinfo{author}{Lebre, A.},
  \bibinfo{author}{Pertin, D.}, \bibinfo{author}{Balderrama, J.R.},
  \bibinfo{author}{Simonet, A.}, \bibinfo{author}{Simonin, M.},
  \bibinfo{year}{2021}.
\newblock \bibinfo{title}{Enoslib: A library for experiment-driven research in
  distributed computing}.
\newblock \bibinfo{journal}{IEEE Transactions on Parallel and Distributed
  Systems} .
\bibitem[{Claesen and De~Moor(2015)}]{claesen2015hyperparameter}
\bibinfo{author}{Claesen, M.}, \bibinfo{author}{De~Moor, B.},
  \bibinfo{year}{2015}.
\newblock \bibinfo{title}{Hyperparameter search in machine learning}.
\newblock \bibinfo{journal}{arXiv preprint arXiv:1502.02127} .
\bibitem[{Coutinho et~al.(2018)Coutinho, Greve, Prazeres and
  Cardoso}]{coutinho2018fogbed}
\bibinfo{author}{Coutinho, A.}, \bibinfo{author}{Greve, F.},
  \bibinfo{author}{Prazeres, C.}, \bibinfo{author}{Cardoso, J.},
  \bibinfo{year}{2018}.
\newblock \bibinfo{title}{Fogbed: A rapid-prototyping emulation environment for
  fog computing}, in: \bibinfo{booktitle}{2018 IEEE International Conference on
  Communications (ICC)}, \bibinfo{organization}{IEEE}. pp.
  \bibinfo{pages}{1--7}.
\bibitem[{Das et~al.(2016)Das, Mullick and Suganthan}]{das2016recent}
\bibinfo{author}{Das, S.}, \bibinfo{author}{Mullick, S.S.},
  \bibinfo{author}{Suganthan, P.N.}, \bibinfo{year}{2016}.
\newblock \bibinfo{title}{Recent advances in differential evolution--an updated
  survey}.
\newblock \bibinfo{journal}{Swarm and Evolutionary Computation}
  \bibinfo{volume}{27}, \bibinfo{pages}{1--30}.
\bibitem[{Dautov and Distefano(2020)}]{dautov2020stream}
\bibinfo{author}{Dautov, R.}, \bibinfo{author}{Distefano, S.},
  \bibinfo{year}{2020}.
\newblock \bibinfo{title}{Stream processing on clustered edge devices}.
\newblock \bibinfo{journal}{IEEE Transactions on Cloud Computing} .
\bibitem[{Dautov et~al.(2017)Dautov, Distefano, Bruneo, Longo, Merlino and
  Puliafito}]{dautov2017pushing}
\bibinfo{author}{Dautov, R.}, \bibinfo{author}{Distefano, S.},
  \bibinfo{author}{Bruneo, D.}, \bibinfo{author}{Longo, F.},
  \bibinfo{author}{Merlino, G.}, \bibinfo{author}{Puliafito, A.},
  \bibinfo{year}{2017}.
\newblock \bibinfo{title}{Pushing intelligence to the edge with a stream
  processing architecture}, in: \bibinfo{booktitle}{2017 IEEE International
  Conference on Internet of Things (iThings) and IEEE Green Computing and
  Communications (GreenCom) and IEEE Cyber, Physical and Social Computing
  (CPSCom) and IEEE Smart Data (SmartData)}, \bibinfo{organization}{IEEE}. pp.
  \bibinfo{pages}{792--799}.
\bibitem[{Dautov et~al.(2018)Dautov, Distefano, Bruneo, Longo, Merlino and
  Puliafito}]{dautov2018data}
\bibinfo{author}{Dautov, R.}, \bibinfo{author}{Distefano, S.},
  \bibinfo{author}{Bruneo, D.}, \bibinfo{author}{Longo, F.},
  \bibinfo{author}{Merlino, G.}, \bibinfo{author}{Puliafito, A.},
  \bibinfo{year}{2018}.
\newblock \bibinfo{title}{Data processing in cyber-physical-social systems
  through edge computing}.
\newblock \bibinfo{journal}{IEEE Access} \bibinfo{volume}{6},
  \bibinfo{pages}{29822--29835}.
\bibitem[{Dautov et~al.(2019)Dautov, Distefano and
  Buyya}]{dautov2019hierarchical}
\bibinfo{author}{Dautov, R.}, \bibinfo{author}{Distefano, S.},
  \bibinfo{author}{Buyya, R.}, \bibinfo{year}{2019}.
\newblock \bibinfo{title}{Hierarchical data fusion for smart healthcare}.
\newblock \bibinfo{journal}{Journal of Big Data} \bibinfo{volume}{6},
  \bibinfo{pages}{1--23}.
\bibitem[{Debauche et~al.(2021)Debauche, Mahmoudi, Manneback and
  Lebeau}]{debauche2021cloud}
\bibinfo{author}{Debauche, O.}, \bibinfo{author}{Mahmoudi, S.},
  \bibinfo{author}{Manneback, P.}, \bibinfo{author}{Lebeau, F.},
  \bibinfo{year}{2021}.
\newblock \bibinfo{title}{Cloud and distributed architectures for data
  management in agriculture 4.0: Review and future trends}.
\newblock \bibinfo{journal}{Journal of King Saud University-Computer and
  Information Sciences} .
\bibitem[{Demeester et~al.(2016)Demeester, Van~Daele, Wauters and
  Hrasnica}]{demeester2016fed4fire}
\bibinfo{author}{Demeester, P.}, \bibinfo{author}{Van~Daele, P.},
  \bibinfo{author}{Wauters, T.}, \bibinfo{author}{Hrasnica, H.},
  \bibinfo{year}{2016}.
\newblock \bibinfo{title}{Fed4fire: the largest federation of testbeds in
  europe}, in: \bibinfo{booktitle}{Building the future internet through FIRE},
  pp. \bibinfo{pages}{87--109}.
\bibitem[{Deng et~al.(2020)Deng, Zhao, Fang, Yin, Dustdar and
  Zomaya}]{deng2020edge}
\bibinfo{author}{Deng, S.}, \bibinfo{author}{Zhao, H.}, \bibinfo{author}{Fang,
  W.}, \bibinfo{author}{Yin, J.}, \bibinfo{author}{Dustdar, S.},
  \bibinfo{author}{Zomaya, A.Y.}, \bibinfo{year}{2020}.
\newblock \bibinfo{title}{Edge intelligence: The confluence of edge computing
  and artificial intelligence}.
\newblock \bibinfo{journal}{IEEE Internet of Things Journal}
  \bibinfo{volume}{7}, \bibinfo{pages}{7457--7469}.
\bibitem[{Dey et~al.(2019)Dey, Mondal and Mukherjee}]{dey2019offloaded}
\bibinfo{author}{Dey, S.}, \bibinfo{author}{Mondal, J.},
  \bibinfo{author}{Mukherjee, A.}, \bibinfo{year}{2019}.
\newblock \bibinfo{title}{Offloaded execution of deep learning inference at
  edge: Challenges and insights}, in: \bibinfo{booktitle}{2019 IEEE
  International Conference on Pervasive Computing and Communications Workshops
  (PerCom Workshops)}, \bibinfo{organization}{IEEE}. pp.
  \bibinfo{pages}{855--861}.
\bibitem[{Diez-Olivan et~al.(2019)Diez-Olivan, Del~Ser, Galar and
  Sierra}]{diez2019data}
\bibinfo{author}{Diez-Olivan, A.}, \bibinfo{author}{Del~Ser, J.},
  \bibinfo{author}{Galar, D.}, \bibinfo{author}{Sierra, B.},
  \bibinfo{year}{2019}.
\newblock \bibinfo{title}{Data fusion and machine learning for industrial
  prognosis: Trends and perspectives towards industry 4.0}.
\newblock \bibinfo{journal}{Information Fusion} \bibinfo{volume}{50},
  \bibinfo{pages}{92--111}.
\bibitem[{Du and Swamy(2016)}]{du2016particle}
\bibinfo{author}{Du, K.L.}, \bibinfo{author}{Swamy, M.}, \bibinfo{year}{2016}.
\newblock \bibinfo{title}{Particle swarm optimization}, in:
  \bibinfo{booktitle}{Search and optimization by metaheuristics}.
  \bibinfo{publisher}{Springer}, pp. \bibinfo{pages}{153--173}.
\bibitem[{Duan et~al.(2019)Duan, Edwards and Dwivedi}]{duan2019artificial}
\bibinfo{author}{Duan, Y.}, \bibinfo{author}{Edwards, J.S.},
  \bibinfo{author}{Dwivedi, Y.K.}, \bibinfo{year}{2019}.
\newblock \bibinfo{title}{Artificial intelligence for decision making in the
  era of big data--evolution, challenges and research agenda}.
\newblock \bibinfo{journal}{International Journal of Information Management}
  \bibinfo{volume}{48}, \bibinfo{pages}{63--71}.
\bibitem[{Elsken et~al.(2019)Elsken, Metzen and Hutter}]{elsken2019neural}
\bibinfo{author}{Elsken, T.}, \bibinfo{author}{Metzen, J.H.},
  \bibinfo{author}{Hutter, F.}, \bibinfo{year}{2019}.
\newblock \bibinfo{title}{Neural architecture search: A survey}.
\newblock \bibinfo{journal}{The Journal of Machine Learning Research}
  \bibinfo{volume}{20}, \bibinfo{pages}{1997--2017}.
\bibitem[{Endo et~al.(2016)Endo, Rodrigues, Gon{\c{c}}alves, Kelner, Sadok and
  Curescu}]{endo2016high}
\bibinfo{author}{Endo, P.T.}, \bibinfo{author}{Rodrigues, M.},
  \bibinfo{author}{Gon{\c{c}}alves, G.E.}, \bibinfo{author}{Kelner, J.},
  \bibinfo{author}{Sadok, D.H.}, \bibinfo{author}{Curescu, C.},
  \bibinfo{year}{2016}.
\newblock \bibinfo{title}{High availability in clouds: systematic review and
  research challenges}.
\newblock \bibinfo{journal}{Journal of Cloud Computing} \bibinfo{volume}{5},
  \bibinfo{pages}{1--15}.
\bibitem[{ETP4HPC(April 29, 2020)}]{etp4-hpc-20}
\bibinfo{author}{ETP4HPC}, \bibinfo{year}{April 29, 2020}.
\newblock \bibinfo{title}{Etp4hpc strategic research agenda}.
\newblock \bibinfo{howpublished}{\url{https://www.etp4hpc.eu/sra.html}}.
\bibitem[{Fafoutis et~al.(2018)Fafoutis, Marchegiani, Elsts, Pope, Piechocki
  and Craddock}]{fafoutis2018extending}
\bibinfo{author}{Fafoutis, X.}, \bibinfo{author}{Marchegiani, L.},
  \bibinfo{author}{Elsts, A.}, \bibinfo{author}{Pope, J.},
  \bibinfo{author}{Piechocki, R.}, \bibinfo{author}{Craddock, I.},
  \bibinfo{year}{2018}.
\newblock \bibinfo{title}{Extending the battery lifetime of wearable sensors
  with embedded machine learning}, in: \bibinfo{booktitle}{2018 IEEE 4th World
  Forum on Internet of Things (WF-IoT)}, \bibinfo{organization}{IEEE}. pp.
  \bibinfo{pages}{269--274}.
\bibitem[{FAR-EDGE(July 5, 2021)}]{faredge}
\bibinfo{author}{FAR-EDGE}, \bibinfo{year}{July 5, 2021}.
\newblock \bibinfo{title}{Far-edge vision}.
\newblock \bibinfo{howpublished}{\url{http://www.faredge.eu/}}.
\bibitem[{Fei et~al.(2019)Fei, Shah, Verba, Chao, Sanchez-Anguix, Lewandowski,
  James and Usman}]{fei2019cps}
\bibinfo{author}{Fei, X.}, \bibinfo{author}{Shah, N.}, \bibinfo{author}{Verba,
  N.}, \bibinfo{author}{Chao, K.M.}, \bibinfo{author}{Sanchez-Anguix, V.},
  \bibinfo{author}{Lewandowski, J.}, \bibinfo{author}{James, A.},
  \bibinfo{author}{Usman, Z.}, \bibinfo{year}{2019}.
\newblock \bibinfo{title}{Cps data streams analytics based on machine learning
  for cloud and fog computing: A survey}.
\newblock \bibinfo{journal}{Future Generation Computer Systems}
  \bibinfo{volume}{90}, \bibinfo{pages}{435--450}.
\bibitem[{Fern{\'a}ndez-Cerero et~al.(2018)Fern{\'a}ndez-Cerero,
  Fern{\'a}ndez-Montes, Jak{\'o}bik, Ko{\l}odziej and
  Toro}]{fernandez2018score}
\bibinfo{author}{Fern{\'a}ndez-Cerero, D.},
  \bibinfo{author}{Fern{\'a}ndez-Montes, A.}, \bibinfo{author}{Jak{\'o}bik,
  A.}, \bibinfo{author}{Ko{\l}odziej, J.}, \bibinfo{author}{Toro, M.},
  \bibinfo{year}{2018}.
\newblock \bibinfo{title}{Score: Simulator for cloud optimization of resources
  and energy consumption}.
\newblock \bibinfo{journal}{Simulation Modelling Practice and Theory}
  \bibinfo{volume}{82}, \bibinfo{pages}{160--173}.
\bibitem[{Ferro and Kelly(2018)}]{ferro2018sigir}
\bibinfo{author}{Ferro, N.}, \bibinfo{author}{Kelly, D.}, \bibinfo{year}{2018}.
\newblock \bibinfo{title}{{SIGIR Initiative to Implement ACM Artifact Review
  and Badging}}, in: \bibinfo{booktitle}{ACM SIGIR Forum},
  \bibinfo{organization}{ACM New York, NY, USA}. pp. \bibinfo{pages}{4--10}.
\bibitem[{Fiuczynski(2006)}]{fiuczynski2006planetlab}
\bibinfo{author}{Fiuczynski, M.E.}, \bibinfo{year}{2006}.
\newblock \bibinfo{title}{Planetlab: overview, history, and future directions}.
\newblock \bibinfo{journal}{ACM SIGOPS Operating Systems Review}
  \bibinfo{volume}{40}, \bibinfo{pages}{6--10}.
\bibitem[{Fu et~al.(2019)Fu, Ghaffar, Davis and Lee}]{fu2019edgewise}
\bibinfo{author}{Fu, X.}, \bibinfo{author}{Ghaffar, T.},
  \bibinfo{author}{Davis, J.C.}, \bibinfo{author}{Lee, D.},
  \bibinfo{year}{2019}.
\newblock \bibinfo{title}{Edgewise: a better stream processing engine for the
  edge}, in: \bibinfo{booktitle}{2019 $\{$USENIX$\}$ Annual Technical
  Conference ($\{$USENIX$\}$$\{$ATC$\}$ 19)}, pp. \bibinfo{pages}{929--946}.
\bibitem[{Ghosh and Grolinger(2020)}]{ghosh2020edge}
\bibinfo{author}{Ghosh, A.M.}, \bibinfo{author}{Grolinger, K.},
  \bibinfo{year}{2020}.
\newblock \bibinfo{title}{Edge-cloud computing for internet of things data
  analytics: Embedding intelligence in the edge with deep learning}.
\newblock \bibinfo{journal}{IEEE Transactions on Industrial Informatics}
  \bibinfo{volume}{17}, \bibinfo{pages}{2191--2200}.
\bibitem[{Gill et~al.(2019)Gill, Tuli, Xu, Singh, Singh, Lindsay, Tuli,
  Smirnova, Singh, Jain et~al.}]{gill2019transformative}
\bibinfo{author}{Gill, S.S.}, \bibinfo{author}{Tuli, S.}, \bibinfo{author}{Xu,
  M.}, \bibinfo{author}{Singh, I.}, \bibinfo{author}{Singh, K.V.},
  \bibinfo{author}{Lindsay, D.}, \bibinfo{author}{Tuli, S.},
  \bibinfo{author}{Smirnova, D.}, \bibinfo{author}{Singh, M.},
  \bibinfo{author}{Jain, U.}, et~al., \bibinfo{year}{2019}.
\newblock \bibinfo{title}{Transformative effects of iot, blockchain and
  artificial intelligence on cloud computing: Evolution, vision, trends and
  open challenges}.
\newblock \bibinfo{journal}{Internet of Things} \bibinfo{volume}{8},
  \bibinfo{pages}{100118}.
\bibitem[{Goodfellow et~al.(2016)Goodfellow, Bengio and
  Courville}]{goodfellow2016deep}
\bibinfo{author}{Goodfellow, I.}, \bibinfo{author}{Bengio, Y.},
  \bibinfo{author}{Courville, A.}, \bibinfo{year}{2016}.
\newblock \bibinfo{title}{Deep learning}.
\newblock \bibinfo{publisher}{MIT press}.
\bibitem[{Grzenda et~al.(2020)Grzenda, Kwasiborska and
  Zaremba}]{grzenda2020hybrid}
\bibinfo{author}{Grzenda, M.}, \bibinfo{author}{Kwasiborska, K.},
  \bibinfo{author}{Zaremba, T.}, \bibinfo{year}{2020}.
\newblock \bibinfo{title}{Hybrid short term prediction to address limited
  timeliness of public transport data streams}.
\newblock \bibinfo{journal}{Neurocomputing} \bibinfo{volume}{391},
  \bibinfo{pages}{305--317}.
\bibitem[{Grzywaczewski(2017)}]{grzywaczewski2017training}
\bibinfo{author}{Grzywaczewski, A.}, \bibinfo{year}{2017}.
\newblock \bibinfo{title}{Training ai for self-driving vehicles: the challenge
  of scale}.
\newblock \bibinfo{journal}{Available from Internet: https://devblogs. nvidia.
  com/training-self-driving-vehicles-challenge-scale} .
\bibitem[{Guo et~al.(2021)Guo, Hu and Hu}]{guo2021mistify}
\bibinfo{author}{Guo, P.}, \bibinfo{author}{Hu, B.}, \bibinfo{author}{Hu, W.},
  \bibinfo{year}{2021}.
\newblock \bibinfo{title}{Mistify: Automating dnn model porting for on-device
  inference at the edge.}, in: \bibinfo{booktitle}{NSDI}, pp.
  \bibinfo{pages}{705--719}.
\bibitem[{Gupta et~al.(2017)Gupta, Vahid~Dastjerdi, Ghosh and
  Buyya}]{gupta2017ifogsim}
\bibinfo{author}{Gupta, H.}, \bibinfo{author}{Vahid~Dastjerdi, A.},
  \bibinfo{author}{Ghosh, S.K.}, \bibinfo{author}{Buyya, R.},
  \bibinfo{year}{2017}.
\newblock \bibinfo{title}{ifogsim: A toolkit for modeling and simulation of
  resource management techniques in the internet of things, edge and fog
  computing environments}.
\newblock \bibinfo{journal}{Software: Practice and Experience}
  \bibinfo{volume}{47}, \bibinfo{pages}{1275--1296}.
\bibitem[{Hamdan et~al.(2020)Hamdan, Ayyash and Almajali}]{hamdan2020edge}
\bibinfo{author}{Hamdan, S.}, \bibinfo{author}{Ayyash, M.},
  \bibinfo{author}{Almajali, S.}, \bibinfo{year}{2020}.
\newblock \bibinfo{title}{Edge-computing architectures for internet of things
  applications: A survey}.
\newblock \bibinfo{journal}{Sensors} \bibinfo{volume}{20},
  \bibinfo{pages}{6441}.
\bibitem[{Hasenburg et~al.(2018)Hasenburg, Werner and
  Bermbach}]{hasenburg_supporting_2018}
\bibinfo{author}{Hasenburg, J.}, \bibinfo{author}{Werner, S.},
  \bibinfo{author}{Bermbach, D.}, \bibinfo{year}{2018}.
\newblock \bibinfo{title}{Supporting the evaluation of fog-based {IoT}
  applications during the design phase}, in: \bibinfo{booktitle}{Proceedings of
  the 5th Workshop on Middleware and Applications for the Internet of Things
  (M4IoT 2018)}, \bibinfo{publisher}{{ACM}}.
\bibitem[{Hauswirth and Le-Phuoc(2020)}]{hauswirth2020autonomous}
\bibinfo{author}{Hauswirth, M.}, \bibinfo{author}{Le-Phuoc, D.},
  \bibinfo{year}{2020}.
\newblock \bibinfo{title}{Autonomous rdf stream processing for iot edge
  devices}, in: \bibinfo{booktitle}{Semantic Technology: 9th Joint
  International Conference, JIST 2019, Hangzhou, China, November 25--27, 2019,
  Proceedings}, \bibinfo{organization}{Springer Nature}. p.
  \bibinfo{pages}{304}.
\bibitem[{Hong and Chandra(2019)}]{hong2019dlion}
\bibinfo{author}{Hong, R.}, \bibinfo{author}{Chandra, A.},
  \bibinfo{year}{2019}.
\newblock \bibinfo{title}{Dlion: Decentralized distributed deep learning in
  micro-clouds}, in: \bibinfo{booktitle}{11th $\{$USENIX$\}$ Workshop on Hot
  Topics in Cloud Computing (HotCloud 19)}.
\bibitem[{Huang et~al.(2018)Huang, Lin, Tsai, Yan and Shih}]{huang2018building}
\bibinfo{author}{Huang, Z.}, \bibinfo{author}{Lin, K.J.},
  \bibinfo{author}{Tsai, B.L.}, \bibinfo{author}{Yan, S.},
  \bibinfo{author}{Shih, C.S.}, \bibinfo{year}{2018}.
\newblock \bibinfo{title}{Building edge intelligence for online activity
  recognition in service-oriented iot systems}.
\newblock \bibinfo{journal}{Future Generation Computer Systems}
  \bibinfo{volume}{87}, \bibinfo{pages}{557--567}.
\bibitem[{Kaur et~al.(2014)Kaur, Singh and Ghumman}]{kaur2014mininet}
\bibinfo{author}{Kaur, K.}, \bibinfo{author}{Singh, J.},
  \bibinfo{author}{Ghumman, N.S.}, \bibinfo{year}{2014}.
\newblock \bibinfo{title}{Mininet as software defined networking testing
  platform}, in: \bibinfo{booktitle}{International Conference on Communication,
  Computing \& Systems (ICCCS)}, pp. \bibinfo{pages}{139--42}.
\bibitem[{Keahey et~al.(2020)Keahey, Anderson, Zhen, Riteau, Ruth, Stanzione,
  Cevik, Colleran, Gunawi, Hammock, Mambretti, Barnes, Halbach, Rocha and
  Stubbs}]{keahey2020lessons}
\bibinfo{author}{Keahey, K.}, \bibinfo{author}{Anderson, J.},
  \bibinfo{author}{Zhen, Z.}, \bibinfo{author}{Riteau, P.},
  \bibinfo{author}{Ruth, P.}, \bibinfo{author}{Stanzione, D.},
  \bibinfo{author}{Cevik, M.}, \bibinfo{author}{Colleran, J.},
  \bibinfo{author}{Gunawi, H.S.}, \bibinfo{author}{Hammock, C.},
  \bibinfo{author}{Mambretti, J.}, \bibinfo{author}{Barnes, A.},
  \bibinfo{author}{Halbach, F.}, \bibinfo{author}{Rocha, A.},
  \bibinfo{author}{Stubbs, J.}, \bibinfo{year}{2020}.
\newblock \bibinfo{title}{Lessons learned from the chameleon testbed}, in:
  \bibinfo{booktitle}{Proceedings of the 2020 USENIX Annual Technical
  Conference (USENIX ATC '20)}. \bibinfo{publisher}{USENIX Association}.
\bibitem[{Keele et~al.(2007)}]{keele2007guidelines}
\bibinfo{author}{Keele, S.}, et~al., \bibinfo{year}{2007}.
\newblock \bibinfo{title}{Guidelines for performing systematic literature
  reviews in software engineering}.
\newblock \bibinfo{type}{Technical Report}. Citeseer.
\bibitem[{Khayyam et~al.(2020)Khayyam, Javadi, Jalili and
  Jazar}]{khayyam2020artificial}
\bibinfo{author}{Khayyam, H.}, \bibinfo{author}{Javadi, B.},
  \bibinfo{author}{Jalili, M.}, \bibinfo{author}{Jazar, R.N.},
  \bibinfo{year}{2020}.
\newblock \bibinfo{title}{Artificial intelligence and internet of things for
  autonomous vehicles}, in: \bibinfo{booktitle}{Nonlinear approaches in
  engineering applications}. \bibinfo{publisher}{Springer}, pp.
  \bibinfo{pages}{39--68}.
\bibitem[{Ko{\l}odziej and Gonz{\'a}lez-V{\'e}lez(2019)}]{kolodziej2019high}
\bibinfo{author}{Ko{\l}odziej, J.}, \bibinfo{author}{Gonz{\'a}lez-V{\'e}lez,
  H.}, \bibinfo{year}{2019}.
\newblock \bibinfo{title}{High-Performance Modelling and Simulation for Big
  Data Applications: Selected Results of the COST Action IC1406 cHiPSet}.
\newblock \bibinfo{publisher}{Springer Nature}.
\bibitem[{Kourtellis et~al.(2021)Kourtellis, Herodotou, Grzenda, Wawrzyniak and
  Bifet}]{kourtellis2021s2ce}
\bibinfo{author}{Kourtellis, N.}, \bibinfo{author}{Herodotou, H.},
  \bibinfo{author}{Grzenda, M.}, \bibinfo{author}{Wawrzyniak, P.},
  \bibinfo{author}{Bifet, A.}, \bibinfo{year}{2021}.
\newblock \bibinfo{title}{S2ce: a hybrid cloud and edge orchestrator for mining
  exascale distributed streams}, in: \bibinfo{booktitle}{Proceedings of the
  15th ACM International Conference on Distributed and Event-based Systems},
  pp. \bibinfo{pages}{103--113}.
\bibitem[{Kukreja et~al.(2019)Kukreja, Shilova, Beaumont, Huckelheim, Ferrier,
  Hovland and Gorman}]{kukreja2019training}
\bibinfo{author}{Kukreja, N.}, \bibinfo{author}{Shilova, A.},
  \bibinfo{author}{Beaumont, O.}, \bibinfo{author}{Huckelheim, J.},
  \bibinfo{author}{Ferrier, N.}, \bibinfo{author}{Hovland, P.},
  \bibinfo{author}{Gorman, G.}, \bibinfo{year}{2019}.
\newblock \bibinfo{title}{Training on the edge: The why and the how}, in:
  \bibinfo{booktitle}{2019 IEEE International Parallel and Distributed
  Processing Symposium Workshops (IPDPSW)}, \bibinfo{organization}{IEEE}. pp.
  \bibinfo{pages}{899--903}.
\bibitem[{Kumar et~al.(2017)Kumar, Goyal and Varma}]{kumar2017resource}
\bibinfo{author}{Kumar, A.}, \bibinfo{author}{Goyal, S.},
  \bibinfo{author}{Varma, M.}, \bibinfo{year}{2017}.
\newblock \bibinfo{title}{Resource-efficient machine learning in 2 kb ram for
  the internet of things}, in: \bibinfo{booktitle}{International Conference on
  Machine Learning}, \bibinfo{organization}{PMLR}. pp.
  \bibinfo{pages}{1935--1944}.
\bibitem[{Kumar et~al.(2019)Kumar, Ramkumar, Sindhu and
  Chandra}]{kumar2019decaf}
\bibinfo{author}{Kumar, D.}, \bibinfo{author}{Ramkumar, A.A.},
  \bibinfo{author}{Sindhu, R.}, \bibinfo{author}{Chandra, A.},
  \bibinfo{year}{2019}.
\newblock \bibinfo{title}{Decaf: Iterative collaborative processing over the
  edge}, in: \bibinfo{booktitle}{2nd $\{$USENIX$\}$ Workshop on Hot Topics in
  Edge Computing (HotEdge 19)}.
\bibitem[{Lee et~al.(2020)Lee, Singh, Azamfar and Pandhare}]{lee2020industrial}
\bibinfo{author}{Lee, J.}, \bibinfo{author}{Singh, J.},
  \bibinfo{author}{Azamfar, M.}, \bibinfo{author}{Pandhare, V.},
  \bibinfo{year}{2020}.
\newblock \bibinfo{title}{Industrial ai and predictive analytics for smart
  manufacturing systems}, in: \bibinfo{booktitle}{Smart Manufacturing}.
  \bibinfo{publisher}{Elsevier}, pp. \bibinfo{pages}{213--244}.
\bibitem[{Lera et~al.(2019)Lera, Guerrero and Juiz}]{lera2019yafs}
\bibinfo{author}{Lera, I.}, \bibinfo{author}{Guerrero, C.},
  \bibinfo{author}{Juiz, C.}, \bibinfo{year}{2019}.
\newblock \bibinfo{title}{Yafs: A simulator for iot scenarios in fog
  computing}.
\newblock \bibinfo{journal}{IEEE Access} \bibinfo{volume}{7},
  \bibinfo{pages}{91745--91758}.
\bibitem[{Li et~al.(2019)Li, Zeng, Zhou and Chen}]{li2019edge}
\bibinfo{author}{Li, E.}, \bibinfo{author}{Zeng, L.}, \bibinfo{author}{Zhou,
  Z.}, \bibinfo{author}{Chen, X.}, \bibinfo{year}{2019}.
\newblock \bibinfo{title}{Edge ai: On-demand accelerating deep neural network
  inference via edge computing}.
\newblock \bibinfo{journal}{IEEE Transactions on Wireless Communications}
  \bibinfo{volume}{19}, \bibinfo{pages}{447--457}.
\bibitem[{Liang et~al.(2006)Liang, Huang, Saratchandran and
  Sundararajan}]{liang2006fast}
\bibinfo{author}{Liang, N.Y.}, \bibinfo{author}{Huang, G.B.},
  \bibinfo{author}{Saratchandran, P.}, \bibinfo{author}{Sundararajan, N.},
  \bibinfo{year}{2006}.
\newblock \bibinfo{title}{A fast and accurate online sequential learning
  algorithm for feedforward networks}.
\newblock \bibinfo{journal}{IEEE Transactions on neural networks}
  \bibinfo{volume}{17}, \bibinfo{pages}{1411--1423}.
\bibitem[{Liaw et~al.(2018)Liaw, Liang, Nishihara, Moritz, Gonzalez and
  Stoica}]{liaw2018tune}
\bibinfo{author}{Liaw, R.}, \bibinfo{author}{Liang, E.},
  \bibinfo{author}{Nishihara, R.}, \bibinfo{author}{Moritz, P.},
  \bibinfo{author}{Gonzalez, J.E.}, \bibinfo{author}{Stoica, I.},
  \bibinfo{year}{2018}.
\newblock \bibinfo{title}{Tune: A research platform for distributed model
  selection and training}.
\newblock \bibinfo{journal}{arXiv preprint arXiv:1807.05118} .
\bibitem[{Liu et~al.(2019)Liu, Tang, Li, Cai, Zhang and Zhou}]{liu2019survey}
\bibinfo{author}{Liu, F.}, \bibinfo{author}{Tang, G.}, \bibinfo{author}{Li,
  Y.}, \bibinfo{author}{Cai, Z.}, \bibinfo{author}{Zhang, X.},
  \bibinfo{author}{Zhou, T.}, \bibinfo{year}{2019}.
\newblock \bibinfo{title}{A survey on edge computing systems and tools}.
\newblock \bibinfo{journal}{Proceedings of the IEEE} \bibinfo{volume}{107},
  \bibinfo{pages}{1537--1562}.
\bibitem[{Lobo et~al.(2020)Lobo, Del~Ser, Bifet and Kasabov}]{lobo2020spiking}
\bibinfo{author}{Lobo, J.L.}, \bibinfo{author}{Del~Ser, J.},
  \bibinfo{author}{Bifet, A.}, \bibinfo{author}{Kasabov, N.},
  \bibinfo{year}{2020}.
\newblock \bibinfo{title}{Spiking neural networks and online learning: An
  overview and perspectives}.
\newblock \bibinfo{journal}{Neural Networks} \bibinfo{volume}{121},
  \bibinfo{pages}{88--100}.
\bibitem[{Loghin et~al.(2020)Loghin, Cai, Chen, Dinh, Fan, Lin, Ng, Ooi, Sun,
  Ta et~al.}]{loghin2020disruptions}
\bibinfo{author}{Loghin, D.}, \bibinfo{author}{Cai, S.}, \bibinfo{author}{Chen,
  G.}, \bibinfo{author}{Dinh, T.T.A.}, \bibinfo{author}{Fan, F.},
  \bibinfo{author}{Lin, Q.}, \bibinfo{author}{Ng, J.}, \bibinfo{author}{Ooi,
  B.C.}, \bibinfo{author}{Sun, X.}, \bibinfo{author}{Ta, Q.T.}, et~al.,
  \bibinfo{year}{2020}.
\newblock \bibinfo{title}{The disruptions of 5g on data-driven technologies and
  applications}.
\newblock \bibinfo{journal}{IEEE transactions on knowledge and data
  engineering} \bibinfo{volume}{32}, \bibinfo{pages}{1179--1198}.
\bibitem[{Lu et~al.(2019)Lu, Yao and Shi}]{lu2019collaborative}
\bibinfo{author}{Lu, S.}, \bibinfo{author}{Yao, Y.}, \bibinfo{author}{Shi, W.},
  \bibinfo{year}{2019}.
\newblock \bibinfo{title}{Collaborative learning on the edges: A case study on
  connected vehicles}, in: \bibinfo{booktitle}{2nd $\{$USENIX$\}$ Workshop on
  Hot Topics in Edge Computing (HotEdge 19)}.
\bibitem[{Luckow et~al.(2021)Luckow, Rattan and Jha}]{luckow2021exploring}
\bibinfo{author}{Luckow, A.}, \bibinfo{author}{Rattan, K.},
  \bibinfo{author}{Jha, S.}, \bibinfo{year}{2021}.
\newblock \bibinfo{title}{Exploring task placement for edge-to-cloud
  applications using emulation}, in: \bibinfo{booktitle}{2021 IEEE 5th
  International Conference on Fog and Edge Computing (ICFEC)},
  \bibinfo{organization}{IEEE}. pp. \bibinfo{pages}{79--83}.
\bibitem[{L’heureux et~al.(2017)L’heureux, Grolinger, Elyamany and
  Capretz}]{l2017machine}
\bibinfo{author}{L’heureux, A.}, \bibinfo{author}{Grolinger, K.},
  \bibinfo{author}{Elyamany, H.F.}, \bibinfo{author}{Capretz, M.A.},
  \bibinfo{year}{2017}.
\newblock \bibinfo{title}{Machine learning with big data: Challenges and
  approaches}.
\newblock \bibinfo{journal}{Ieee Access} \bibinfo{volume}{5},
  \bibinfo{pages}{7776--7797}.
\bibitem[{Mahmood(2018)}]{mahmood2018fog}
\bibinfo{author}{Mahmood, Z.}, \bibinfo{year}{2018}.
\newblock \bibinfo{title}{{Fog Computing: Concepts, Frameworks and
  Technologies}}.
\newblock \bibinfo{publisher}{Springer}.
\bibitem[{Malik et~al.(2020)Malik, Qayyum, Rahman, Khan, Khalid and
  Khan}]{malik2020xfogsim}
\bibinfo{author}{Malik, A.W.}, \bibinfo{author}{Qayyum, T.},
  \bibinfo{author}{Rahman, A.U.}, \bibinfo{author}{Khan, M.A.},
  \bibinfo{author}{Khalid, O.}, \bibinfo{author}{Khan, S.U.},
  \bibinfo{year}{2020}.
\newblock \bibinfo{title}{Xfogsim: A distributed fog resource management
  framework for sustainable iot services}.
\newblock \bibinfo{journal}{IEEE Transactions on Sustainable Computing}
  \bibinfo{volume}{6}, \bibinfo{pages}{691--702}.
\bibitem[{Mayer et~al.(2017)Mayer, Graser, Gupta, Saurez and
  Ramachandran}]{mayer2017emufog}
\bibinfo{author}{Mayer, R.}, \bibinfo{author}{Graser, L.},
  \bibinfo{author}{Gupta, H.}, \bibinfo{author}{Saurez, E.},
  \bibinfo{author}{Ramachandran, U.}, \bibinfo{year}{2017}.
\newblock \bibinfo{title}{Emufog: Extensible and scalable emulation of
  large-scale fog computing infrastructures}, in: \bibinfo{booktitle}{2017 IEEE
  Fog World Congress (FWC)}, \bibinfo{organization}{IEEE}. pp.
  \bibinfo{pages}{1--6}.
\bibitem[{Medina et~al.(2001)Medina, Lakhina, Matta and
  Byers}]{medina2001brite}
\bibinfo{author}{Medina, A.}, \bibinfo{author}{Lakhina, A.},
  \bibinfo{author}{Matta, I.}, \bibinfo{author}{Byers, J.},
  \bibinfo{year}{2001}.
\newblock \bibinfo{title}{Brite: An approach to universal topology generation},
  in: \bibinfo{booktitle}{MASCOTS 2001, Proceedings Ninth International
  Symposium on Modeling, Analysis and Simulation of Computer and
  Telecommunication Systems}, \bibinfo{organization}{IEEE}. pp.
  \bibinfo{pages}{346--353}.
\bibitem[{Midya et~al.(2018)Midya, Roy, Majumder and Phadikar}]{midya2018multi}
\bibinfo{author}{Midya, S.}, \bibinfo{author}{Roy, A.},
  \bibinfo{author}{Majumder, K.}, \bibinfo{author}{Phadikar, S.},
  \bibinfo{year}{2018}.
\newblock \bibinfo{title}{Multi-objective optimization technique for resource
  allocation and task scheduling in vehicular cloud architecture: A hybrid
  adaptive nature inspired approach}.
\newblock \bibinfo{journal}{Journal of Network and Computer Applications}
  \bibinfo{volume}{103}, \bibinfo{pages}{58--84}.
\bibitem[{Mijuskovic et~al.(2021)Mijuskovic, Chiumento, Bemthuis, Aldea and
  Havinga}]{mijuskovic2021resource}
\bibinfo{author}{Mijuskovic, A.}, \bibinfo{author}{Chiumento, A.},
  \bibinfo{author}{Bemthuis, R.}, \bibinfo{author}{Aldea, A.},
  \bibinfo{author}{Havinga, P.}, \bibinfo{year}{2021}.
\newblock \bibinfo{title}{Resource management techniques for cloud/fog and edge
  computing: An evaluation framework and classification}.
\newblock \bibinfo{journal}{Sensors} \bibinfo{volume}{21},
  \bibinfo{pages}{1832}.
\bibitem[{Mirjalili(2019)}]{mirjalili2019genetic}
\bibinfo{author}{Mirjalili, S.}, \bibinfo{year}{2019}.
\newblock \bibinfo{title}{Genetic algorithm}, in:
  \bibinfo{booktitle}{Evolutionary algorithms and neural networks}.
  \bibinfo{publisher}{Springer}, pp. \bibinfo{pages}{43--55}.
\bibitem[{Mohammadi et~al.(2018)Mohammadi, Al-Fuqaha, Sorour and
  Guizani}]{mohammadi2018deep}
\bibinfo{author}{Mohammadi, M.}, \bibinfo{author}{Al-Fuqaha, A.},
  \bibinfo{author}{Sorour, S.}, \bibinfo{author}{Guizani, M.},
  \bibinfo{year}{2018}.
\newblock \bibinfo{title}{Deep learning for iot big data and streaming
  analytics: A survey}.
\newblock \bibinfo{journal}{IEEE Communications Surveys \& Tutorials}
  \bibinfo{volume}{20}, \bibinfo{pages}{2923--2960}.
\bibitem[{Mrozek et~al.(2020)Mrozek, Koczur and
  Ma{\l}ysiak-Mrozek}]{mrozek2020fall}
\bibinfo{author}{Mrozek, D.}, \bibinfo{author}{Koczur, A.},
  \bibinfo{author}{Ma{\l}ysiak-Mrozek, B.}, \bibinfo{year}{2020}.
\newblock \bibinfo{title}{Fall detection in older adults with mobile iot
  devices and machine learning in the cloud and on the edge}.
\newblock \bibinfo{journal}{Information Sciences} \bibinfo{volume}{537},
  \bibinfo{pages}{132--147}.
\bibitem[{Mwase et~al.(2022)Mwase, Jin, Westerlund, Tenhunen and
  Zou}]{mwase2022communication}
\bibinfo{author}{Mwase, C.}, \bibinfo{author}{Jin, Y.},
  \bibinfo{author}{Westerlund, T.}, \bibinfo{author}{Tenhunen, H.},
  \bibinfo{author}{Zou, Z.}, \bibinfo{year}{2022}.
\newblock \bibinfo{title}{Communication-efficient distributed ai strategies for
  the iot edge}.
\newblock \bibinfo{journal}{Future Generation Computer Systems} .
\bibitem[{Nair et~al.(2018)Nair, Shetty and Shetty}]{nair2018applying}
\bibinfo{author}{Nair, L.R.}, \bibinfo{author}{Shetty, S.D.},
  \bibinfo{author}{Shetty, S.D.}, \bibinfo{year}{2018}.
\newblock \bibinfo{title}{Applying spark based machine learning model on
  streaming big data for health status prediction}.
\newblock \bibinfo{journal}{Computers \& Electrical Engineering}
  \bibinfo{volume}{65}, \bibinfo{pages}{393--399}.
\bibitem[{Nandan~Jha et~al.(2019)Nandan~Jha, Alwasel, Alshoshan, Huang, Naha,
  Battula, Garg, Puthal, James, Zomaya et~al.}]{nandan2019iotsim}
\bibinfo{author}{Nandan~Jha, D.}, \bibinfo{author}{Alwasel, K.},
  \bibinfo{author}{Alshoshan, A.}, \bibinfo{author}{Huang, X.},
  \bibinfo{author}{Naha, R.K.}, \bibinfo{author}{Battula, S.K.},
  \bibinfo{author}{Garg, S.}, \bibinfo{author}{Puthal, D.},
  \bibinfo{author}{James, P.}, \bibinfo{author}{Zomaya, A.Y.}, et~al.,
  \bibinfo{year}{2019}.
\newblock \bibinfo{title}{Iotsim-edge: A simulation framework for modeling the
  behaviour of iot and edge computing environments}.
\newblock \bibinfo{journal}{arXiv e-prints} , \bibinfo{pages}{arXiv--1910}.
\bibitem[{Nguyen et~al.(2019)Nguyen, Dlugolinsky, Bob{\'a}k, Tran, Garc{\'\i}a,
  Heredia, Mal{\'\i}k and Hluch{\`y}}]{nguyen2019machine}
\bibinfo{author}{Nguyen, G.}, \bibinfo{author}{Dlugolinsky, S.},
  \bibinfo{author}{Bob{\'a}k, M.}, \bibinfo{author}{Tran, V.},
  \bibinfo{author}{Garc{\'\i}a, {\'A}.L.}, \bibinfo{author}{Heredia, I.},
  \bibinfo{author}{Mal{\'\i}k, P.}, \bibinfo{author}{Hluch{\`y}, L.},
  \bibinfo{year}{2019}.
\newblock \bibinfo{title}{Machine learning and deep learning frameworks and
  libraries for large-scale data mining: a survey}.
\newblock \bibinfo{journal}{Artificial Intelligence Review}
  \bibinfo{volume}{52}, \bibinfo{pages}{77--124}.
\bibitem[{Nikouei et~al.(2019)Nikouei, Chen, Song, Choi and
  Faughnan}]{nikouei2019toward}
\bibinfo{author}{Nikouei, S.Y.}, \bibinfo{author}{Chen, Y.},
  \bibinfo{author}{Song, S.}, \bibinfo{author}{Choi, B.Y.},
  \bibinfo{author}{Faughnan, T.R.}, \bibinfo{year}{2019}.
\newblock \bibinfo{title}{Toward intelligent surveillance as an edge network
  service (isense) using lightweight detection and tracking algorithms}.
\newblock \bibinfo{journal}{IEEE Transactions on Services Computing} .
\bibitem[{Nussbaum(2019)}]{nussbaum2019overview}
\bibinfo{author}{Nussbaum, L.}, \bibinfo{year}{2019}.
\newblock \bibinfo{title}{An overview of fed4fire testbeds--and beyond?}, in:
  \bibinfo{booktitle}{GEFI-Global Experimentation for Future Internet
  Workshop}.
\bibitem[{Ometov et~al.(2022)Ometov, Molua, Komarov and
  Nurmi}]{ometov2022survey}
\bibinfo{author}{Ometov, A.}, \bibinfo{author}{Molua, O.L.},
  \bibinfo{author}{Komarov, M.}, \bibinfo{author}{Nurmi, J.},
  \bibinfo{year}{2022}.
\newblock \bibinfo{title}{A survey of security in cloud, edge, and fog
  computing}.
\newblock \bibinfo{journal}{Sensors} \bibinfo{volume}{22},
  \bibinfo{pages}{927}.
\bibitem[{ORBIT(2016)}]{orbit2016open}
\bibinfo{author}{ORBIT, P.}, \bibinfo{year}{2016}.
\newblock \bibinfo{title}{Open-access research testbed for next-generation
  wireless networks}.
\bibitem[{P{\"a}{\"a}kk{\"o}nen and Pakkala(2020)}]{paakkonen2020extending}
\bibinfo{author}{P{\"a}{\"a}kk{\"o}nen, P.}, \bibinfo{author}{Pakkala, D.},
  \bibinfo{year}{2020}.
\newblock \bibinfo{title}{Extending reference architecture of big data systems
  towards machine learning in edge computing environments}.
\newblock \bibinfo{journal}{Journal of Big Data} \bibinfo{volume}{7},
  \bibinfo{pages}{1--29}.
\bibitem[{Pan and Yang(2009)}]{pan2009survey}
\bibinfo{author}{Pan, S.J.}, \bibinfo{author}{Yang, Q.}, \bibinfo{year}{2009}.
\newblock \bibinfo{title}{A survey on transfer learning}.
\newblock \bibinfo{journal}{IEEE Transactions on knowledge and data
  engineering} \bibinfo{volume}{22}, \bibinfo{pages}{1345--1359}.
\bibitem[{Parashar(2022)}]{parashar2022eic}
\bibinfo{author}{Parashar, M.}, \bibinfo{year}{2022}.
\newblock \bibinfo{title}{Eic editorial--advancing reproducibility in parallel
  and distributed systems research}.
\newblock \bibinfo{journal}{IEEE Transactions on Parallel \& Distributed
  Systems} \bibinfo{volume}{33}, \bibinfo{pages}{2010--2010}.
\bibitem[{Pathak et~al.(2018)Pathak, Pandey and
  Rautaray}]{pathak2018construing}
\bibinfo{author}{Pathak, A.R.}, \bibinfo{author}{Pandey, M.},
  \bibinfo{author}{Rautaray, S.}, \bibinfo{year}{2018}.
\newblock \bibinfo{title}{Construing the big data based on taxonomy, analytics
  and approaches}.
\newblock \bibinfo{journal}{Iran Journal of Computer Science}
  \bibinfo{volume}{1}, \bibinfo{pages}{237--259}.
\bibitem[{Pechlivanidou et~al.(2014)Pechlivanidou, Katsalis, Igoumenos,
  Katsaros, Korakis and Tassiulas}]{pechlivanidou2014nitos}
\bibinfo{author}{Pechlivanidou, K.}, \bibinfo{author}{Katsalis, K.},
  \bibinfo{author}{Igoumenos, I.}, \bibinfo{author}{Katsaros, D.},
  \bibinfo{author}{Korakis, T.}, \bibinfo{author}{Tassiulas, L.},
  \bibinfo{year}{2014}.
\newblock \bibinfo{title}{Nitos testbed: A cloud based wireless experimentation
  facility}, in: \bibinfo{booktitle}{2014 26th International Teletraffic
  Congress (ITC)}, \bibinfo{organization}{IEEE}. pp. \bibinfo{pages}{1--6}.
\bibitem[{P{\'e}rez et~al.(2018)P{\'e}rez, Gutierrez-Torre, Berral and
  Carrera}]{perez2018resilient}
\bibinfo{author}{P{\'e}rez, J.L.}, \bibinfo{author}{Gutierrez-Torre, A.},
  \bibinfo{author}{Berral, J.L.}, \bibinfo{author}{Carrera, D.},
  \bibinfo{year}{2018}.
\newblock \bibinfo{title}{A resilient and distributed near real-time traffic
  forecasting application for fog computing environments}.
\newblock \bibinfo{journal}{Future Generation Computer Systems}
  \bibinfo{volume}{87}, \bibinfo{pages}{198--212}.
\bibitem[{Peuster et~al.(2018)Peuster, Kampmeyer and
  Karl}]{peuster2018containernet}
\bibinfo{author}{Peuster, M.}, \bibinfo{author}{Kampmeyer, J.},
  \bibinfo{author}{Karl, H.}, \bibinfo{year}{2018}.
\newblock \bibinfo{title}{Containernet 2.0: A rapid prototyping platform for
  hybrid service function chains}, in: \bibinfo{booktitle}{2018 4th IEEE
  Conference on Network Softwarization and Workshops (NetSoft)},
  \bibinfo{organization}{IEEE}. pp. \bibinfo{pages}{335--337}.
\bibitem[{Pham and Karaboga(2012)}]{pham2012intelligent}
\bibinfo{author}{Pham, D.}, \bibinfo{author}{Karaboga, D.},
  \bibinfo{year}{2012}.
\newblock \bibinfo{title}{Intelligent optimisation techniques: genetic
  algorithms, tabu search, simulated annealing and neural networks}.
\newblock \bibinfo{publisher}{Springer Science \& Business Media}.
\bibitem[{Prabhu(2019)}]{prabhu2019fog}
\bibinfo{author}{Prabhu, C.}, \bibinfo{year}{2019}.
\newblock \bibinfo{title}{Fog Computing, Deep Learning and Big Data
  Analytics-Research Directions}.
\newblock \bibinfo{publisher}{Springer}.
\bibitem[{Qayyum et~al.(2018)Qayyum, Malik, Khattak, Khalid and
  Khan}]{qayyum2018fognetsim++}
\bibinfo{author}{Qayyum, T.}, \bibinfo{author}{Malik, A.W.},
  \bibinfo{author}{Khattak, M.A.K.}, \bibinfo{author}{Khalid, O.},
  \bibinfo{author}{Khan, S.U.}, \bibinfo{year}{2018}.
\newblock \bibinfo{title}{Fognetsim++: A toolkit for modeling and simulation of
  distributed fog environment}.
\newblock \bibinfo{journal}{IEEE Access} \bibinfo{volume}{6},
  \bibinfo{pages}{63570--63583}.
\bibitem[{RADICAL-DREAMER(Feb 12, 2022)}]{radicaldreamer}
\bibinfo{author}{RADICAL-DREAMER}, \bibinfo{year}{Feb 12, 2022}.
\newblock \bibinfo{title}{Radical-dreamer: Dynamic runtime and execution
  adaptive middleware emulator (rd)}.
\newblock
  \bibinfo{howpublished}{\url{https://github.com/radical-project/radical.dreamer/}}.
\bibitem[{Ranjan et~al.(2018)Ranjan, Rana, Nepal, Yousif, James, Wen, Barr,
  Watson, Jayaraman, Georgakopoulos et~al.}]{ranjan2018next}
\bibinfo{author}{Ranjan, R.}, \bibinfo{author}{Rana, O.},
  \bibinfo{author}{Nepal, S.}, \bibinfo{author}{Yousif, M.},
  \bibinfo{author}{James, P.}, \bibinfo{author}{Wen, Z.},
  \bibinfo{author}{Barr, S.}, \bibinfo{author}{Watson, P.},
  \bibinfo{author}{Jayaraman, P.P.}, \bibinfo{author}{Georgakopoulos, D.},
  et~al., \bibinfo{year}{2018}.
\newblock \bibinfo{title}{The next grand challenges: Integrating the internet
  of things and data science}.
\newblock \bibinfo{journal}{IEEE Cloud Computing} \bibinfo{volume}{5},
  \bibinfo{pages}{12--26}.
\bibitem[{Rao et~al.(2019)Rao, Mitra, Bhatt and Goswami}]{rao2019big}
\bibinfo{author}{Rao, T.R.}, \bibinfo{author}{Mitra, P.},
  \bibinfo{author}{Bhatt, R.}, \bibinfo{author}{Goswami, A.},
  \bibinfo{year}{2019}.
\newblock \bibinfo{title}{The big data system, components, tools, and
  technologies: a survey}.
\newblock \bibinfo{journal}{Knowledge and Information Systems}
  \bibinfo{volume}{60}, \bibinfo{pages}{1165--1245}.
\bibitem[{Rocha~Neto et~al.(2020)Rocha~Neto, Delicato, Batista and
  Pires}]{rocha2020distributed}
\bibinfo{author}{Rocha~Neto, A.F.}, \bibinfo{author}{Delicato, F.C.},
  \bibinfo{author}{Batista, T.V.}, \bibinfo{author}{Pires, P.F.},
  \bibinfo{year}{2020}.
\newblock \bibinfo{title}{Distributed machine learning for iot applications in
  the fog}.
\newblock \bibinfo{journal}{Fog Computing: Theory and Practice} ,
  \bibinfo{pages}{309--345}.
\bibitem[{Rong et~al.(2021)Rong, Xu, Tong and Fan}]{rong2021edge}
\bibinfo{author}{Rong, G.}, \bibinfo{author}{Xu, Y.}, \bibinfo{author}{Tong,
  X.}, \bibinfo{author}{Fan, H.}, \bibinfo{year}{2021}.
\newblock \bibinfo{title}{An edge-cloud collaborative computing platform for
  building aiot applications efficiently} .
\bibitem[{Rosendo et~al.(2021a)Rosendo, Costan, Antoniu, Simonin, Lombardo,
  Joly and Valduriez}]{rosendo2021reproducible}
\bibinfo{author}{Rosendo, D.}, \bibinfo{author}{Costan, A.},
  \bibinfo{author}{Antoniu, G.}, \bibinfo{author}{Simonin, M.},
  \bibinfo{author}{Lombardo, J.C.}, \bibinfo{author}{Joly, A.},
  \bibinfo{author}{Valduriez, P.}, \bibinfo{year}{2021}a.
\newblock \bibinfo{title}{Reproducible performance optimization of complex
  applications on the edge-to-cloud continuum}.
\newblock \bibinfo{journal}{arXiv preprint arXiv:2108.04033} .
\bibitem[{Rosendo et~al.(2021b)Rosendo, Costan, Antoniu and
  Valduriez}]{rosendo2021e2clab}
\bibinfo{author}{Rosendo, D.}, \bibinfo{author}{Costan, A.},
  \bibinfo{author}{Antoniu, G.}, \bibinfo{author}{Valduriez, P.},
  \bibinfo{year}{2021}b.
\newblock \bibinfo{title}{E2clab: Reproducible analysis of complex workflows on
  the edge-to-cloud continuum}, in: \bibinfo{booktitle}{IEEE 35th International
  Parallel and Distributed Processing Symposium (IPDS 2021)}.
\bibitem[{Rosendo et~al.(2020)Rosendo, Silva, Simonin, Costan and
  Antoniu}]{rosendo2020e2clab}
\bibinfo{author}{Rosendo, D.}, \bibinfo{author}{Silva, P.},
  \bibinfo{author}{Simonin, M.}, \bibinfo{author}{Costan, A.},
  \bibinfo{author}{Antoniu, G.}, \bibinfo{year}{2020}.
\newblock \bibinfo{title}{E2clab: Exploring the computing continuum through
  repeatable, replicable and reproducible edge-to-cloud experiments}, in:
  \bibinfo{booktitle}{2020 IEEE International Conference on Cluster Computing
  (CLUSTER)}, \bibinfo{organization}{IEEE}. pp. \bibinfo{pages}{176--186}.
\bibitem[{Sadiq et~al.(2004)Sadiq, Orlowska, Sadiq and Foulger}]{sadiq2004data}
\bibinfo{author}{Sadiq, S.}, \bibinfo{author}{Orlowska, M.},
  \bibinfo{author}{Sadiq, W.}, \bibinfo{author}{Foulger, C.},
  \bibinfo{year}{2004}.
\newblock \bibinfo{title}{{Data Flow and Validation in Workflow Modelling}},
  in: \bibinfo{booktitle}{Proceedings of the 15th Australasian database
  conference-Volume 27}, pp. \bibinfo{pages}{207--214}.
\bibitem[{Samie et~al.(2019)Samie, Bauer and Henkel}]{samie2019cloud}
\bibinfo{author}{Samie, F.}, \bibinfo{author}{Bauer, L.},
  \bibinfo{author}{Henkel, J.}, \bibinfo{year}{2019}.
\newblock \bibinfo{title}{From cloud down to things: An overview of machine
  learning in internet of things}.
\newblock \bibinfo{journal}{IEEE Internet of Things Journal}
  \bibinfo{volume}{6}, \bibinfo{pages}{4921--4934}.
\bibitem[{Sanchez et~al.(2014)Sanchez, Mu{\~n}oz, Galache, Sotres, Santana,
  Gutierrez, Ramdhany, Gluhak, Krco, Theodoridis
  et~al.}]{sanchez2014smartsantander}
\bibinfo{author}{Sanchez, L.}, \bibinfo{author}{Mu{\~n}oz, L.},
  \bibinfo{author}{Galache, J.A.}, \bibinfo{author}{Sotres, P.},
  \bibinfo{author}{Santana, J.R.}, \bibinfo{author}{Gutierrez, V.},
  \bibinfo{author}{Ramdhany, R.}, \bibinfo{author}{Gluhak, A.},
  \bibinfo{author}{Krco, S.}, \bibinfo{author}{Theodoridis, E.}, et~al.,
  \bibinfo{year}{2014}.
\newblock \bibinfo{title}{Smartsantander: Iot experimentation over a smart city
  testbed}.
\newblock \bibinfo{journal}{Computer Networks} \bibinfo{volume}{61},
  \bibinfo{pages}{217--238}.
\bibitem[{Sankaranarayanan et~al.(2020)Sankaranarayanan, Rodrigues, Sugumaran,
  Kozlov et~al.}]{sankaranarayanan2020data}
\bibinfo{author}{Sankaranarayanan, S.}, \bibinfo{author}{Rodrigues, J.J.},
  \bibinfo{author}{Sugumaran, V.}, \bibinfo{author}{Kozlov, S.}, et~al.,
  \bibinfo{year}{2020}.
\newblock \bibinfo{title}{Data flow and distributed deep neural network based
  low latency iot-edge computation model for big data environment}.
\newblock \bibinfo{journal}{Engineering Applications of Artificial
  Intelligence} \bibinfo{volume}{94}, \bibinfo{pages}{103785}.
\bibitem[{Sarabia-J{\'a}come et~al.(2019)Sarabia-J{\'a}come, Lacalle, Palau and
  Esteve}]{sarabia2019efficient}
\bibinfo{author}{Sarabia-J{\'a}come, D.}, \bibinfo{author}{Lacalle, I.},
  \bibinfo{author}{Palau, C.E.}, \bibinfo{author}{Esteve, M.},
  \bibinfo{year}{2019}.
\newblock \bibinfo{title}{Efficient deployment of predictive analytics in edge
  gateways: Fall detection scenario}, in: \bibinfo{booktitle}{2019 IEEE 5th
  World Forum on Internet of Things (WF-IoT)}, \bibinfo{organization}{IEEE}.
  pp. \bibinfo{pages}{41--46}.
\bibitem[{Sarabia-J{\'a}come et~al.(2020)Sarabia-J{\'a}come, Usach, Palau and
  Esteve}]{sarabia2020highly}
\bibinfo{author}{Sarabia-J{\'a}come, D.}, \bibinfo{author}{Usach, R.},
  \bibinfo{author}{Palau, C.E.}, \bibinfo{author}{Esteve, M.},
  \bibinfo{year}{2020}.
\newblock \bibinfo{title}{Highly-efficient fog-based deep learning aal fall
  detection system}.
\newblock \bibinfo{journal}{Internet of Things} \bibinfo{volume}{11},
  \bibinfo{pages}{100185}.
\bibitem[{Sergeev and Del~Balso(2018)}]{sergeev2018horovod}
\bibinfo{author}{Sergeev, A.}, \bibinfo{author}{Del~Balso, M.},
  \bibinfo{year}{2018}.
\newblock \bibinfo{title}{Horovod: fast and easy distributed deep learning in
  tensorflow}.
\newblock \bibinfo{journal}{arXiv preprint arXiv:1802.05799} .
\bibitem[{Sezer et~al.(2017)Sezer, Dogdu and Ozbayoglu}]{sezer2017context}
\bibinfo{author}{Sezer, O.B.}, \bibinfo{author}{Dogdu, E.},
  \bibinfo{author}{Ozbayoglu, A.M.}, \bibinfo{year}{2017}.
\newblock \bibinfo{title}{Context-aware computing, learning, and big data in
  internet of things: a survey}.
\newblock \bibinfo{journal}{IEEE Internet of Things Journal}
  \bibinfo{volume}{5}, \bibinfo{pages}{1--27}.
\bibitem[{Sharma and Wang(2017)}]{sharma2017live}
\bibinfo{author}{Sharma, S.K.}, \bibinfo{author}{Wang, X.},
  \bibinfo{year}{2017}.
\newblock \bibinfo{title}{Live data analytics with collaborative edge and cloud
  processing in wireless iot networks}.
\newblock \bibinfo{journal}{IEEE Access} \bibinfo{volume}{5},
  \bibinfo{pages}{4621--4635}.
\bibitem[{Shin et~al.(2017)Shin, Lee, Kim and Kim}]{shin2017continual}
\bibinfo{author}{Shin, H.}, \bibinfo{author}{Lee, J.K.}, \bibinfo{author}{Kim,
  J.}, \bibinfo{author}{Kim, J.}, \bibinfo{year}{2017}.
\newblock \bibinfo{title}{Continual learning with deep generative replay}.
\newblock \bibinfo{journal}{arXiv preprint arXiv:1705.08690} .
\bibitem[{Simpson et~al.(2001)Simpson, Mauery, Korte and
  Mistree}]{simpson2001kriging}
\bibinfo{author}{Simpson, T.W.}, \bibinfo{author}{Mauery, T.M.},
  \bibinfo{author}{Korte, J.J.}, \bibinfo{author}{Mistree, F.},
  \bibinfo{year}{2001}.
\newblock \bibinfo{title}{Kriging models for global approximation in
  simulation-based multidisciplinary design optimization}.
\newblock \bibinfo{journal}{AIAA journal} \bibinfo{volume}{39},
  \bibinfo{pages}{2233--2241}.
\bibitem[{Sitton~Candanedo et~al.(2020)}]{sitton2020geca}
\bibinfo{author}{Sitton~Candanedo, I.X.}, et~al., \bibinfo{year}{2020}.
\newblock \bibinfo{title}{Geca: A global edge computing architecture} .
\bibitem[{Snoek et~al.(2012)Snoek, Larochelle and Adams}]{snoek2012practical}
\bibinfo{author}{Snoek, J.}, \bibinfo{author}{Larochelle, H.},
  \bibinfo{author}{Adams, R.P.}, \bibinfo{year}{2012}.
\newblock \bibinfo{title}{Practical bayesian optimization of machine learning
  algorithms}.
\newblock \bibinfo{journal}{arXiv preprint arXiv:1206.2944} .
\bibitem[{Sonmez et~al.(2018)Sonmez, Ozgovde and
  Ersoy}]{sonmez2018edgecloudsim}
\bibinfo{author}{Sonmez, C.}, \bibinfo{author}{Ozgovde, A.},
  \bibinfo{author}{Ersoy, C.}, \bibinfo{year}{2018}.
\newblock \bibinfo{title}{Edgecloudsim: An environment for performance
  evaluation of edge computing systems}.
\newblock \bibinfo{journal}{Transactions on Emerging Telecommunications
  Technologies} \bibinfo{volume}{29}, \bibinfo{pages}{e3493}.
\bibitem[{Spataru(2021)}]{spataru2021review}
\bibinfo{author}{Spataru, A.}, \bibinfo{year}{2021}.
\newblock \bibinfo{title}{A review of blockchain-enabled fog computing in the
  cloud continuum context}.
\newblock \bibinfo{journal}{Scalable Computing: Practice and Experience}
  \bibinfo{volume}{22}, \bibinfo{pages}{463--468}.
\bibitem[{Sreerangaraju(2020)}]{sreerangaraju}
\bibinfo{author}{Sreerangaraju, S.}, \bibinfo{year}{2020}.
\newblock \bibinfo{title}{Emulation vs. simulation}.
\newblock
  \bibinfo{howpublished}{\url{https://www.perfecto.io/blog/emulation-vs-simulation}}.
\bibitem[{Stodden and Miguez(2013)}]{stodden2013best}
\bibinfo{author}{Stodden, V.}, \bibinfo{author}{Miguez, S.},
  \bibinfo{year}{2013}.
\newblock \bibinfo{title}{{Best Practices for Computational Science: Software
  Infrastructure and Environments for Reproducible and Extensible Research}}.
\newblock \bibinfo{journal}{Available at SSRN 2322276} .
\bibitem[{Struye et~al.(2018)Struye, Braem, Latr{\'e} and
  Marquez-Barja}]{struye2018citylab}
\bibinfo{author}{Struye, J.}, \bibinfo{author}{Braem, B.},
  \bibinfo{author}{Latr{\'e}, S.}, \bibinfo{author}{Marquez-Barja, J.},
  \bibinfo{year}{2018}.
\newblock \bibinfo{title}{The citylab testbed—large-scale multi-technology
  wireless experimentation in a city environment: Neural network-based
  interference prediction in a smart city}, in: \bibinfo{booktitle}{IEEE
  INFOCOM 2018-IEEE Conference on Computer Communications Workshops (INFOCOM
  WKSHPS)}, \bibinfo{organization}{IEEE}. pp. \bibinfo{pages}{529--534}.
\bibitem[{Sutton and Barto(2018)}]{sutton2018reinforcement}
\bibinfo{author}{Sutton, R.S.}, \bibinfo{author}{Barto, A.G.},
  \bibinfo{year}{2018}.
\newblock \bibinfo{title}{Reinforcement learning: An introduction}.
\newblock \bibinfo{publisher}{MIT press}.
\bibitem[{Svorobej et~al.(2019)Svorobej, Takako~Endo, Bendechache,
  Filelis-Papadopoulos, Giannoutakis, Gravvanis, Tzovaras, Byrne and
  Lynn}]{svorobej2019simulating}
\bibinfo{author}{Svorobej, S.}, \bibinfo{author}{Takako~Endo, P.},
  \bibinfo{author}{Bendechache, M.}, \bibinfo{author}{Filelis-Papadopoulos,
  C.}, \bibinfo{author}{Giannoutakis, K.M.}, \bibinfo{author}{Gravvanis, G.A.},
  \bibinfo{author}{Tzovaras, D.}, \bibinfo{author}{Byrne, J.},
  \bibinfo{author}{Lynn, T.}, \bibinfo{year}{2019}.
\newblock \bibinfo{title}{Simulating fog and edge computing scenarios: An
  overview and research challenges}.
\newblock \bibinfo{journal}{Future Internet} \bibinfo{volume}{11},
  \bibinfo{pages}{55}.
\bibitem[{Talagala et~al.(2018)Talagala, Sundararaman, Sridhar, Arteaga, Luo,
  Subramanian, Ghanta, Khermosh and Roselli}]{talagala2018eco}
\bibinfo{author}{Talagala, N.}, \bibinfo{author}{Sundararaman, S.},
  \bibinfo{author}{Sridhar, V.}, \bibinfo{author}{Arteaga, D.},
  \bibinfo{author}{Luo, Q.}, \bibinfo{author}{Subramanian, S.},
  \bibinfo{author}{Ghanta, S.}, \bibinfo{author}{Khermosh, L.},
  \bibinfo{author}{Roselli, D.}, \bibinfo{year}{2018}.
\newblock \bibinfo{title}{$\{$ECO$\}$: Harmonizing edge and cloud with ml/dl
  orchestration}, in: \bibinfo{booktitle}{$\{$USENIX$\}$ Workshop on Hot Topics
  in Edge Computing (HotEdge 18)}.
\bibitem[{Ulusar et~al.(2020)Ulusar, Ozcan and Al-Turjman}]{ulusar2020open}
\bibinfo{author}{Ulusar, U.D.}, \bibinfo{author}{Ozcan, D.G.},
  \bibinfo{author}{Al-Turjman, F.}, \bibinfo{year}{2020}.
\newblock \bibinfo{title}{Open source tools for machine learning with big data
  in smart cities}, in: \bibinfo{booktitle}{Smart Cities Performability,
  Cognition, \& Security}. \bibinfo{publisher}{Springer}, pp.
  \bibinfo{pages}{153--168}.
\bibitem[{Vanhove et~al.(2015)Vanhove, Van~Seghbroeck, Wauters, De~Turck,
  Vermeulen and Demeester}]{vanhove2015tengu}
\bibinfo{author}{Vanhove, T.}, \bibinfo{author}{Van~Seghbroeck, G.},
  \bibinfo{author}{Wauters, T.}, \bibinfo{author}{De~Turck, F.},
  \bibinfo{author}{Vermeulen, B.}, \bibinfo{author}{Demeester, P.},
  \bibinfo{year}{2015}.
\newblock \bibinfo{title}{Tengu: An experimentation platform for big data
  applications}, in: \bibinfo{booktitle}{2015 IEEE 35th International
  Conference on Distributed Computing Systems Workshops},
  \bibinfo{organization}{IEEE}. pp. \bibinfo{pages}{42--47}.
\bibitem[{Verma and Fatima(2020)}]{verma2020smart}
\bibinfo{author}{Verma, P.}, \bibinfo{author}{Fatima, S.},
  \bibinfo{year}{2020}.
\newblock \bibinfo{title}{Smart healthcare applications and real-time analytics
  through edge computing}, in: \bibinfo{booktitle}{Internet of Things Use Cases
  for the Healthcare Industry}. \bibinfo{publisher}{Springer}, pp.
  \bibinfo{pages}{241--270}.
\bibitem[{Verma et~al.(2017)Verma, Kawamoto, Fadlullah, Nishiyama and
  Kato}]{verma2017survey}
\bibinfo{author}{Verma, S.}, \bibinfo{author}{Kawamoto, Y.},
  \bibinfo{author}{Fadlullah, Z.M.}, \bibinfo{author}{Nishiyama, H.},
  \bibinfo{author}{Kato, N.}, \bibinfo{year}{2017}.
\newblock \bibinfo{title}{A survey on network methodologies for real-time
  analytics of massive iot data and open research issues}.
\newblock \bibinfo{journal}{IEEE Communications Surveys \& Tutorials}
  \bibinfo{volume}{19}, \bibinfo{pages}{1457--1477}.
\bibitem[{Vermesan et~al.(2020)Vermesan, Bahr, Ottella, Serrano, Karlsen,
  Wahlstr{\o}m, Sand, Ashwathnarayan and Gamba}]{vermesan2020internet}
\bibinfo{author}{Vermesan, O.}, \bibinfo{author}{Bahr, R.},
  \bibinfo{author}{Ottella, M.}, \bibinfo{author}{Serrano, M.},
  \bibinfo{author}{Karlsen, T.}, \bibinfo{author}{Wahlstr{\o}m, T.},
  \bibinfo{author}{Sand, H.E.}, \bibinfo{author}{Ashwathnarayan, M.},
  \bibinfo{author}{Gamba, M.T.}, \bibinfo{year}{2020}.
\newblock \bibinfo{title}{Internet of robotic things intelligent connectivity
  and platforms}.
\newblock \bibinfo{journal}{Frontiers in Robotics and AI} \bibinfo{volume}{7},
  \bibinfo{pages}{104}.
\bibitem[{V{\'e}stias et~al.(2020)V{\'e}stias, Duarte, de~Sousa and
  Neto}]{vestias2020moving}
\bibinfo{author}{V{\'e}stias, M.P.}, \bibinfo{author}{Duarte, R.P.},
  \bibinfo{author}{de~Sousa, J.T.}, \bibinfo{author}{Neto, H.C.},
  \bibinfo{year}{2020}.
\newblock \bibinfo{title}{Moving deep learning to the edge}.
\newblock \bibinfo{journal}{Algorithms} \bibinfo{volume}{13},
  \bibinfo{pages}{125}.
\bibitem[{Wang et~al.(2018)Wang, Chen, Song, Guizani, Yu and Du}]{wang2018iot}
\bibinfo{author}{Wang, D.}, \bibinfo{author}{Chen, D.}, \bibinfo{author}{Song,
  B.}, \bibinfo{author}{Guizani, N.}, \bibinfo{author}{Yu, X.},
  \bibinfo{author}{Du, X.}, \bibinfo{year}{2018}.
\newblock \bibinfo{title}{From iot to 5g i-iot: The next generation iot-based
  intelligent algorithms and 5g technologies}.
\newblock \bibinfo{journal}{IEEE Communications Magazine} \bibinfo{volume}{56},
  \bibinfo{pages}{114--120}.
\bibitem[{Wang and Li(2018)}]{wang2018adaptive}
\bibinfo{author}{Wang, J.}, \bibinfo{author}{Li, D.}, \bibinfo{year}{2018}.
\newblock \bibinfo{title}{Adaptive computing optimization in software-defined
  network-based industrial internet of things with fog computing}.
\newblock \bibinfo{journal}{Sensors} \bibinfo{volume}{18},
  \bibinfo{pages}{2509}.
\bibitem[{Wang et~al.(2000)Wang, Chen, Qian and Ye}]{wang2000optimization}
\bibinfo{author}{Wang, X.}, \bibinfo{author}{Chen, B.}, \bibinfo{author}{Qian,
  G.}, \bibinfo{author}{Ye, F.}, \bibinfo{year}{2000}.
\newblock \bibinfo{title}{On the optimization of fuzzy decision trees}.
\newblock \bibinfo{journal}{Fuzzy Sets and Systems} \bibinfo{volume}{112},
  \bibinfo{pages}{117--125}.
\bibitem[{Wette et~al.(2014)Wette, Dr{\"a}xler, Schwabe, Wallaschek, Zahraee
  and Karl}]{wette2014maxinet}
\bibinfo{author}{Wette, P.}, \bibinfo{author}{Dr{\"a}xler, M.},
  \bibinfo{author}{Schwabe, A.}, \bibinfo{author}{Wallaschek, F.},
  \bibinfo{author}{Zahraee, M.H.}, \bibinfo{author}{Karl, H.},
  \bibinfo{year}{2014}.
\newblock \bibinfo{title}{Maxinet: Distributed emulation of software-defined
  networks}, in: \bibinfo{booktitle}{2014 IFIP Networking Conference},
  \bibinfo{organization}{IEEE}. pp. \bibinfo{pages}{1--9}.
\bibitem[{Wiering and Van~Otterlo(2012)}]{wiering2012reinforcement}
\bibinfo{author}{Wiering, M.A.}, \bibinfo{author}{Van~Otterlo, M.},
  \bibinfo{year}{2012}.
\newblock \bibinfo{title}{Reinforcement learning}.
\newblock \bibinfo{journal}{Adaptation, learning, and optimization}
  \bibinfo{volume}{12}.
\bibitem[{Xia et~al.(2018)Xia, Etchevers, Letondeur, Lebre, Coupaye and
  Desprez}]{xia2018combining}
\bibinfo{author}{Xia, Y.}, \bibinfo{author}{Etchevers, X.},
  \bibinfo{author}{Letondeur, L.}, \bibinfo{author}{Lebre, A.},
  \bibinfo{author}{Coupaye, T.}, \bibinfo{author}{Desprez, F.},
  \bibinfo{year}{2018}.
\newblock \bibinfo{title}{Combining heuristics to optimize and scale the
  placement of iot applications in the fog}, in: \bibinfo{booktitle}{2018
  IEEE/ACM 11th International Conference on Utility and Cloud Computing (UCC)},
  \bibinfo{organization}{IEEE}. pp. \bibinfo{pages}{153--163}.
\bibitem[{Xian(2020)}]{xian2020parallel}
\bibinfo{author}{Xian, G.}, \bibinfo{year}{2020}.
\newblock \bibinfo{title}{Parallel machine learning algorithm using
  fine-grained-mode spark on a mesos big data cloud computing software
  framework for mobile robotic intelligent fault recognition}.
\newblock \bibinfo{journal}{IEEE Access} \bibinfo{volume}{8},
  \bibinfo{pages}{131885--131900}.
\bibitem[{Xiao et~al.(2017)Xiao, Xue, Miao, Li, Chen, Wu, Li and
  Zhou}]{xiao2017tux2}
\bibinfo{author}{Xiao, W.}, \bibinfo{author}{Xue, J.}, \bibinfo{author}{Miao,
  Y.}, \bibinfo{author}{Li, Z.}, \bibinfo{author}{Chen, C.},
  \bibinfo{author}{Wu, M.}, \bibinfo{author}{Li, W.}, \bibinfo{author}{Zhou,
  L.}, \bibinfo{year}{2017}.
\newblock \bibinfo{title}{Tux$^2$: Distributed graph computation for machine
  learning}, in: \bibinfo{booktitle}{14th $\{$USENIX$\}$ symposium on networked
  systems design and implementation ($\{$NSDI$\}$ 17)}, pp.
  \bibinfo{pages}{669--682}.
\bibitem[{Xiong et~al.(2018)Xiong, Sun, Xing and Huang}]{xiong2018extend}
\bibinfo{author}{Xiong, Y.}, \bibinfo{author}{Sun, Y.}, \bibinfo{author}{Xing,
  L.}, \bibinfo{author}{Huang, Y.}, \bibinfo{year}{2018}.
\newblock \bibinfo{title}{Extend cloud to edge with kubeedge}, in:
  \bibinfo{booktitle}{2018 IEEE/ACM Symposium on Edge Computing (SEC)},
  \bibinfo{organization}{IEEE}. pp. \bibinfo{pages}{373--377}.
\bibitem[{Xu et~al.(2018)Xu, Yu, Griffith and Golmie}]{xu2018survey}
\bibinfo{author}{Xu, H.}, \bibinfo{author}{Yu, W.}, \bibinfo{author}{Griffith,
  D.}, \bibinfo{author}{Golmie, N.}, \bibinfo{year}{2018}.
\newblock \bibinfo{title}{A survey on industrial internet of things: A
  cyber-physical systems perspective}.
\newblock \bibinfo{journal}{IEEE Access} \bibinfo{volume}{6},
  \bibinfo{pages}{78238--78259}.
\bibitem[{Xu et~al.(2021)Xu, Venkataraman, Gupta, Mai and
  Potharaju}]{xu2021move}
\bibinfo{author}{Xu, L.}, \bibinfo{author}{Venkataraman, S.},
  \bibinfo{author}{Gupta, I.}, \bibinfo{author}{Mai, L.},
  \bibinfo{author}{Potharaju, R.}, \bibinfo{year}{2021}.
\newblock \bibinfo{title}{Move fast and meet deadlines: Fine-grained real-time
  stream processing with cameo}, in: \bibinfo{booktitle}{18th $\{$USENIX$\}$
  Symposium on Networked Systems Design and Implementation ($\{$NSDI$\}$ 21)},
  pp. \bibinfo{pages}{389--405}.
\bibitem[{Xu et~al.(2019)Xu, Liu, Liu, Lin, Liu and Liu}]{xu2019first}
\bibinfo{author}{Xu, M.}, \bibinfo{author}{Liu, J.}, \bibinfo{author}{Liu, Y.},
  \bibinfo{author}{Lin, F.X.}, \bibinfo{author}{Liu, Y.}, \bibinfo{author}{Liu,
  X.}, \bibinfo{year}{2019}.
\newblock \bibinfo{title}{A first look at deep learning apps on smartphones},
  in: \bibinfo{booktitle}{The World Wide Web Conference}, pp.
  \bibinfo{pages}{2125--2136}.
\bibitem[{Yang(2017)}]{yang2017iot}
\bibinfo{author}{Yang, S.}, \bibinfo{year}{2017}.
\newblock \bibinfo{title}{Iot stream processing and analytics in the fog}.
\newblock \bibinfo{journal}{IEEE Communications Magazine} \bibinfo{volume}{55},
  \bibinfo{pages}{21--27}.
\bibitem[{Yousefpour et~al.(2019)Yousefpour, Fung, Nguyen, Kadiyala, Jalali,
  Niakanlahiji, Kong and Jue}]{yousefpour2019all}
\bibinfo{author}{Yousefpour, A.}, \bibinfo{author}{Fung, C.},
  \bibinfo{author}{Nguyen, T.}, \bibinfo{author}{Kadiyala, K.},
  \bibinfo{author}{Jalali, F.}, \bibinfo{author}{Niakanlahiji, A.},
  \bibinfo{author}{Kong, J.}, \bibinfo{author}{Jue, J.P.},
  \bibinfo{year}{2019}.
\newblock \bibinfo{title}{All one needs to know about fog computing and related
  edge computing paradigms: A complete survey}.
\newblock \bibinfo{journal}{Journal of Systems Architecture}
  \bibinfo{volume}{98}, \bibinfo{pages}{289--330}.
\bibitem[{Zhang et~al.(2019)Zhang, Patras and Haddadi}]{zhang2019deep}
\bibinfo{author}{Zhang, C.}, \bibinfo{author}{Patras, P.},
  \bibinfo{author}{Haddadi, H.}, \bibinfo{year}{2019}.
\newblock \bibinfo{title}{Deep learning in mobile and wireless networking: A
  survey}.
\newblock \bibinfo{journal}{IEEE Communications surveys \& tutorials}
  \bibinfo{volume}{21}, \bibinfo{pages}{2224--2287}.
\bibitem[{Zhang et~al.(2020)Zhang, Zhang, Lane, Shu, Zeng, Fang, Yan and
  Xu}]{zhang2020deep}
\bibinfo{author}{Zhang, M.}, \bibinfo{author}{Zhang, F.},
  \bibinfo{author}{Lane, N.D.}, \bibinfo{author}{Shu, Y.},
  \bibinfo{author}{Zeng, X.}, \bibinfo{author}{Fang, B.}, \bibinfo{author}{Yan,
  S.}, \bibinfo{author}{Xu, H.}, \bibinfo{year}{2020}.
\newblock \bibinfo{title}{Deep learning in the era of edge computing:
  Challenges and opportunities}.
\newblock \bibinfo{journal}{Fog Computing: Theory and Practice} ,
  \bibinfo{pages}{67--78}.
\bibitem[{Zhang et~al.(2018)Zhang, Wang and Shi}]{zhang2018pcamp}
\bibinfo{author}{Zhang, X.}, \bibinfo{author}{Wang, Y.}, \bibinfo{author}{Shi,
  W.}, \bibinfo{year}{2018}.
\newblock \bibinfo{title}{pcamp: Performance comparison of machine learning
  packages on the edges}, in: \bibinfo{booktitle}{$\{$USENIX$\}$ Workshop on
  Hot Topics in Edge Computing (HotEdge 18)}.
\bibitem[{Zhang and Yang(2017)}]{zhang2017survey}
\bibinfo{author}{Zhang, Y.}, \bibinfo{author}{Yang, Q.}, \bibinfo{year}{2017}.
\newblock \bibinfo{title}{A survey on multi-task learning}.
\newblock \bibinfo{journal}{arXiv preprint arXiv:1707.08114} .
\bibitem[{Zhou et~al.(2019)Zhou, Wen, Teodorescu and Du}]{zhou2019distributing}
\bibinfo{author}{Zhou, L.}, \bibinfo{author}{Wen, H.},
  \bibinfo{author}{Teodorescu, R.}, \bibinfo{author}{Du, D.H.},
  \bibinfo{year}{2019}.
\newblock \bibinfo{title}{Distributing deep neural networks with containerized
  partitions at the edge}, in: \bibinfo{booktitle}{2nd $\{$USENIX$\}$ Workshop
  on Hot Topics in Edge Computing (HotEdge 19)}.
\bibitem[{Zhou et~al.(2020)Zhou, Wu, Liang, Sun, Xu and Luo}]{zhou2020saface}
\bibinfo{author}{Zhou, Z.}, \bibinfo{author}{Wu, B.}, \bibinfo{author}{Liang,
  Z.}, \bibinfo{author}{Sun, G.}, \bibinfo{author}{Xu, C.},
  \bibinfo{author}{Luo, G.}, \bibinfo{year}{2020}.
\newblock \bibinfo{title}{Saface: Towards scenario-aware face recognition via
  edge computing system}, in: \bibinfo{booktitle}{3rd $\{$USENIX$\}$ Workshop
  on Hot Topics in Edge Computing (HotEdge 20)}.

\end{thebibliography}

\end{document}